Kurdistan Region – Iraq
Kurdistan Regional Government
Ministry of Higher Education and Scientific Research

Sulaimani Polytechnic University

Kurdistan Institution for Strategic Studies and Scientific Research

# Artificial Intelligence Algorithms for Natural Language Processing and the Semantic Web Ontology Learning

A Doctoral Thesis

Submitted to the Council of Technical College of Informatics, Sulaimani Polytechnic University in Collaboration with Kurdistan Institution for Strategic Studies and Scientific Research in Partial Fulfilment of the Requirements for the Joint Degree of Doctor of Philosophy (Ph.D.) of Science in Information Technology

By

**Bryar Ahmad Hassan**

MSc. Software Engineering - University of Southampton, UK (2013)

Supervised by

**Dr Tarik A. Rashid**

Professor

February, 2021 Rebandan 2720

# Acknowledgements

This thesis would not have been possible without the sincere encouragements and support of my supervisor, Professor Dr Tarik A. Rashid. His supports and recommendations have tremendously assisted me while the thesis was on progress. Thus, I am truly indebted and thankful to him for his inspiration and guidance throughout the course of my research.

I should also take this opportunity to thank Sulaimani Polytechnic University and Kurdistan Institution for Strategic Studies and Scientific Research that offered me endless support.

# Abstract

Evolutionary clustering algorithms have considered as the most popular and widely used evolutionary algorithms for minimising optimisation and practical problems in nearly all fields, including natural language processing and the Semantic Web ontology learning. In this thesis, a new evolutionary clustering algorithm star (ECA*) is proposed, which is based on statistical and meta-heuristic approaches for analysing heterogeneous and multi-featured clustering datasets stochastically. The rationale behind choosing BSA among other evolutionary and swarm algorithms is related to the experiment conducted in this thesis. The experimental results indicate the success of BSA over its counterpart algorithms in solving different cohorts of numerical optimisation problems such as problems with different levels of hardness score, problem dimensions, and search spaces.

Additionally, a number of experiments were conducted to evaluate ECA* against five state-of-the-art approaches: K-means (KM), K-means++ (KM++), expectation maximisation (EM), learning vector quantisation (LVQ), and genetic algorithm for clustering++ (GENCLUST++). For this, 32 heterogeneous and multi-featured datasets were used to examine their performance using internal and external clustering measures, and to measure the sensitivity of their performance towards dataset features in the form of operational framework. The results indicate that ECA* overcomes its competitive techniques in terms of the ability to find the right clusters. Significantly, ECA* was less sensitive towards the dataset properties compared with its counterpart techniques. Thus, ECA* was the best-performed one among the aforementioned counterpart algorithms. Meanwhile, the overall performance rank of ECA* is 1.1** for 32 datasets according to the dataset features. Based on its superior performance, exploiting and adapting ECA* on the ontology learning had a vital possibility. In the process of deriving concept hierarchies from corpora, generating formal context may lead to a time-consuming process. Therefore, formal context size reduction results in removing uninterested and erroneous pairs, taking less time to extract the concept lattice and concept hierarchies accordingly. In this premise, this work aims to propose a framework to reduce the ambiguity of the formal context of the existing framework using an adaptive version of ECA*. In turn, an experiment was conducted by applying 385 sample corpora from Wikipedia on the two frameworks to examine the reduction of formal context size, which leads to yield concept lattice and concept hierarchy. The resulting lattice of formal context was evaluated to the original one using concept lattice-invariants. Accordingly, the homomorphic between the two lattices preserves the quality of resulting concept hierarchies by 89% in contrast to the basic ones, and the reduced concept lattice inherits the structural relation of the original one.

# Table of Contents

# List of Figures

# List of Tables

# List of Algorithms

# List of Abbreviations

| Abbreviation | Description |
|---|---|
| ABC | Artificial Bee Colony |
| ABS | Abaffy Broyden Spedicato |
| AI | Artificial Intelligence |
| BSA | Backtracking Search Optimisation Algorithm |
| C | Cluster centroids |
| CIn | Computational Intelligence |
| CI | Centroid Index |
| CLPSO | Comprehensive Learning Particle Swarm Optimisation |
| CMAES | Covariance Matrix Adaptation Evolution Strategy |
| COBSA | Improved BSA Based on Population Control Factor And Optimal Learning Strategy |
| F | Control parameter |
| COPs | Constrained Optimisation Problems |
| CSI | Centroid Similarity Index |
| DcoRef | Coreference Resolution |
| DE | Differential Evolution |
| DS | Differential Search |
| EA | Evolutionary Algorithms |
| ECA* | Evolutionary Clustering Algorithm Star |
| EM | Expectation Maximisation |
| FCA | Formal Context Analysis |
| FF | Firefly Algorithm |
| GA | Genetic Algorithm |
| GENCLUST++ | Genetic Algorithm for Clustering++ |
| GUI | Graphical User Interface |
| HI | Historical Cluster Centroids |
| JDE | Self-adaptive Differential Evolution |
| K | Number of Clusters |
| KM | K-Means |
| KM++ | K-Means++ |
| LFO | Levy Flight Optimisation |
| LVQ | Learning Vector Quantisation |
| MSE | Mean Squared Error |
| Mut-over | Mutation-crossover |
| NER | Named Entity Recognition |
| NIA | Nature-Inspired Algorithms |
| NIC | Nature-Inspired Computing |
| NLP | Natural Language Processing |
| NMI | Normalised Mutual Information |
| Parse | Syntactic Parsing |
| POS | Part-Of-Speech |
| POST | Part-Of-Speech Tagger |
| PSO | Particle Swarm Optimisation |
| S | The Number of Social Class Ranks |
| SSE | Sum of Squared Error |
| SSplit | Sentence Splitting |
| ε- ratio | Approximation ratio |

# List of Symbols

| Symbol | Description |
|---|---|
| α | Alpha |
| $\sigma$ | Sigma |
| ε- ratio | Approximation ratio |
| $\varepsilon$ | A very small number, near zero |
| * | Star |

# Chapter One: Introduction

## 1.1 Overview

Since the invention of computer, a great deal of attention has been given to the quest for the unknown and the best solution. The first form of a search-based algorithm was developed by Turing to crack German Enigma ciphers in World War II [1]. Hundreds of types of algorithms have been developed to date, including optimisation problems, for various purposes. Optimisation algorithms are used to find acceptable problem solutions. Many different solutions may be found for a single problem, but the optimal solution is preferred. Optimisation concerns are typically nonlinear with a dynamic environment. Optimisation algorithms can generally be divided into Traditional and evolutionary algorithms. Traditional algorithms include gradient and quadratic programming. Heuristic algorithms and other hybrid strategies are used in evolutionary algorithms. This thesis primarily combines several multi-disciplinary areas of research, such as Soft Computing, Artificial Intelligence, Natural Language Processing, Text Mining, and Ontology Learning of the Semantic Web. On this premise, this section specifically consists of three specific sub-sections, which are: evolutionary algorithms, evolutionary clustering algorithms, and the Semantic Web ontology learning.

### 1.1.1 Evolutionary Algorithms

Evolutionary algorithms are a heuristic approach to problem-solving that cannot be solved easily in polynomial cycles, such as non-deterministic polynomial-time (NP) problems, and anything that would take the way too long to complete [2]. They are usually applied on their own to combinatorial problems. However, genetic algorithms are also used in combination with other approaches as quickly as possible to find a slightly suitable starting point to work off another algorithm. The idea of evolutionary algorithm (EA) is very simple if you know the natural selection process. An EA usually includes four overall steps [3]: initialisation, selection, genetic operators, and termination/exit. Each of these steps is roughly a part of the natural selection and offers simple ways to modularise implementations for this group of algorithms. Fitter members of an EA survive and spread, while unfit members i.e., and not, like natural selection, add to the gene pool of other generations.

### 1.1.2 Evolutionary Clustering Algorithms

Due to its unsupervised nature, clustering is considered one of the most difficult and challenging problems in machine learning. The unsupervised nature of the problem means that its structural characteristics are not known unless certain domain knowledge is available in advance. In particular, unknown is the spatial distribution of data in terms of cluster numbers, sizes, densities, shapes, and orientations [4]. This can be further potentialised by the possible need to deal with data objects defined in different nature characteristics (binary, continuous, discrete, and categorical), conditions (partially missing and complete), and scales (ordinal and nominal) [5]. Furthermore, clustering is formally known as a specific type of NP-hard grouping problem from an optimisation perspective [6]. It has stimulated the quest for efficient approximation algorithms, not only by using ad hoc heuristics for particular groups or problems but also by using metaheuristics for general purposes. Metaheuristics are commonly believed to be efficient on NP-hard problems and can provide near-optimal solutions to these problems in a reasonable time. According to this theory, various evolutionary algorithms have been proposed in the literature to solve cluster problems. These algorithms are based on optimising certain objective functions (fitness function), which direct the evolutionary search.

### 1.1.3 The Semantic Web Ontology Learning

Since the beginning of the twenty-first century, unstructured data in the form of electronic news and scientific literature grew rapidly with the development of technology in different domains. By the beginning of this century, however, the web was not effective. When one author wrote on one website about a certain topic, another author may provide conflicting details on a different website about the same topic. In other words, the Web has been disconnected, unreliable, and dumb. It was an incorrect method to derive valuable information from such a web system. To tackle this issue, Maedche and Staab presented the idea of the Semantic Web in 2001 [7]. The underlying purpose behind this concept was to build a highly linked, reliable, and intelligent web platform. Ontologies play a key role in the implementation of the Semantic Web concept. An ontology describes the principles and relationships of a specific conceptualisation systematically and syntactically [8]. It can be defined more precisely as concepts, relationships, attributes, and hierarchies in a domain. Ontologies can be generated using the method called ontology population by extracting specific instances of information from text. Nevertheless, constructing these large ontologies is a difficult task, and ontology for all the available domains cannot be defined [9]. Thus, instead of handcrafting ontologies, research trends are now heading towards automated ontology learning. The process of ontology derivation starts by extracting terms and their synonyms from corpora. After that. corresponding terms and synonyms are combined to construct concepts. Then, the hierarchies between these concepts are found. Lastly, axiom schemata are instantiated and general axioms are extracted from unstructured text. The whole process is considered an ontology learning layer cake.

## 1.2 Problem Statement and Significance of the Thesis

The complexity of the problems of the real world around human beings inhibits the search for any solution simply because of time, space, and costs. This needs low cost, quick and smarter mechanisms. Hence, scholars have investigated the behaviours of animals and natural phenomena to understand how they solve real-world problems. For instance, owing to the adaptability of BSA to different applications and optimisation problems, several scholars have proposed new algorithms based on the original BSA, whereas few others have attempted to employ the original BSA in different applications to solve a variety of problems. Although BSA has been recently developed, it has already been used in several real-world applications and academic fields. It has also been affirmed by [10] that BSA is one of the most used EAs in the field of engineering. On this basis, BSA is considered as one of the most popular and widely used evolutionary algorithms for minimising optimisation and practical problems. Meanwhile, the problem is lack of examining the performance of BSA. To examine this hypothesis, a fair and equitable performance analysis of BSA against its counterpart techniques is necessary.

On the other hand, clustering algorithms have a wide range of applications in nearly all fields, including ontology learning. In the past, several clustering techniques have been proposed; however, these techniques could have the following problems:

- Each of these algorithms was mostly dedicated to a specific type of problem [11];
- Further, not every clustering algorithm needs to perform well in all or almost cohort of datasets and real-world applications, but also it is instead a need to show how sensitive the algorithm towards different types of benchmarking and real-world problems;
- Despite that, almost all the clustering techniques have some common limitations. Therefore, there is an enormous demand for developing a new evolutionary clustering algorithm that is free from the shortcomings of current clustering techniques.

In addition, the Semantic Web is a web of machine-readable data, which provides a program to process data via machine directly or indirectly [12]. As an extension of the current Web, the Semantic Web can add meaning to the World Wide Web content; and thus, it supports automated services based on semantic descriptions. The Semantic Web, meanwhile, relies on formal ontologies that structure the underlying data to enable a comprehensive and transportable understanding of machines [13]. Ontologies, as an essential part of the Semantic Web, are commonly used in information systems. Moreover, the proliferation of ontologies demands that ontology development should be derived quickly and efficiently to bring about the Semantic Web's success [12]. Nonetheless, manual ontology building is still a repetitive, cumbersome task, and the main problem in manually building ontologies is the bottleneck of knowledge acquisition and time-consuming development, maintenance complexity, and the integration of different ontologies for different applications and domains. To solve this issue, ontology learning is a solution to the bottleneck of acquiring knowledge and constructing ontologies on time. Ontology learning is defined as a subtask of extracting information, and its objective is to semi-automatically or automatically extract relevant concepts and relationships from a given corpus or other data sets to construct ontology. The current approach for automatic acquisition of concepts and concept hierarchies from a text corpus is introduced by [14]. This approach is based on formal concept analysis (FCA) to construct a concept lattice that can be converted to a particular type of partial order that constitutes a hierarchy of concepts. The overall process includes a set of steps to construct concept hierarchies from the text. One of the primary steps of this construction is how to extract the phrase dependencies or word pairs, particularly the pairs required and interested because not all the pairs are right or interesting. Hence, eliminating the erroneous and uninterested pairs is an issue that need to be addressed to reduce the size of formal context to consume less time for deriving the concept lattice. After that, the pairs are then weighted based on some statistical measures, and only the pairs over a certain threshold are converted into a formal context to which formal concept analysis is applied. Finally, the concept hierarchies are produced from the concept lattice.

## 1.3 Thesis Objectives

Despite the advances in using evolutionary computing for data clustering analysis and Semantic Web ontology learning, there are many opportunities for carrying out further research in this area. These opportunities can be focused on this thesis:

- Studying the performance of soft computing algorithms on different cohorts of datasets;
- The use of new meta-heuristic operators for clustering algorithm;
- Designing a novel and multi-disciplinary evolutionary clustering algorithm for heterogeneous datasets and real-world applications;
- Proposing a performance framework to measure the performance sensibility of evolutionary clustering algorithms sensitive towards heterogeneous datasets and practical applications;
- Proposing a framework to reduce the size and ambiguity of the formal context for deriving concept hierarchies from corpora using evolutionary clustering algorithms.

## 1.4 Thesis Contributions

The main contributions of this thesis are:

- An equitable and fair performance evaluation of the popular soft computing algorithms is conducted;
- An operational framework of BSA variants is proposed;
- A new ensemble and heterogenous clustering evolutionary algorithm is proposed.
  - It is based on BSA and several statistical and heuristic techniques,
  - It is used for multifeatured and heterogenous datasets
- An adaptive version of ECA* is applied on constructing concept hierarchies to reduce the size of concept lattice and produce reduced formal context.

## 1.5 Thesis Organisation

This thesis is divided into five chapters. As it has presented, the first chapter is mainly about an overview of the related topics in this thesis, problem statement and significance of the thesis, thesis objectives, thesis contributions and thesis organisation. The remainder of the thesis is organised as follows:

Chapter two is about the theoretical background and algorithms related to this thesis. Firstly, the standard BSA with its convergence analysis are introduced. Secondly, the background and theory of four principal fields of this thesis are presented: (i) Nature-inspired computing; (ii) Clustering in data mining; (ii) The Semantic Web ontology learning, including ontologies, ontology learning, input sources for ontology learning; (v) Information extraction and text mining, including natural language process, Stanford natural language processing framework, and wordnet.

Chapter three presents the literal work related to the previous works on analytical evaluation of BSA, evolutionary clustering algorithms and formal context reduction in deriving concept hierarchies in ontology learning from corpora.

Chapter four is divided into six sections. In the first section, an operational framework of the main expansions of BSA and its principal implementations is proposed. In the second section, the performance of BSA is evaluated against its competitive algorithms via conducting experiments. In the third section, a multi-disciplinary evolutionary algorithm called ECA* is proposed for clustering heterogeneous datasets, followed by experimenting to evaluate its performance compared with its counterpart techniques. In the fifth section, a framework is proposed using adaptive ECA* to reduce formal context size to derive concept hierarchies from corpora. Finally, an experiment is carried out to investigate the reduced formal context size.

In chapter five, the results are presented, then directly analysed and discussed for readability purposes. Firstly, the results related to performance evaluation of BSA is argued, followed by the limitations and contributions of BSA. Secondly, the evaluation methods of the clustering results of ECA*, performance evaluation of the same algorithm, and a proposed performance ranking framework of ECA* are presented. Finally, the proposed algorithm applied to practical applications such as concept hierarchies in ontology learning and also used to reduce the size of formal context to construct concept hierarchies from free texts.

Chapter six presents the concluding remarks for the work done and discusses possible directions for further work in the future.

# Chapter Two: Theoretical Background

Chapter two is about the theoretical background and algorithms related to this thesis. Firstly, the standard BSA with its convergence analysis are introduced. Secondly, the background and theory of four principal fields of this thesis are presented: (i) Nature-inspired computing; (ii) Clustering in data mining; (ii) The Semantic Web ontology learning, including ontologies, ontology learning, input sources for ontology learning; (v) Information extraction and text mining, including natural language process, Stanford natural language processing framework, and wordnet.

## 2.1 Backtracking Search Optimisation Algorithm

In recent years, BSA has gained significant attention owing to its various applications, especially in the fields of science and engineering. Backtracking was introduced in 1848 and was then known as '8-queens puzzle'. Further, Nauck created an 8*8 chessboard in 1950 for finding a feasible solution to this problem [15]. Later, Pinar Civicioglu introduced BSA as an EA for solving numerical optimisation problems [16]. Currently, BSA is one of the popular EAs used widely for real-valued numerical, non-linear, and non-differentiable optimisation problems. The naive idea of backtracking originates in an improvement of the brute force approach. Backtracking automatically searches for an optimal solution to a problem among all the available solutions. The available solutions are presented as list of vectors, and the optimal solution is reached by moving back and forth or sideways in the vector's domain. Algorithm (2.1) outlines the structure of BSA.

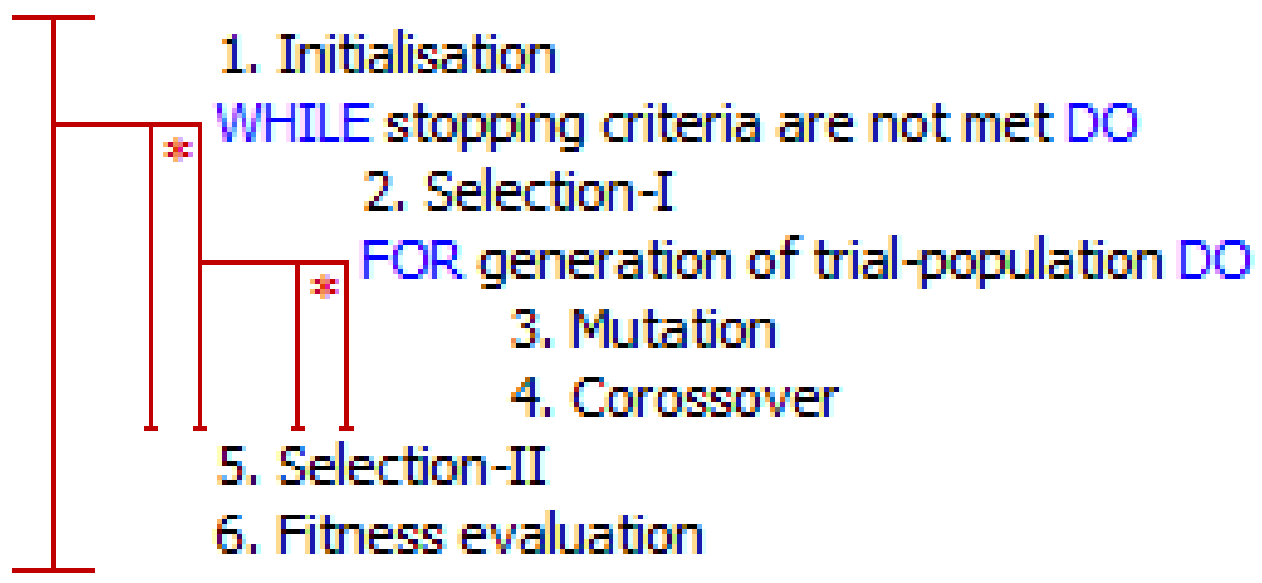

**Algorithm (2.1): The structure of BSA (adapted from** [16]**)**

Some websites related to BSA share the BSA source codes in different programming languages and the ideas, publications, and latest advances on BSA. Such websites are listed in Table (2.1).

**Table (2.1): Public websites on BSA**

| Website name | Reference, author(s) | URL |
|---|---|---|
| backtracking search optimisation algorithm (BSA) | [16], P. Civicioglu | www.pinarcivicioglu.com |
| Mathworks for backtracking search optimisation algorithm | [17], P. Civicioglu | www.mathworks.com |
| Backtracking tutorial using C program Code | [18], K. DUSKO | www.thegeekstuff.com |
| Applications of backtracking | [19], Huybers | www.huybers.net |

### 2.1.1 Standard BSA

BSA is a population-based EA that uses the DE mutation as a recombination-based strategy [20]. According to [16], [21]–[23], BSA comprises six steps: initialisation, selection-I, mutation, crossover, selection-II, and fitness evaluation. Figure (2.1) illustrates the flowchart of BSA. The steps in the flowchart are described in the following sub-sections [16].

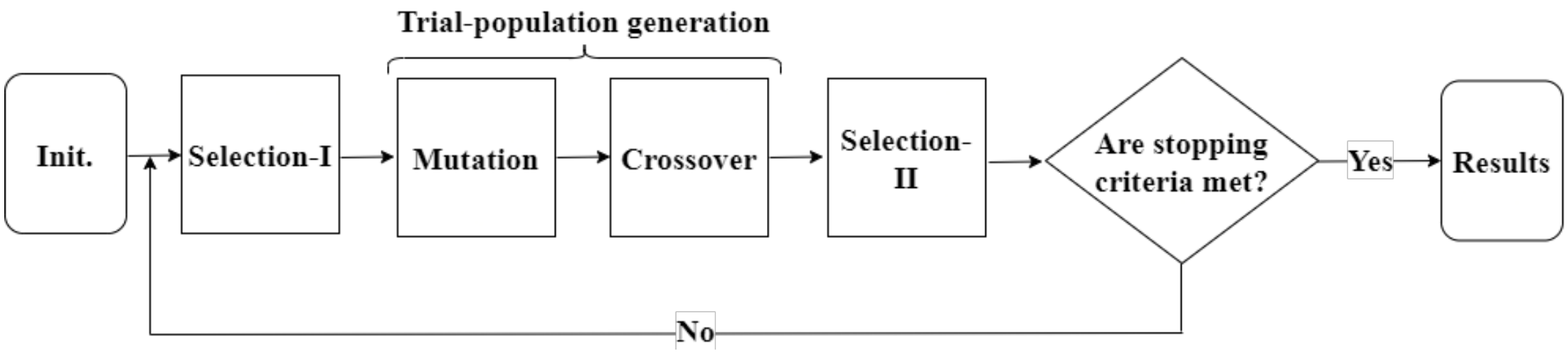

**Figure (2.1): The flowchart of BSA**

**A. Initialisation:** The initial population (*P*) in BSA is generated randomly, and it comprises D variables and N individuals. It is expressed by Equation (2.1) as follows:

$$P_{ij} \sim U(low_j,\ up_j) \tag{2.1}$$

For *i=1, 2,…., N and n=1, 2,…., D,* where *N* is the population size, *D* is the problem dimension, and *U* is the uniform distribution.

**B. Selection-I.** There are two selection operators in BSA: the pre-selection operator, selection-I, and the post-selection operator, selection-II. The pre-selection operator is used to obtain the historical population ($P^{old}$) that is further used to calculate the direction of the search. The value of $P^{old}$ is calculated by using the following three steps:

1. The initial historical population is generated randomly by using Equation (2.2) as follows:

$$P_{ij} \sim U(low_j,\ up_j) \quad (2.2)$$

2. At the beginning of each iteration, $P^{old}$ is redefined as by using Equation (2.3) as follows:

$$P_{ij}^{old} \sim U(low_j, up_j) \quad (2.3)$$

if $a < b$ then $P^{old} := P | a, b \sim U(0,1)$,

where: = is the update operation and $a$ and $b$ are random numbers.

3. After determining the historical population, the order of its individuals is changed by using Equation (2.4) as follows:

$$P^{old} := Permuting(P^{old}) \quad (2.4)$$

The permutation function is a random shuffling function.

**C. Mutation.** BSA generates the initial trial population (mutant) by applying the mutation operator as shown in Equation (2.5).

$$Mutant = P + F.(P^{old} - P) \quad (2.5)$$

where $F$ controls the amplitude of the search direction matrix (Pold - P). This thesis uses the value $F = 3 \,.\, rndn$, where $rndn \sim N(0, 1)$ , and $N$ is the standard normal distribution.

**D. Crossover.** BSA's crossover generates the final form of the trial population ($T$). It comprises two steps:

1. A binary-valued matrix (map) is generated, where the map size is N X D, which indicates the individuals of trail population T.

2. The initial value of the binary integer matrix is set to 1, where $n \in \{1,2,\ldots, N\}$ and $m \in \{1,2,\ldots, D\}$.

The value of $T$ is updated by using Equation (2.6) as follows:

$$T_{n,m} := P_{n,m} \quad (2.6)$$

**E. Selection-II.** Selection-II is called BSA greedy selection. The individuals of trail population T are replaced with the individuals in population $P$ when their fitness values are better than those of the individuals in population *P*. The global best solution is selected based on the overall best individual having the best fitness value.

F. **Fitness evaluation.** A set of individuals is evaluated by the fitness evaluation. The list of individuals is used as input, and fitness is considered as the output.

These aforementioned processes are repeated, except for the initialisation step until the stopping criteria are fulfilled. The detailed flowchart of BSA is illustrated in Figure (2.2) as a set of processes. A population of the individuals *(P)* is created randomly in the initialisation step. Further, individual historical populations (*oldP*) are created in the Selection-I step. The value of *oldP* gets updated in this step, and further, the individuals of *oldP* are randomly re-arranged. The value of *oldP* remains unchanged until further iterations. Next, a population of the initial trial *(mutant)* is created from *oldP* and *P* by a mutation process. Moreover, by using *P* and *Mutant*, a population of the final cross-trial (*T*) is created. Finally, *P* gets updated with *T* individuals in the Selection-II step by considering the best fitness and selecting the corresponding individual. The steps from Selection-I to Selection-II are iterated until the stopping criteria are fulfilled.

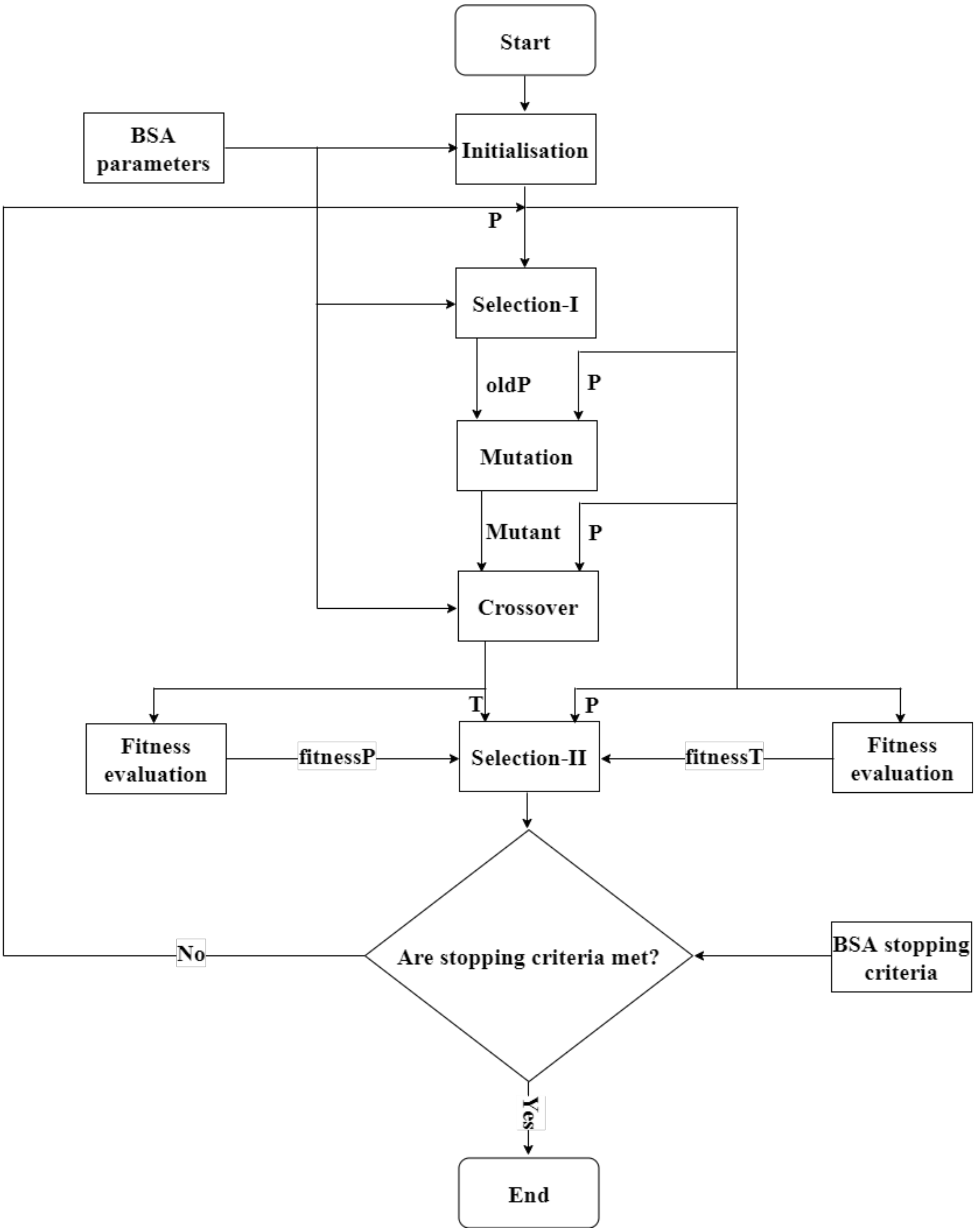

**Figure (2.2): Detailed flowchart of BSA (adapted from** [21]**)**

### 2.1.2 BSA Convergence Analysis

Currently, most researchers are focusing on the performance improvement of BSA. Convergence analysis is an essential tool to analyse the speed of an algorithm during its iterative process [24], [25]. This analysis process is usually represented as a convergence rate to find the desired solution. Therefore, the crossover and mutation parameters of BSA may differ owing to the dependence on the suitable solution of convergence. Additionally, it is recommended to adjust the scale factors within the mutation procedure [26]. This ensures that the convergence level and the variety of populations are approached to their balance. In this manner, the BSA performance is improved by optimisation through an iterative process.

BSA can be mathematically analysed in the context of initialisation, selection, mutation, and crossover. The initial step of BSA selects the current population, *P*, and historic population, *oldP*. This step is denoted by Equation (2.7) as follows:

$$P_{ij} = P_j^{min} + \left(P_j^{max} - P_j^{min}\right) * rand\ (0,1) \tag{2.7}$$

where, i = 1, 2, …, *N* and J = 1, 2, ……, *D*

To contribute more, *N* approaches the population size and *D* approaches the problem dimension. At this stage, the selection process chooses the historical population. It can be expressed by Equations (2.8), and (2.9) as follows:

$$oldP := \begin{cases} P & a,b \\ oldP & otherwise \end{cases} \tag{2.8}$$

$$oldP := permutting\ (oldP) \tag{2.9}$$

Further, the BSA mutation procedure creates a real sample vector. The search direction matrix, (*oldP* – *P*), is estimated and controlled by F that is randomly generated within a standard normal distribution range. The BSA mutation is expressed by Equation (2.10) as follows:

$$Mutant = P + F * (oldP - P) \tag{2.10}$$

Then, the BSA generates a heterogeneous and complicated crossover strategy. The crossover process creates the final version of the trial population. At the forefront, the matrix map of N * D binary integers is obtained. Moreover, depending on the BSA map value, the suitable size of the mutated individual is updated by using the corresponding individual in P. The crossover process can be expressed by Equation (2.11) as follows:

$$V_{ij} = \begin{cases} P_{ij} & map_{ij}=1 \\ Mutant_{ij} & otherwise \end{cases} \tag{2.11}$$

The rate of crossover for a modified scaling factor is expressed in Equation (2.12) [27]

$$F = F_{min} + (F_{max} - F_{min}) \frac{F_{i\,max} - F_{i\,min}}{F_{0\,max} - F_{0\,min}} \quad (2.12)$$

Therefore, the rate of convergence of BSA is expressed by minimising the sequence of the error term among the current and the historical solution. It is known that if the sequence approaches to zero 0, then it provides an insight into how speedily the convergence happens. The convergence analysis can be explained as the Marciniak and Kuczynski (M-K) method [28]–[31].

$$X_{n+1} = Z_n - \frac{f\,(Zn)}{f\,[Z_n, y_n\ ] + f\,[z_n, x_n, y_n](z_n - y_n)} \quad (2.13)$$

Assume that the error in $x_n$ is $e^n = x_{n-}$ α with the beneficial approach of Taylor expansion, $x =$ α. The convergence is six, and the order is seven

$$f_{(x_n)=f'}\,(\alpha)\,[e_n + c_2\,e_n^2 + c_3\,e_n^3 + c_4\,e_n^4 + c_5\,e_n^5 + c_6\,e_n^6 + c_7\,e_n^7 + O(e_n^8)] \quad (2.14)$$

In addition

$$f_{(x_n)=f'}\,(\alpha)\,[1 + 2\,c_2\,e_n^2 + 3\,c_3\,e_n^3 + 4\,c_4\,e_n^4 + 5\,c_5\,e_n^5 + 6\,c_6\,e_n^6 + 7\,c_7\,e_n^7 + O(e_n^7)] \quad (2.15)$$

whereas,

$$C_k = \frac{1}{k}\,!\,\frac{\left(f^k\,(\alpha)\right)}{f'\,(\alpha)}, k = 2, 3, \ldots.. \quad (2.16)$$

Therefore, the iterative method can be estimated by using the Marciniak-Kuczynski (M-K) method.

## 2.2 Two-sided Wilcoxon Singed-Rank

The Wilcoxon test is a nonparametric statistical test that compares two paired groups, and comes in two versions the Rank Sum test or the Signed Rank test. The goal of the test is to determine if two or more sets of pairs are different from one another in a statistically significant manner. In addition, the Wilcoxon Signed-Rank Test was used for pairwise comparisons, with the statistical significance value α = 0.05. The null hypothesis H0 for this test is: ‘There is no difference between the median of the solutions achieved by algorithm A and the median of the solutions obtained by algorithm B for same benchmark problem', i.e., median (A) = median (B). To determine whether algorithm A reached a statistically better solution than algorithm B, or if not, whether the alternative hypothesis was valid, the sizes of the ranks provided by the Wilcoxon Signed-Rank Test (i.e., R+ and R- as defined in [46]) were examined.

- Ranks = n (n+1)/2
- R+: Sum of positive ranks
- R-: Sum of negative ranks
- P-value: The P value answers this question: If the data were sampled from a population with a median equal to the hypothetical value you entered, what is the chance of randomly selecting N data points and finding a median as far (or further) from the hypothetical value as observed here?
- If the P value is small, you can reject the idea that the difference is a due to chance and conclude instead that the population has a median distinct from the hypothetical value you entered;
- If the P value is large, the data do not give you any reason to conclude that the population median differs from the hypothetical median. This is not the same as saying that the medians are the same. You just have no compelling evidence that they differ. If you have small samples, the Wilcoxon test has little power. In fact, if you have five or fewer values, the Wilcoxon test will always give a P value greater than 0.05, no matter how far the sample median is from the hypothetical median.

## 2.3 Nature-Inspired Computing

The term computational intelligence (CIn) typically refers to a computer's ability to learn from data or experimental observation of a particular function. While it is widely considered a synonym for soft computing, but there is still no commonly accepted definition of computational intelligence. Generally, computational intelligence is a collection of computer methods and solutions inspired by nature that solves complex and real problems that, for certain purposes, mathematical or conventional modelling may be unusable: systems may be too complicated for the use of mathematical logic, may involve some uncertainties in the processes or may merely be stochastic [32].

Therefore, CIn provides solutions to these issues. Figure (2.3) structures the classification of CI.

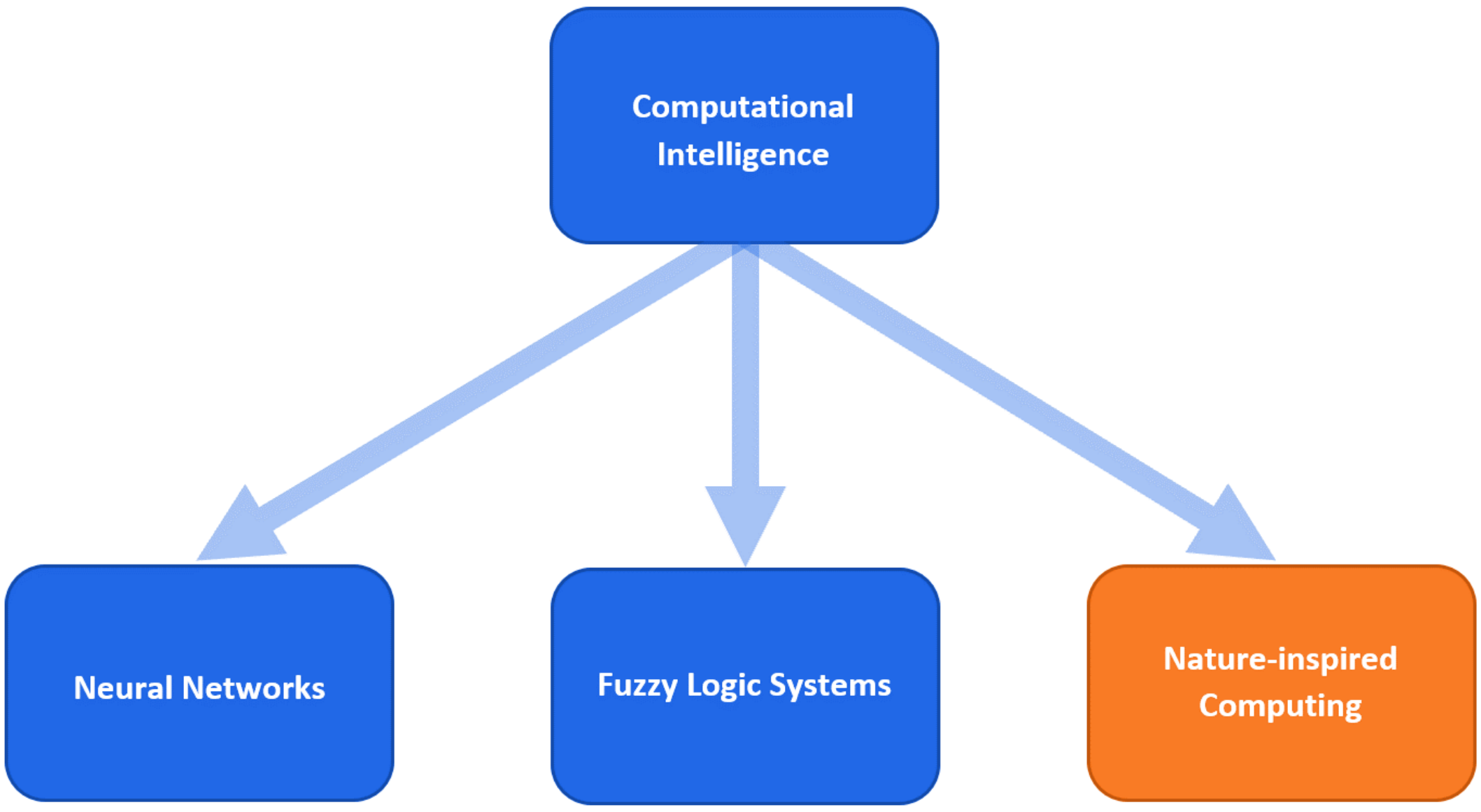

**Figure (2.3): The classification structure of computation intelligence**

The principal techniques of CIn are nature-inspired computing (NIC), fuzzy logic systems, and neural networks. Nature is a big part of science; it efficiently and effectively solves problems. This can be done through the transition of information from natural systems to computer systems. In this manner, NIC is a very modern discipline that aims to develop modern computational techniques by studying the natural actions of the phenomenon in various environmental conditions to solve complex problems. It has enabled ground-breaking research that has developed new branches, such as neural networks, swarm intelligence, computational computing, and artificial immune systems. In the fields of physics, biology, engineering, economy, and management, NIC techniques are used. NIC techniques, such as evolutionary computing and swarm optimisation algorithms have helped in solving complex and real-world problems and provide the optimum solution.

In NIC, besides the existing traditional symbolic and statistical methods [33], evolutionary and swarm optimisation algorithms as part of nature-inspired computing have emerged as a recent research area [34]. The classification of NIC is depicted in Figure (2.4). According to [32], nature-inspired computing deals with collections of nature-inspired algorithms (NIA) that, in turn, have been used to deal with complex practical problems, which cannot be feasibly or effectively solved by traditional methods. Thoroughly, NIA deals with artificial and natural systems that comprise several individuals and possess the ability of self-organisation and decentralised control. This concept, which was primarily initiated by Gerardo Beni and Jing Wang in 1989 in the context of cellular robotic systems, is used in artificial intelligence (AI) [35]. Typically, the NIA system comprises a collection of agents. The agents interact with their environment or with each other. NIA approaches often draw inspiration from the biological systems and nature [36], [37]. Consequently, the ideas of termite and ant colonies, fish schooling, birds flocking, evolution, and human genetics have been used in the EAs. Based on the bio-inspired operators, the idea of a genetic algorithm (GA) was proposed by Emanuel Falkenauer [38], whereas differential evolution (DE) inspiring by evolutionary development was invented by Storn and Price [39]. Furthermore, an algorithm inspiring by the ant foraging behaviour in the real-world is called an ant optimisation algorithm [40], whereas an algorithm based on bird flocking that is called particle swarm optimisation (PSO), which was introduced by Kennedy and Eberhart [41]. Moreover, several creatures have motivated NIA scholars to develop original optimisation algorithms.

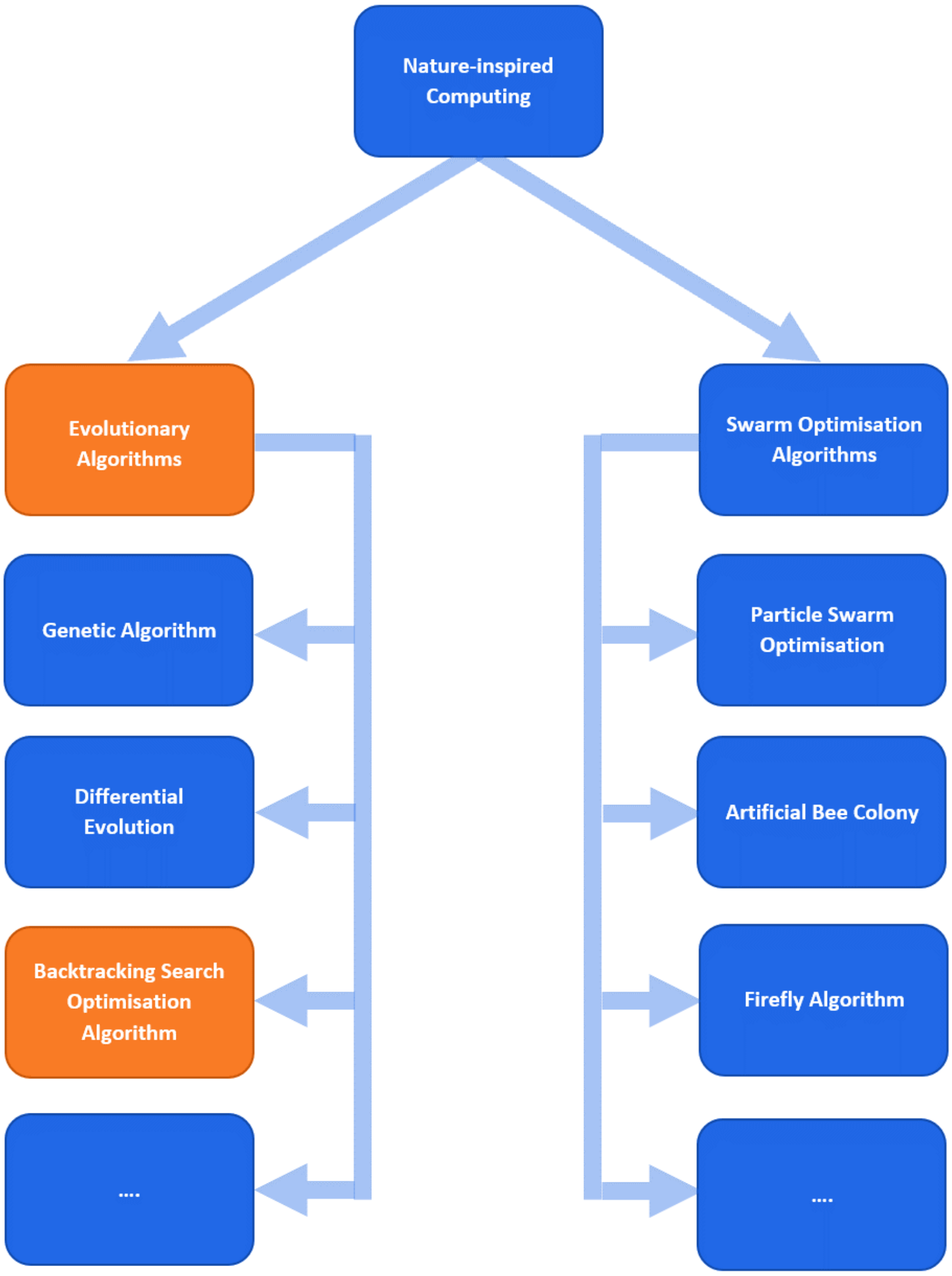

**Figure (2.4): Classification structure of nature-inspired computing (Adapted from** [42]**)**

Additionally, some human artifacts, such as multi-robot systems and some computer programs to solve data analysis and optimisation problems can also be considered to fall in the EAs domain. In this context, Pinar Civicioglu introduced BSA as an EA [16]. BSA is one of the recent evolutionary algorithms (EAs) used to solve and optimise numerical problems. It was developed in 2013, and since then, several scholars have cited and elaborated it in the EAs research community. Consequently, several research articles on the developments and applications of BSA are being published each year. Figure (2.5) depicts the number of research articles on BSA that have been published since 2013 to late 2019 on the platforms of Google Scholar, Web of Science, and Scopus databases. The Figure illustrates the cumulative growth of BSA citations in the literature from the last few years. The number of published articles citing the standard BSA reached a peak in 2016, followed by 2018 and 2017. Overall, the number of articles based on BSA has increased yearly. The increasing use of BSA has attracted more researchers to study its development, and different types of applications and problems have been solved effectively and efficiently by using this meta-heuristic algorithm.

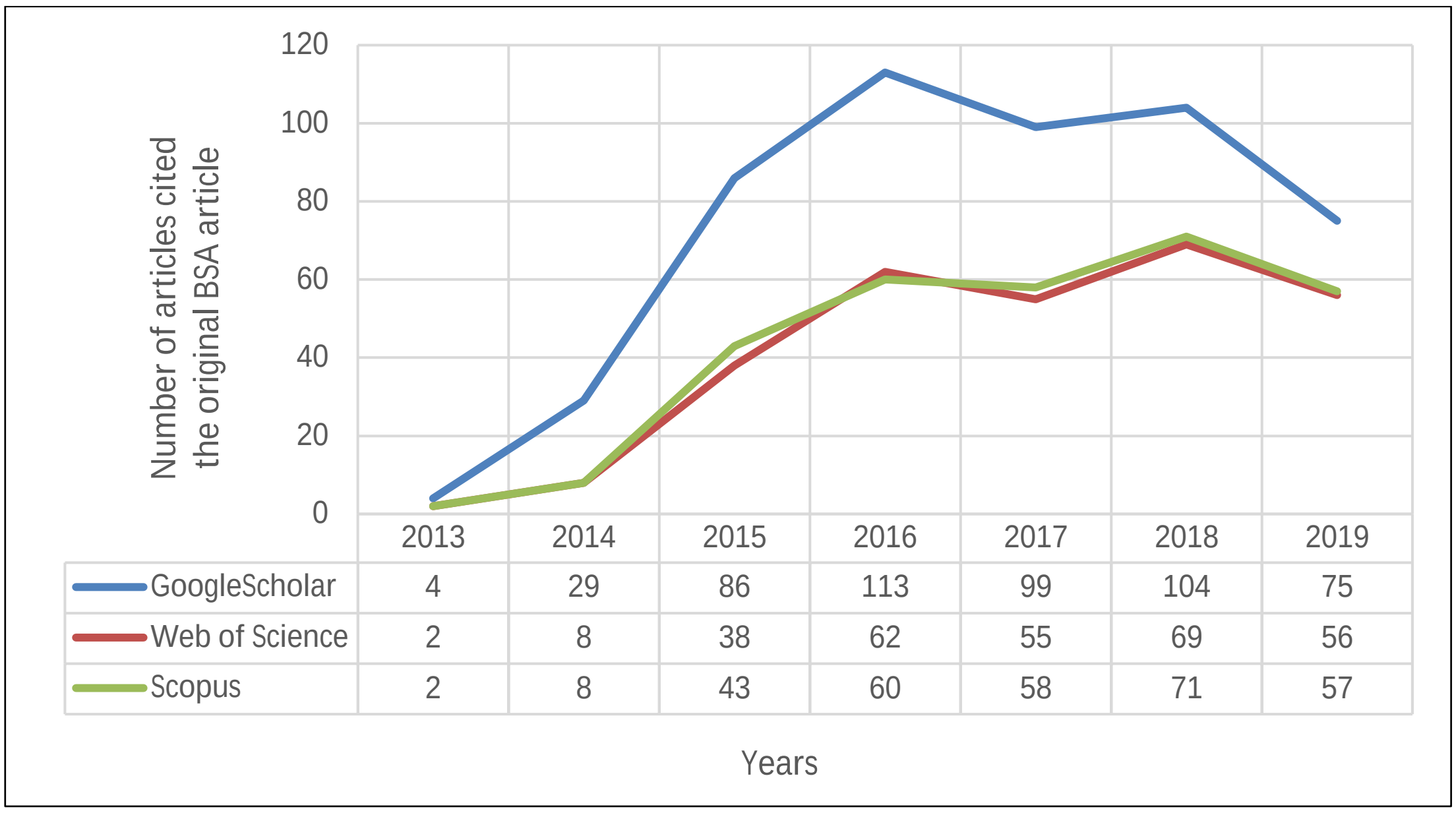

**Figure (2.5): Number of yearly published articles cited the standard BSA algorithm from 2013 to late 2019**

Owing to the adaptability of BSA to different applications and optimisation problems, several scholars have proposed new algorithms based on the original BSA, whereas few others have attempted to employ the original BSA in different applications to solve a variety of problems. Numerous research articles are focusing on the advancements of BSA such as modifications, hybridisation, and extensions. The number of published articles on the modifications of BSA is considerably higher than those on hybridising the BSA with other techniques. BSA was primarily used to deal with numerical optimisation problems; however, currently, it has been extended to deal with multi-objective, constrained, and binary optimisation problems. Currently, several published articles on extending the BSA for different optimisation problems are available. Although BSA was proposed to solve numerical optimisation problems, and it has also tackled several other optimisation problems. The problems tackled by BSA can be categorised into three types: the fundamental problems handled by this algorithm (n-queens problem, knight's tour puzzle, random word generator for the Ambrosia game, and maze generation) [15], [43]–[46], travelling of salesperson (TSP) [47], [48], and scheduling program [49], [50].

Although BSA has been recently developed, it has already been used in several real-world applications and academic fields. The latest applications and implementation categories of BSA are control and robotics engineering, electrical and mechanical engineering, AI, information and communication engineering, and material and civil engineering. Additionally, it has been intensively used in the area of electrical and mechanical engineering. However, few articles have been published on BSA regarding the applications of AI, material and civil engineering, and control and robotics engineering, in decreasing order. It has also been affirmed by [10] that BSA is one of the most used EAs in the field of engineering. On this basis, BSA is considered as one of the most popular and widely used evolutionary optimisation algorithms. Furthermore, the performance evaluation of BSA is an interesting research area. However, owing to several reasons, a fair and equitable performance analysis of BSA is a challenging task. These reasons can be selecting initialisation parameters of the competitive algorithms, system performance during the evaluation, programming style used in the algorithms, the balance of randomisation, and the cohorts and hardness scores of the selected benchmark problems. In this thesis, an experiment was conducted to fairly compare BSA with the other algorithms: PSO, artificial bee colony (ABC), firefly algorithm (FF), and DE by considering all the aforementioned aspects, owing to the lack of equitable experimental evaluation studies on BSA in the literature. Consequently, the uses and performance evaluation of BSA were focused on providing a baseline of the studies, and the recent development of BSA was also concentrated to find footprints to guide scholars for conducting more studies in the future.

## 2.4 Clustering in Data Mining

Clustering groups the observations of a dataset in a way with similar observations are grouped together in a cluster, and dissimilar observations are positioned in different clusters. Clustering algorithms have a wide range of applications in nearly all fields, including ontology learning, medical imaging, gene analysis, and market analysis. Over the past decades, several clustering techniques have been proposed, but each of these algorithms was mostly dedicated to a specific type of problem [11]. For instance, [51] concluded that K-means works poorly for the datasets that have a significant number of clusters, cluster size, or un-balance among the clusters. Not every clustering algorithm needs to perform well in all or almost cohort of datasets and real-world applications, but it is instead a need to show how sensitive the algorithm towards different types of benchmarking and real-world problems. Despite that, almost all the clustering techniques have some common limitations: (i) Finding the right number of clusters is a challenging task [52]. (ii) The sensitivity of current clustering algorithms towards a random selection of cluster centroids is another constraint of the clustering techniques since selecting bad cluster centroids could easily lead to inadequate clustering solutions [53]. (iii) Since almost all the clustering techniques are the process of hill-climbing towards its objective function, they could easily be stuck in local optimum, and this may lead to poor clustering results [54]. (iv) Data is not empty of outliers and noise. Nonetheless, it is assumed that clusters have a similar spread and roughly equal density. As a consequence, noise and outliers may misguide the clustering techniques to produce an excellent clustering result. (v) There are a minimal number of studies that show the sensitivity of the clustering algorithms towards the cohorts of datasets and real-world applications. (vi) Most of the techniques and algorithms define a cluster in terms of the value of the attributes, density, distance, etc. Nevertheless, these definitions might not give a clear clustering result [55]. (vii) Moreover, most of the prior algorithms are deterministic [11], in which their clustering results are entirely related to their initial states and inputs. The given starting conditions and initial input parameters may directly affect the generation of output. This may achieve a balance between exploitation and exploration search spaces and cannot easily trap into local and global optimum accordingly [22].

Therefore, there is an enormous demand for developing a new clustering algorithm that is free from the shortcomings of current clustering techniques. Due to integrating several techniques in our proposed algorithm, the algorithm is called ECA*. The integrated techniques of ECA* are social class ranking, quartiles, percentiles, and percentile ranking, the operators of evolutionary algorithms, the involvement of randomness, and the Euclidean distance in the K-means algorithm. Moreover, the operators of evolutionary algorithms are crossover, mutation, a recombination strategy of crossover and mutation, called mut-over, in BSA, and random walk in the levy flight optimisation (LFO).

One important ingredient of ECA* is the use of two evolutionary algorithm operators: (1) BSA is one of the popular meta-heuristic algorithms. Apart from the initialisation and fitness evaluation component of BSA, it consists of four ingredients:: Selection I, Mutation, Crossover, Selection II [22]. The most important operators of BSA are mutation and crossover since these operators are acting as a recombination-based strategy. Crossover and mutation operators are also used in genetic algorithm and differential evolution algorithm. The reason for using BSA's operators in this study are two-fold [22], [23]: (i) BSA is a popular meta-heuristic algorithm that uses the DE crossover and mutation operators as a recombination-based strategy. (ii) This recombination helps BSA to outperform its counterpart algorithms. Additionally, BSA gets a balance between local and global searches to avoid trapping into local and global optima. This strategy of BSA provides robustness and stability to our proposed technique. (2) Levy flight optimisation (LFO) is a basic technique that uses several particles in each generation. From a better-known spot, the algorithm can create a whole new generation at distances randomly distributed by levy flights. Then, the new generation is tested to choose the most promising. The process is repeated until the best solution is selected, or the stopping conditions have been met. The implementation of LFO is straightforward [56]. The pseudocode algorithm based on the idea of LFO is detailed from [56].

Another significant ingredient of this algorithm is the utilisation of social class ranking. Each cluster is considered as social class rank and represented by percentile rank in the dataset. Table (2.2) represents an instance of a class name, cluster representation, and percentile rank boundary for each cluster in the proposed algorithm. This notion leads us to design the proposed algorithm based on an adaptive number of clusters.

**Table (2.2): A sample representation of the social class ranking**

| **Class name** | **Cluster representation** | **Percentile rank** |
|---|---|---|
| Upper cluster (Elite) | Cluster1 | 80-100 |
| Middle Cluster (Upper) | Cluster2 | 60-80 |
| Middle cluster (Lower) | Cluster3 | 40-60 |
| Working-cluster | Cluster4 | 20-40 |
| Poor cluster | Cluster5 | 0-20 |

There are several motives behind proposing ECA*. First and foremost, suggesting a dynamic clustering strategy to determine the right number of clusters is a remarkable achievement in clustering techniques. This strategy removes empty clusters, merging less dynamic cluster density with its neighbouring clustering, and merging two clusters based on their distance measures. Another reason is to find the right cluster centroids for the clustering results. Outliers and noise are always part of data, and this sometimes misleads the clustering results. To impede this misguidance, quartile boundaries and evolutionary operators can be exploited. These methods are useful to adjust the clustering centroids towards its objective function and to result in excellent clustering centroids accordingly. Additionally, developing a new clustering algorithm that does not trap into local and global optima is an incentive. In turn, a boundary control mechanism can be defined as a very effective technique in gaining diversity and ensuring efficient searches for finding the best cluster centroids. It is also concluded that having statistical techniques and involvement of randomness in the clustering techniques could help to generate meaningful clusters in a set of numerical data [55]. On this premise, developing a stochastic clustering algorithm is a need to seek a balance between exploration and exploitation search spaces and produce an excellent clustering result accordingly. Last but not least, defining an approach for clustering multi-feature and different types of problems is another rationale behind introducing ECA* since no clustering algorithm could have a good result on multi-feature datasets [51].

Clustering can be in two primary forms: (i) in the presence of cluster labels, the clustering technique is called supervised analysis. (ii) Unsupervised clustering techniques do not require cluster labels in datasets as a target pattern for the output. This form of clustering is the focus of this study. In addition to introducing these two primary forms of clustering techniques, a hybrid clustering technique was proposed as a combination of supervised and unsupervised techniques to provide compelling prediction performances in a case of unbalanced datasets [57]. The critical contribution of ECA* is as follows: At first glance, this study integrates the notion of social class ranking, quartiles, and percentiles, recombined operators of meta-heuristic algorithms, randomness process, and the Euclidean distance in K-means algorithm as combinational techniques of our introduced technique Section three demonstrates an analysis to present the advantages of the above techniques in this study's proposed algorithm. One of the essential benefits of ECA* is to find the right number of clusters. In almost all the current algorithms, the number of clusters should be previously defined. In contrast to the prior clustering algorithms, the most crucial contribution of ECA* is its adaptivity in determining the number of clusters. ECA* has an adaptive number of clusters since it is based on social class rankings. The social rankings initially set the number of clusters, and then the empty clusters will be removed. Moreover, the less-density clusters are merged with its neighbouring clusters. Further, selecting the right cluster centroids is one of the contributions of ECA*. Three steps are used for computing cluster centroids as follows: (i) Initially, cluster centroids are calculated based on the mean quartile of the clusters. (ii) Historical cluster centroids are selected between the lower and upper quartile to avoid the misleading of the algorithm by outliers and noise. (iii) Operators of meta-heuristic algorithms are used to adjust the cluster centroids to produce an excellent clustering result. Another value of ECA* is the involvement of stochastic and randomness processes. The advantage of the stochastic process in ECA* is to seek a balance between the search space exploration and utilising the learning process in the search space to hone in on the local and global optimum. Furthermore, the excellent performance of ECA* is contributed by using the operators of meta-heuristic algorithms produces [23]. Meanwhile, mutation strategy of this study's algorithm is improved in two ways [58]. Firstly, an adaptive control parameter (F) is introduced by using LFO to balance the exploitation and the exploration of the algorithm. Secondly, the cluster centroids learn knowledge from the historical cluster centroids (HI) to increase the learning ability of the cluster centroids and find the best cluster centroids for clusters accordingly. Mut-over could also provide a recombination strategy of mutation and crossover. Plus, mut-over with the aid of F and HI could overcome the global and local optima issue that may occur in the other algorithms, such as K-means [59]. This recombined evolutionary operator also provides stability and robustness to this study's proposed algorithm [58]. Hence, these strategies correctly balance the relationship between global optima and local optima. Finally, another contribution of ECA* is that it could handle scalable and multi-feature datasets with metric attributes, such as two dimensional and multi-dimensional datasets with a different number of clusters, as well as overlapped, well-structured and unstructured, and dimensional clusters. The components of ECA* are depicted in Figure (2.6) as a basic flowchart.

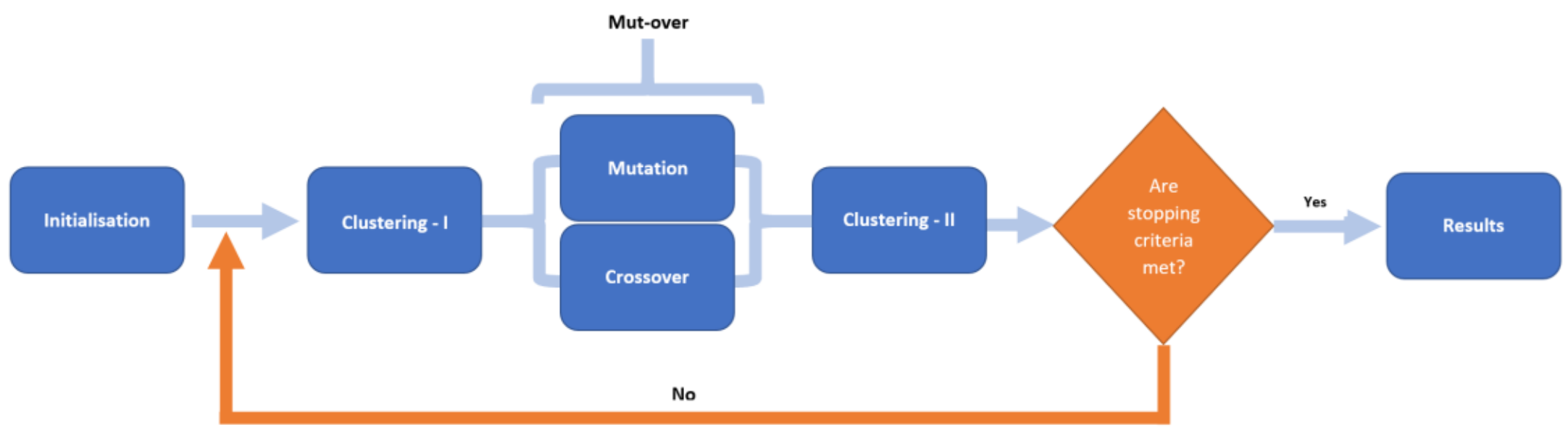

**Figure (2.6): The basic flowchart of ECA***

## 2.5 The Semantic Web Ontology Learning

This section consists of three sub-sections: ontologies, ontology learning, and input sources for ontology learning.

### 2.5.1 Ontologies

Ontology is a body of formally represented knowledge and a specification of conceptualisation as a set of concepts within a domain. In other words, it could represent objects, concepts, and other entities that exist in an area of interest, and the relationship among them to model domain support reasoning about concepts. From a philosophical perspective, on the other hand, ontology means the representation of what exists in the world. There are two common reasons for developing ontology information systems [60], [61]. The first reason is to establish a shared understanding of a domain and facilitate sharing information accordingly. The second reason is to enable the construction of intelligent or semantic applications. The latter reason can be seen in the context of the Semantic Web where is necessarily used for representing domain knowledge that allows machines to perform reasoning tasks on documents. Further, Ontology consists of concepts, taxonomic relationships that define a concept hierarchy, and non-taxonomic relationships between them, and concept instances and assertions between them. The ontology should be machine-readable and formal. In that sense, ontology could provide a common vocabulary between different applications. This knowledge representation structure often composed of a set of frame-based classes that hierarchically organised to describe a domain and provides a skeleton of the knowledge base. The represented concepts for describing a dynamic reality are used by the knowledge base and it changes by the changes in reality. Nevertheless, if there are changes in only the described domain, the ontology changes. Formally, ontology can be defined as a tuple of ontology concepts, a set of taxonomic and non-taxonomic relationships, a set of axioms, and instances of concepts and relationships [7].

### 2.5.2 Ontology Learning

Ontology learning refers to the semi-automatic or automatic support for constructing an ontology when the semi-automatic and automatic support for the instantiation of a given ontology is referred to as the ontology population [62]. Ontology learning concerns with knowledge discovery in different types of data sources and with its representation via an approach for automating or semi-automating the knowledge acquisition process. Whenever an author writes a text, he writes it by following a domain model in his mind. He or she knows the definitions of various concepts in a specific domain and then passes this domain information implicitly and explicitly in the text using this model. That means ontology learning is a reverse engineering process, as the domain model is reconstructed using the formal structure saved in the mind of the author [63]. Figure (2.7) summarises the steps possible to achieve unstructured text ontology.

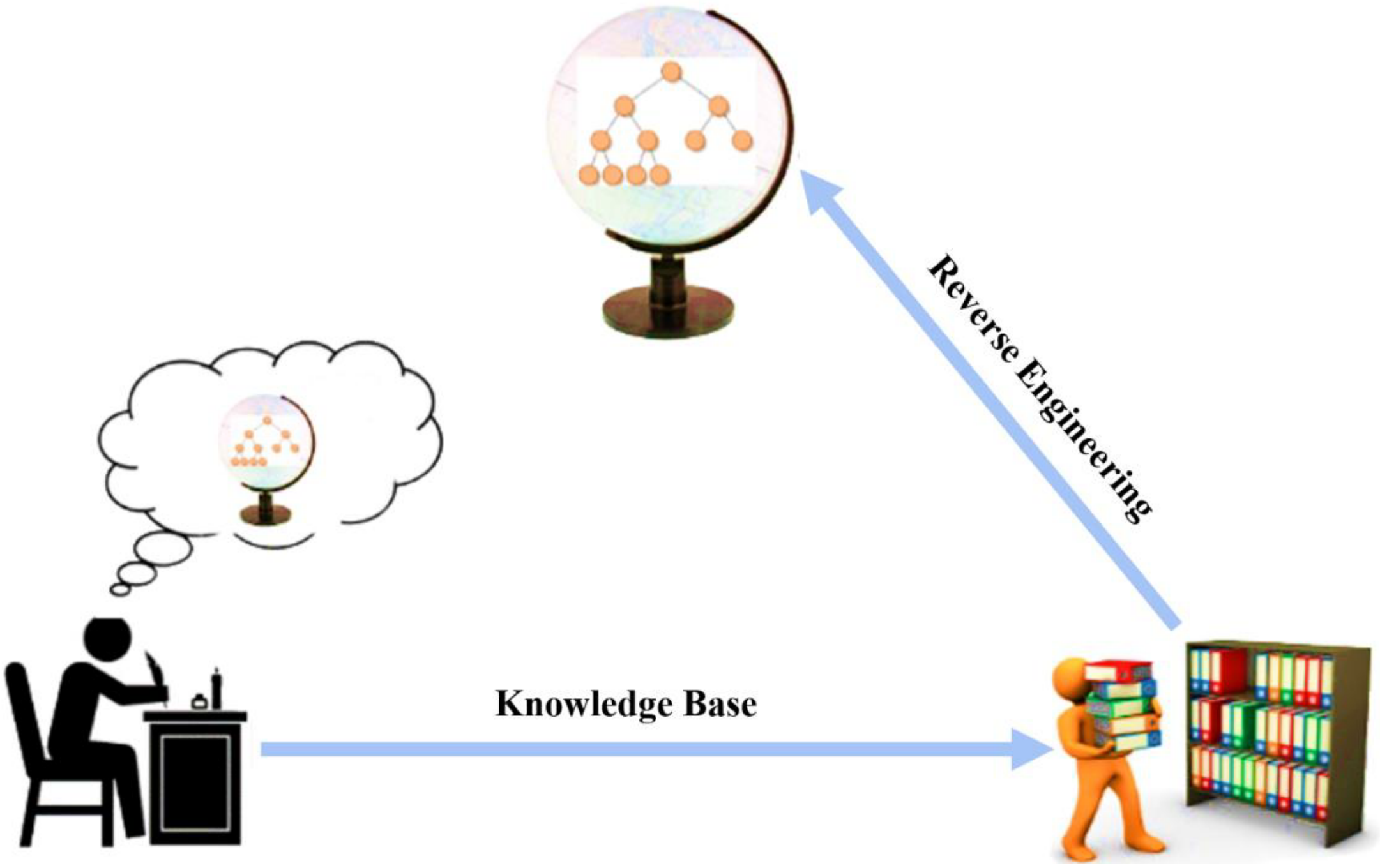

**Figure (2.7): Ontology learning from corpus as a reverse engineering process (reprinted from [64])**

Further, ontology learning development consists of a set of layers [64], [65] Such layers are shown in Figure (2.8). Ontology learning layer cake consists of concepts, relationships between them, and axioms. It is necessary to identify the natural language terms that refer to them to identify the concepts of a domain. This is particularly essential for ontology learning from a domain of the free text. Similarly, identifying Synonym helps to avoid redundant concepts when two or more natural language terms can represent the same concept. In the future, this identification helps to uniquely identify their respective concepts. Identifying the taxonomic relationships (generalisation and specialisation) between the concepts is the next step. Nontaxonomic relations are also necessary to extract.

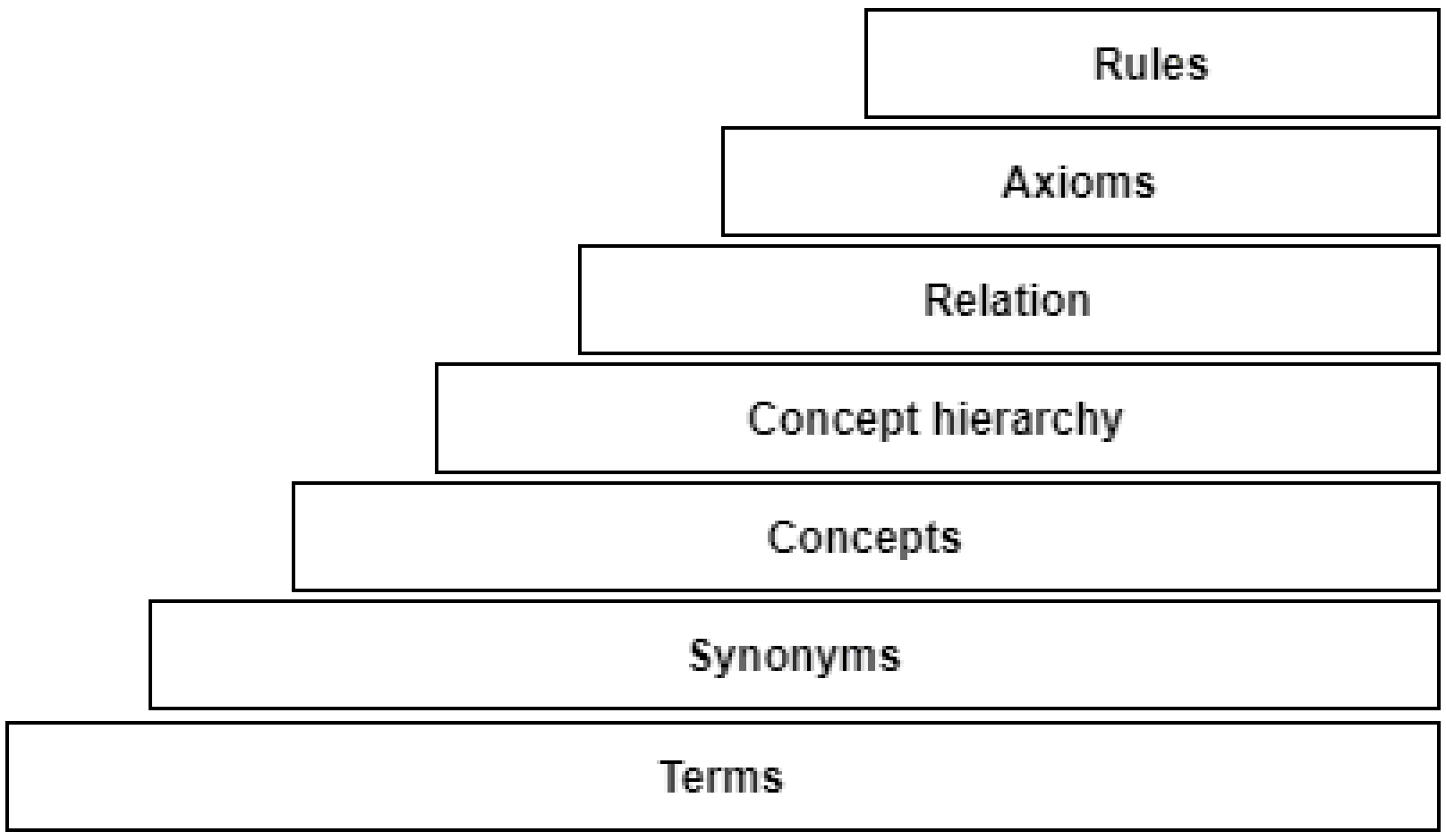

**Figure (2.8): Ontology learning layer cake (adapted from** [64], [65]**)**

These layers and their related concepts are detailed below:

1. **Terms:** Terms are a pre-requisite for all aspects of ontology learning from text and they can be defined as linguistic realisations of domain-specific concepts and thereby central to further complex tasks.
2. **Synonyms:** Synonyms addresses the acquisition of semantic term. The use of synonyms is provided by WordNet and related lexical resources.
3. **Concepts:** Concepts can be anything about which something is said [66]. They can be concrete or abstract, composite or elementary, fictitious or real, or the description of a task, action, function, strategy, or a reasoning process. Concepts are represented by nodes in the ontology graphs.
4. **Conceptual Relations:** Relations can be studies in two ways[13], [66]. The first way is a node in the ontology that is a concept and maybe learned as the other concepts. The second way is to relate two or more concepts and thereby it should be learned as a subset of a product of n concepts (for $n > 1$). In other words, relations may be learned intentionally independent of the relations of concepts or extensionally considering relations between concepts. For example, the part-of binary relation is a concept under the super concept "relation" and has its features and may be related to other concepts by some other relations as well. Alternatively, it relates the concepts, for example, "door" and "house" or "hand" and "human" can be shown as (part-of door house) or (part-of door house). Besides, the conceptual hierarchy may be non-taxonomic or taxonomic relations. Non-taxonomic relations are any relations between concepts except IS-A relations, such as synonymy, antonymy, meronymy, attribute-of, possession, causality and other relations. On the other hand, taxonomic relations are used for organising ontological knowledge by using generalisation or specialisation relationships through which simple or multiple inheritances can be applied.

5. **Axioms:** Axioms are used to model sentences that are always true [13], [66]. In ontology, they aim to constrain the information contained in the ontology, deducing new information, or verifying its correctness.
6. **Rules:** Despite learning ontological knowledge by systems, several systems may learn how to extract and learn ontological knowledge. Some systems may learn met-knowledge, such as rules for extracting relations, instances, and specific fields from the Web.

### 2.5.3 Input Sources for Ontology Learning

Input resources mainly can be of three types: unstructured, semi-structured, and structured. There are also different types of information retrieval and extraction, such as Natural Language Processing techniques, clustering, machine learning, semantics, morphological or lexical or combination of them. The learning process can be performed from scratch or some prior knowledge can be used. The first aspect is the availability of prior knowledge. Prior knowledge is used in the construction of the first version of ontology and such prior knowledge can demand little effort to be transformed into the first version of the ontology. Then, this version can be extended automatically via learning procedures and by a knowledge engineer manually. The second aspect is about the type of input used by the ontology process. The three different types of input sources are detailed below [13]:

1. **Ontology Learning from Structured Data:** These ontology learning procedures extract the parts of the ontology using the available structured information. database schemas, existing ontologies, and knowledge base are examples of structured information sources. The main issue in learning ontology with structured information sources is to determine which pieces of structured information can provide suitable and relevant knowledge [67]. For example, a database schema might be exploited for identifying concepts and their relationships as well.
2. **Ontology Learning from Semi-structured Data:** The resulting quality of ontology learning procedures using structural information is usually better than the ones using unstructured input information completely. Dictionaries are an example of semi-structured data, such as WordNet. WordNet is one of the manually compiled electronic dictionaries that is not restricted to any specific domain and covers most English nouns, verbs, adjectives, and adverbs [68]. WordNet is the most successful product among its similar products because it offers researchers an ideal, many of which were not initially envisaged by the authors, for disambiguation of meaning, information retrieval, and semantic tagging. Along with this, it is a well-documented open source and cost-free.
3. **Ontology Learning from Unstructured Data:** Learning ontology from unstructured data are those methods that do not rely upon structural information rather than unstructured data. Unstructured data is an important source of data for learning ontology because unstructured data is the most available format for ontology learning input. Unstructured data composes of natural texts, such as PDF and Word documents, or webpages.

## 2.6 Information Extracting and Text Mining

Several methods from a diverse spectrum of fields are used by ontology learning [64], [66], such as artificial intelligence and machine learning, knowledge acquisition, natural language processing, information retrieval, and reasoning and database management. Information extraction is to extract structured information from unstructured data, whereas text mining is related to information extraction as it is used to find relevant patterns and trends. Text mining usually involves the process of deriving linguistic features from the text and the removal of unimportant words for mining. The categorisation of text is a typical task of text mining.

### 2.6.1 Natural Language Processing

The main goal of research about Natural Language Processing (NLP) is to understand and parse language. Natural Language Processing is a subfield of linguistics that consists of an understanding of natural human language and automatic generation. Natural Language Processing systems analyse text to find group words and phrase structure by their semantic and syntactic type. Text mining and information extraction are usually dependent on NLP. For instance, text mining could be used in finding relations between nouns in a text. As a result, text is pre-processed by a part-of-speech tagger (POS) that groups the words in grammatical types. The most important components of Natural Language Processing that are used in this thesis are tokenisation, sentence splitter, POS, and lemmatisation.

### 2.6.2 Stanford Natural Language Processing Framework

The Stanford Natural Language Processing Group is a team of faculty, postdocs, research scientists, students, and programmers who work together on algorithms that allows the computer to process and understand human languages, such as English and Kurdish languages. Stanford Natural Language Processing tools range from basic research in computational linguistics to key applications in human language technology and covers areas, such as sentence understanding, probabilistic parsing and tagging, machine translation, biomedical information extraction, word sense disambiguation, grammar induction, and automatic question answering. One of the sets of natural language analysis tools of this group is the Stanford Natural Language Processing framework. It is an integrated framework and a suite of open-source and natural language processing tools for the English language that is written in Java [69]. Besides, the Stanford Natural Language Processing framework provides a set of natural language analysis tools that can take free English language text as an input and give the based forms of words and their parts of speech. Furthermore, it includes tokenisation, part-of-speech tagging, named entity recognition, coreference, and parsing. On the other hand, the Stanford Natural Language Processing framework API is used by Annotation and Annotator classes. Both of them form the backbone of the Stanford Natural Language Processing package. Annotations are the data structure that can hold the result values of annotations, which are maps, such as part-of-speech, parse, or named entity tags. On the other hand, Annotators are relatively like functions that they do things, such as parse and tokenise, but they operate over Annotations instead of Objects. Both Annotations and Annotators are integrated by Annotation Pipelines that create sequences of generic Annotators. Stanford Natural Language Processing framework inherits from the Annotation Pipeline class and is customised with Natural Language Processing Annotators. Figure (2.9) depicts the overall framework architecture: the raw text is put into an Annotation object and then a sequence of Annotators adds information in an analysis pipeline. The resulting Annotation, containing all the analysis information added by the Annotators, can be output in XML or plain text forms.

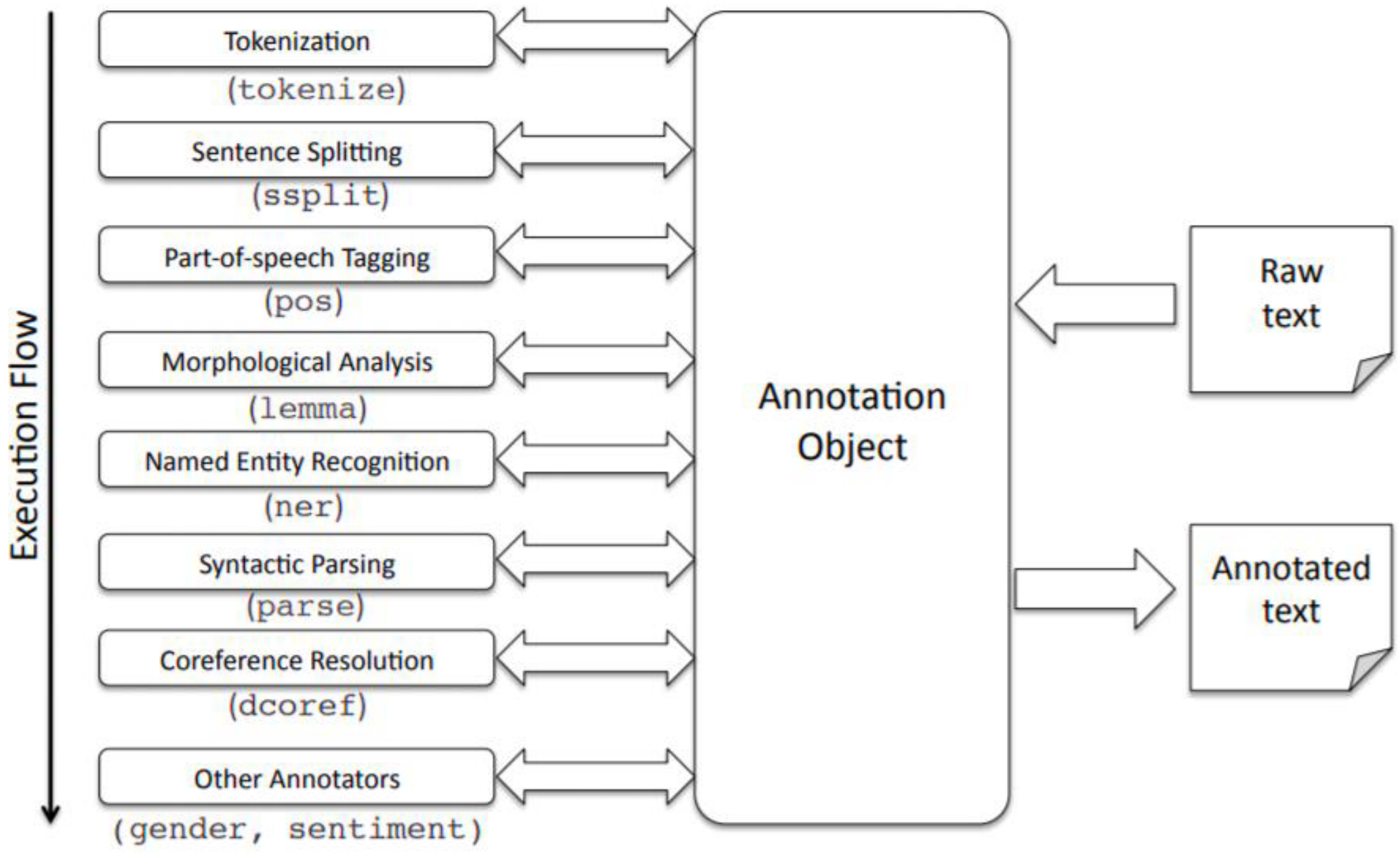

**Figure (2.9): The architecture of the Stanford CoreNLP framework (reprinted from** [69]**)**

### 2.6.3 WordNet

WordNet is a lexical electronic database that is considered to be the most important resource available to researchers in computational linguistics [70][71]. The purpose of WordNet is to support automatic text analysis and artificial intelligence applications. It is also known as one of the most used manually compiled electronic dictionaries that are not restricted to any specific domain and cover almost all English verbs, nouns, adjectives, and adverbs. Even though there are similar products, such as CYC, Cycorp, and Roget's International Thesaurus, WordNet is the most successful and growing one and it has been used in several applications over the last ten years [68].

Furthermore, both verbs and nouns are structured into a hierarchy which is defined by hypernym or IS-A relationships [71]. For example, the first sense of the word artifact has the following hyponym hierarchy as shown in Figure (2.10). The words at the same level are synonyms of each other. Moreover, some sense of cat is synonymous with some other senses of domestic cat and wildcats and so on. Besides, each set of synonyms or synset has a unique index which shares its properties, such as a dictionary or gloss definition.

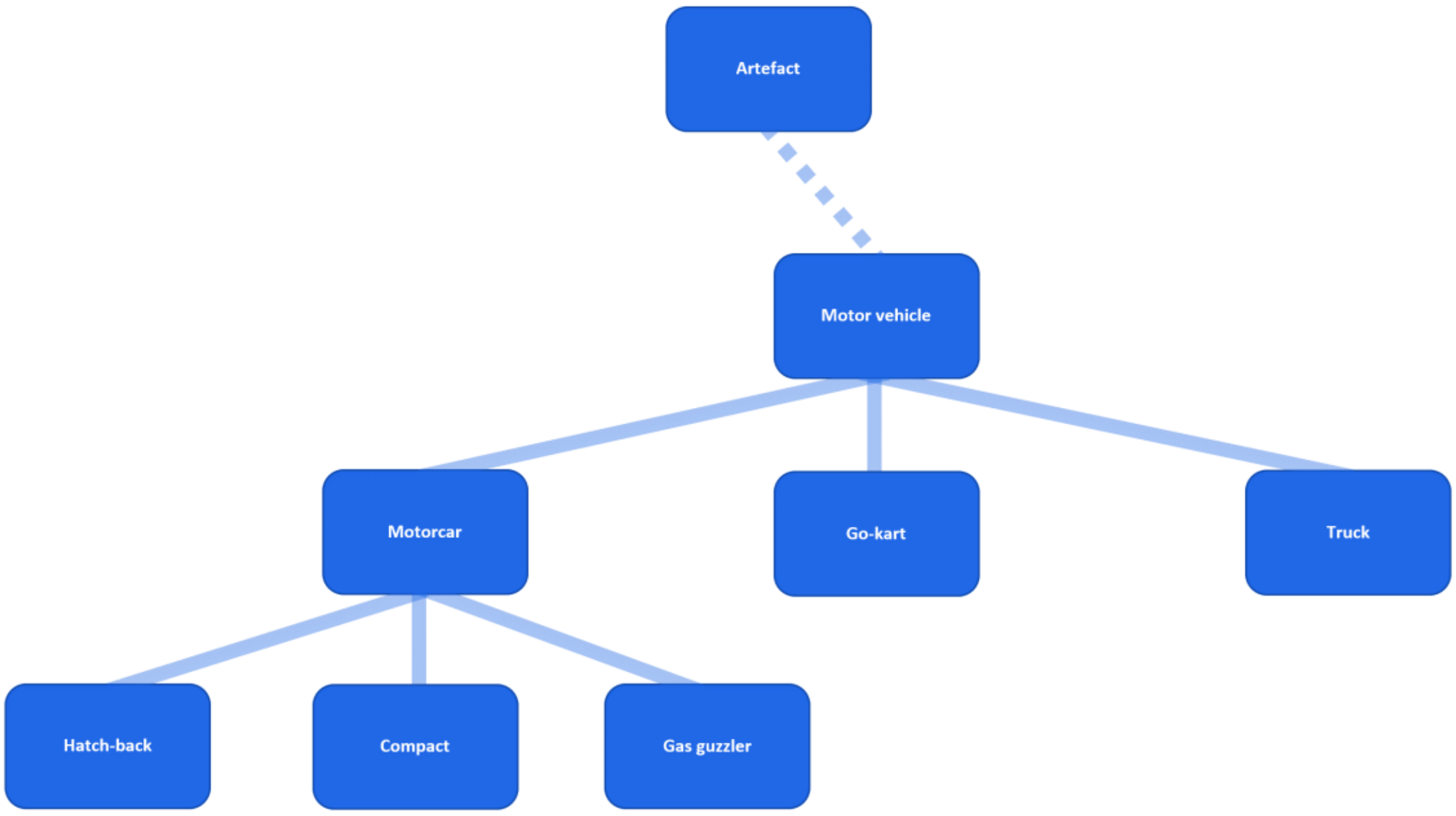

**Figure (2.10): A Sample of WordNet knowledge structure**

# Chapter Three: Literature Review

In this chapter, three related domains related to the thesis are reviewed:

- Analytical evaluation of BSA in the literature.
- Related work on evolutionary clustering algorithms.
- Related work on formal context reduction.

## 3.1 Analytical Evaluation of BSA in the Literature

This section presents the experiments and statistical analysis carried out in the literature on BSA by using the benchmark functions. Additionally, the performance of BSA in the previous studies is presented.

### 3.1.1 Statistical Analysis

In this analysis, 75 benchmark functions were used in three tests to examine the success of BSA in comparison with the other heuristic algorithms [16]. Test 1 uses 50 benchmark test functions. The details of these benchmark test functions are provided in [72], [73]. Furthermore, 25 benchmark functions are used in Test 2. The details of these benchmark functions are provided in [16]. Additionally, three benchmark test functions are used in Test 3. The details of these benchmark functions are provided in [74]. In the literature, the Wilcoxon signed-rank test was used to determine which two algorithms achieved a better solution in terms of solving numerical problems statistically. In the same study [16], necessary statistics measures such as standard deviation, best, mean, and runtime were achieved by BSA, PSO, ABC, CMAES, JDE, SADE, and CLPSO in 75 mathematical benchmark problems for 30 solutions in 3 tests, considering the statistical significant value (α) as 0.05 and the null hypothesis (H0). Tests 1, 2, and 3 were regarding the basic statistical measures (standard deviation, best, mean, and runtime) achieved by BSA, PSO, ABC, CMAES, JDE, SADE, and CLPSO for 30 solutions for 50 (F1-F50), 25 (F51-F75), and 3 (F76-F78) benchmark problems, respectively. These tests were used to compare BSA with PSO, ABC, CMAES, JDE, SADE, and CLPSO to statistically determine which of them provided the best solution. Figure (3.1) illustrates a graphical comparison of BSA versus CMAES, PSO, ABS, CLPSO, JDE, and SADE to statistically determine which of them provided the best solution for the benchmark problems used in Tests 1, 2, and 3.

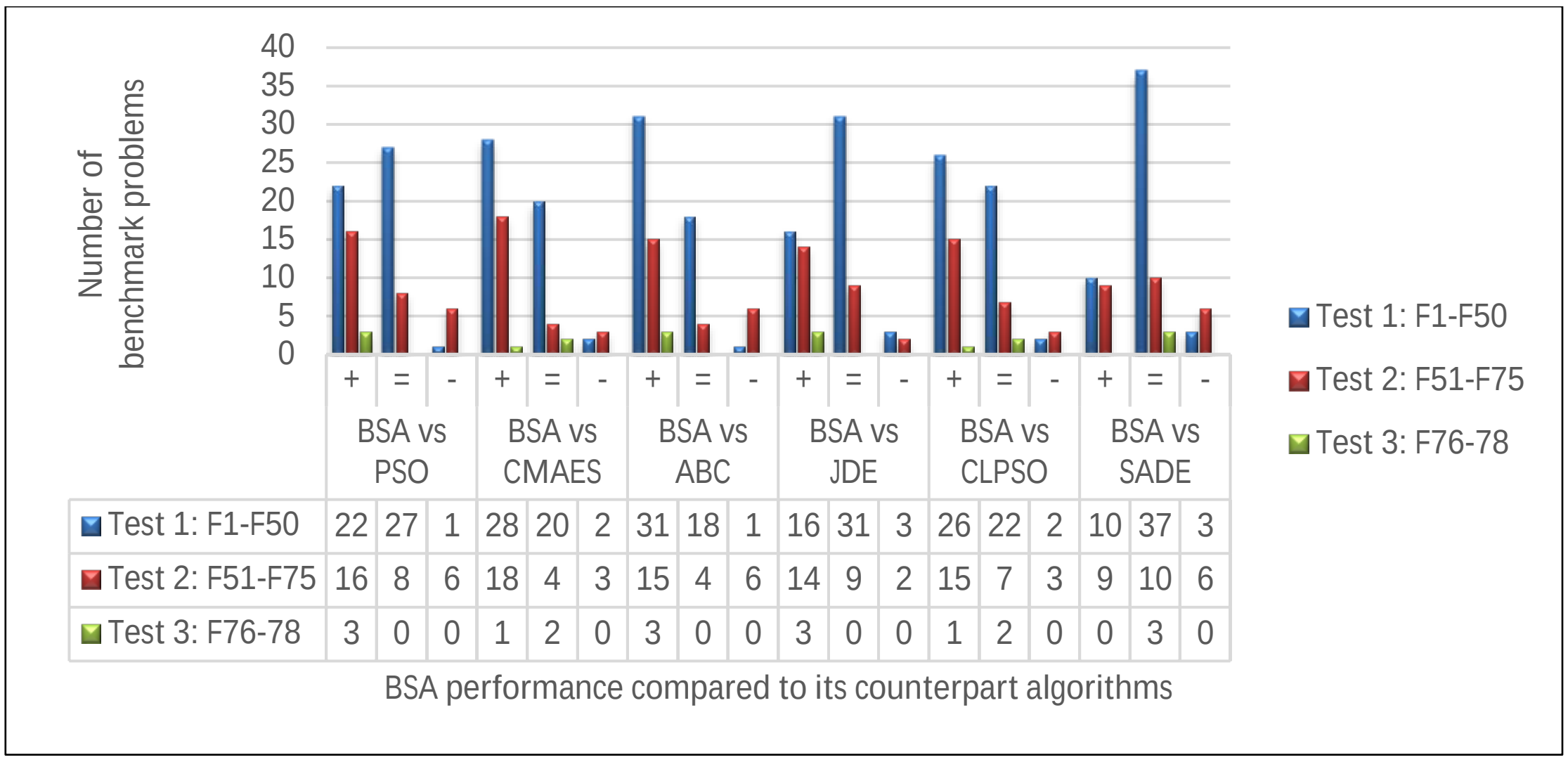

**Figure (3.1): Statistical performance of BSA in comparison to counterpart algorithms on seventy-eight benchmark problems used in Tests 1, 2, and 3**

In Figure (3.1), '+' means that BSA has statistically better performance than the other algorithm at null hypothesis H0 and α = 0.05; '-' indicates that BSA has statistically lower performance compared with its competitor at null hypothesis H0 and α = 0.05; and '=' indicates that BSA has statistically the same performance as its competitor at null hypothesis H0 and α = 0.05. Furthermore, Table (3.1) compares BSA with other EAs for the statistical pairwise multi-problem at a significant statistical value (α = 0.05) and the null hypothesis (H0). The result of the experiment conducted by [11] showed that (i) CMAES, PSO, ABS, CLPSO, and JDE are statistically less successful than BSA in Tests 1 and 2, (ii) BSA is statistically as successful as SADE in Test 3, and (iii) BSA is more successful than PSO, ABC, CMAES, JDE, and CLPSO in Test 3.

**Table (3.1): Comparing BSA with other EAs for statistical pairwise multi-problem (α=0.05) (adapted from** [16]**)**

| | **BSA versus PSO** | **BSA versus CMAES** | **BSA versus ABC** | **BSA versus JDE** | **BSA versus CLPSO** | **BSA versus SADE** |
|---|---|---|---|---|---|---|
| **P-value** | 4.406E-07 | 4.081E-10 | 2.007E-02 | 5.526E-08 | 1.320E-07 | 2.607E-04 |
| **T+** | 145 | 150 | 415 | 121.5 | 186 | 194 |
| **T-** | 1286 | 2196 | 911 | 1418.5 | 1584 | 841 |
| **Winner** | BSA | BSA | BSA | BSA | BSA | BSA |

### 3.1.2 Performance Evaluation

Based on the studies reviewed in the literature, BSA has been used to deal with different types of problems, and it has been evaluated against other popular EAs. Table (3.2) summarises the performance evaluation of BSA against its competitors for different types of problems. For most of the problems, BSA showed superior performance than its competitors of BSA that performed better than BSA. For example, GA gained more improvement in its performance than BSA for the personal scheduling problem.

**Table (3.2): Performance evaluation of the standard BSA against its competitors for different types of problems**

| Type of the problem | Source, author(s) | BSA's competitor(s) | Type of comparison | Winner |
|---|---|---|---|---|
| Numerical optimisation problem | [16], P. Civicioglu | CMAES, PSO, ABS, CLPSO, JDE, and SADE | Effectiveness and performance | BSA |
| Personal scheduling | [75], A.C. Adamuthe, and R.S. Bichkar | GA | Performance for value and variable orderings and techniques of consistency enforcement | GA |
| Combination of centric circular antenna arrays | [76], K. Guney, and A. Durmus | Differential search (DS) and bacterial foraging algorithm (BFA) | Iterative performance | BSA |
| Three design problems circular antenna array | [77], P. Civicioglu | ABC, E2-DSA, CK, CLPSO, CMAES, ACS, JADE, DE, EPSDE, JDE, SADE, GSA, S-DSA, PSO | Side lobe levels minimisation, maximum directivity acquisition, non-uniform null control, and array of planar circular antenna. | S-DSA |
| Economic dispatch problem | [78], A.F. Ali | GA, PSO, BA | Performance | Memetic BSA |
| Optimal power flow problem | [79], U. Kılıç | Those heuristic algorithms that have solved OPF previously, such as GA, SA, Hybrid SFLA-SA, and ABC | Effectiveness and efficiency | Effect and quick reach of BSA to global optimum |
| Optimal power flow problem | [80], A.E. Chaib et al. | DE, PSO, ABC, GA, BBO | Effectiveness and efficiency | BSA |
| Fed-batch fermentation optimisation | [81], M.Z. bin Mohd Zain et al. | DE, CMAES, AAA, and ABC | Performance and robustness of convergence | BSA |
| Global continuous optimisation | [82], S. Mandal et al | ALEP, CPSO-H6, HEA, HTGA, GAAPI | CPU Efficiency and solution of global optima | BSA |
| Fuzzy proportional-integral controller (EPCI) for scaling factor optimisation | [83], P. Gupta et al. | GWO, DE, BA | Performance | GWO |
| stabilisers design of multi-machine power systems | [84], M. Shafiullah et al. | PSO | Robustness | BSA |
| Allocating Multi-type distributed generators | [85], A. El-Fergany | Golden search, Grid search, Analytical, Analytical LSF, Analytical ELF, PSO, ABC | Optimal result | BSA |
| Linear antenna arrays | [76], K. Guney, and A. Durmus | PSO, GA, MTACO, QPM, BFA, TSA, MA, NSGA-2, MODE, MODE/D-DE, CLPSO, HAS, SOA, MVMO, | Pattern nulling | BSA |
| Estimating parameter for non-linear Muskingum model | [86], X. Yuan et al. | PSO, GA, DE | Efficiency | BSA |
| Cost-sensitive feature selection | [87], H. Zhao et al. | Heuristic algorithm | Efficiency and effectiveness | Efficiency: BSA Effectiveness: a heuristic algorithm |
| load frequency control (LFC) in power systems engineering | [88], V. Kumar et al. | BFOA, PSO, GA, DE, and GSA | Performance, robustness, and effectiveness | BSA |

### 3.2 Related Work on Evolutionary Clustering Algorithms

Many clustering algorithms have been developed recently. K-means is one of the most popular and oldest algorithms used as a basis for developing new clustering techniques [89]. The K-means algorithm categorises several data points into $K$ clusters by calculating the summation of squared distances between every point and its nearest cluster centroids [90]. Despite its inaccuracy, the speed and simplicity of K-means are practically attractive for many scholars. Besides, K-means has several drawbacks: (i) It is difficult to predict the number of clusters. (ii) Different initial cluster centroids may result in different final clusters. (iii) K-means does not work well with datasets in different size and density. (v) Noise and outliers may misguide the clustering results. (vi) Increasing the dimension of datasets might result in bad clusters.

To overcome the shortcomings mentioned above, another clustering algorithm based on K-means, called K-means++, was proposed to generate optimal clustering results [53]. In K-means++, a smarter method is used for choosing the cluster centroids and improving the quality of clusters accordingly. Except for the initialisation process, the rest of the K-means++ algorithm is the same as the original K-means technique. That is, K-means++ is the standard K-means linked to a smarter initialisation of cluster centroids. In its initialisation procedure, K-means++ randomly chooses cluster centroids that are far from each other. This process increases the likelihood of initially collecting cluster centroids in various clusters. Because cluster centroids are obtained from the data points, each cluster centroid has certain data points at the end. In the same study, experiments conducted to compare the performance of K-means with K-means++. Though the initialisation process of K-means++ is computationally more expensive than the original K-means, K-means++ consistently outperformed the K-means algorithm. However, the main drawback of K-means++ is its intrinsic sequential behaviour. That means several scans needed over the data to be able to find good initial cluster centroids. This nature limits the usability of K-means++ in big datasets.

This study [91] proposed scalable K-means++ to reduce the number of passes needed to generate proper initialisation. Lately, another technique, called expectation maximisation (EM) was introduced by [92] to maximise the overall likelihood or probability of the data and to give a better clustering result. EM was extended from the standard K-means algorithm by computing the likelihood of the cluster membership, based on one or more probability distributions. On this basis, this algorithm is suitable for estimation problems. It was also addressed that future work on EM would include its use in different and new areas and its enhancement to improve computational structure and convergence speed.

Learning vector quantisation (LVQ) is another crucial algorithm for clustering datasets, which was introduced by [93] to classify the patterns where each output unit is represented by a cluster. Additionally, three variants of LVQ (LVQ2, LVQ2.1, and LVQ3) were developed by [94] for building better clustering results. Nevertheless, LVQ and its variants have reference vector diverging that may decrease their recognition capability. Several studies have been conducted to solve this problem. One of the successful studies was introduced by [95] to update the reference vectors and to minimise the cost function. The experiments indicated that the recognition capability of GLVQ is high in comparison with VLQ.

Furthermore, the involvement of meta-heuristic algorithms in clustering techniques has become popular and successful in many ways. At first sight, a genetic algorithm (GA) with K-means was used as a combined technique to avoid the user input requirement of *k* and enhance the quality of the cluster by exploring good quality optimum [54]. Recently, several GAs with K-means have been proposed. In these approaches, new genetic operators have been introduced to obtain faster algorithms than their predecessors and achieve better quality clustering results [96]. Meanwhile, a new clustering technique was proposed to enhance energy exploitation in wireless sensor networks [97]. The proposed algorithm was combined with bat and chaotic map algorithms. The experimental results achieved with the implementation of the introduced technique had a significant impact on energy usage enhancement, increasing network lifetime, and the number of live nodes in different algorithm rounds.

One of the common GA-based algorithms is called GenClust. Two versions of GenClust were suggested by [98] and [99]. On the one hand, the first GenClust algorithm was suggested by [98] for clustering gene expression datasets. In GenClust, a new GA was used to add up two key features: (i) a smart coding of search space, which is compact, simple, and easy to update. (ii) It can naturally be employed together with data-driven validation methods. Based on experimental results on real datasets, the validity of GenClust was assured compared with other clustering techniques. On the other hand, the second version proposed by [99] as a combination of a new GA with a K-means technique. Several experiments, which were conducted to evaluate the superiority of GenClust with five new algorithms on real datasets, showed a better performance of GenClust compared with its counterpart techniques. However, the second version of GenClust is not free of shortcomings. Therefore, this algorithm was advanced by [100] namely GENCLUST++ by combining GA with the K-means algorithm as a novel re-arrangement of GA operators to produce high-quality clusters. According to the results of the experiments, the introduced algorithm was faster than its predecessor and could achieve a higher quality of clustering results.

After reviewing the above-mentioned clustering techniques based on meta-heuristic algorithms, it was found that these algorithms could produce better clustering results in comparison with the other classical techniques. Consequently, a new cluster evolutionary algorithm was proposed in this study for clustering multi-feature and heterogeneous datasets. This suggested technique is based on the social class ranking with a combination of hybrid techniques of percentiles and quartiles, operators of the meta-heuristic algorithm, random walks in levy flight optimisation, and Euclidean distance of K-means algorithm. This newly introduced algorithm aims for seven-folds: (i) Automatic selection of the number of clusters. (ii) Finding the right cluster centroids for each cluster. (iii) Preserving a balance between population diversity to avoid the algorithm to trap local and global optima. (v) Providing capability of handling scalable benchmarking problems. (vi) Providing a framework to show the sensitivity of ECA* towards different cohorts of datasets in comparison to its counterpart algorithms. (vii) Avoiding the outliers and noise to misguide the ECA* to generate good clustering results. (viii) Providing stability and robustness to our proposed algorithm.

### 3.3 Related Work on Formal Context Reduction

Generating automatic ontologies is one of the on-going research studies in the Semantic Web area. Ontology learning is considered as a resource-demanding and time-consuming task with the involvement of domain experts to generate the ontology manually [101][64]. As a result, endorsing this method would be partially or entirely helpful for semi-automatically or ultimately constructing ontology. Ontology learning can be defined as a subtask of information extraction to semi-automatically derive relations and relevant concepts from text corpora or other types of data sources to construct an ontology.

Various techniques from the fields of natural language processing, machine learning, information retrieval, data mining, and knowledge representation have contributed to enhancing ontology development over the last decade [64]. Data mining, machine learning, and retrieval of information provide statistical techniques to extract specific domain terms, concepts, and associations among them. On the other hand, by providing linguistic techniques, natural language processing plays its role in almost every level of the ontology learning layer cake. FCA is one of the interesting statistical techniques used to construct concept hierarchies in ontology learning. This approach relies on the basic idea whose objects are related to their corresponding attributes property. It takes a matrix attribute of an object as input and finds all the natural attributes and object clusters together. It yields a lattice in the shape of a hierarchy, with definitions and attributes. The concept of the general use of FCA in NLP is undoubtedly not new. In [102], some possible applications of the FCA are stated in the analysis of linguistic structures, lexical semantics, and lexical tuning. [103] and [104] applied FCA to construct more concise hierarchies of lexical inheritance regarding morphological characteristics, such as numerous, gender. In [105], the function of learning sub-categorisation frames from corpora was also extended to FCA. Using Formal Concept Analysis [106] presented a new approach to automatically acquiring concept hierarchies from the text. The results indicated that this approach performed relatively well compared with its counterpart algorithms. Lately, [107] used FCA for the extraction of the taxonomic relation. They also compared it with the taxonomic relation-extraction approach based on agglomeration clustering. On the scientific corpus, FCA was 47 percent accurate, while agglomerated clustering was 71 percent accurate. Their tests showed that not only FCA was complicated in terms of time, but its findings were also less reliable compared with agglomeration.

Recently, [14] presented a novel approach based on FCA to automatically acquire concept hierarchies from domain-specific texts. The approach was also contrasted with a hierarchical agglomerative clustering algorithm as well as with Bi-Bisecting K-Means and found that it yielded better results on the two datasets considered. Further various information measures were examined to assess the importance of an attribute/object pair and concluded that the conditional probability performs well in comparison with other more complicated information measures. The same study automatically depicted the overall process of deriving concept hierarchies from text corpora. In the overall process of automating concept hierarchies using Formal Concept Analysis, a set of steps are included. First, the text is labelled as part-of-speech (POS) and then parsed for each sentence to produce a parse tree. Next, it extracts verb/object, verb/subject, and verb/prepositional phrase dependencies from the parse trees. In particular, pairs containing the verb and the head of the object, subject, or prepositional sentence are extracted. The verb and the head are then lemmatised and transferred to its base form. The process of word pair extraction from text corpora is called NLP components, which is illustrated in Figure (3.2).

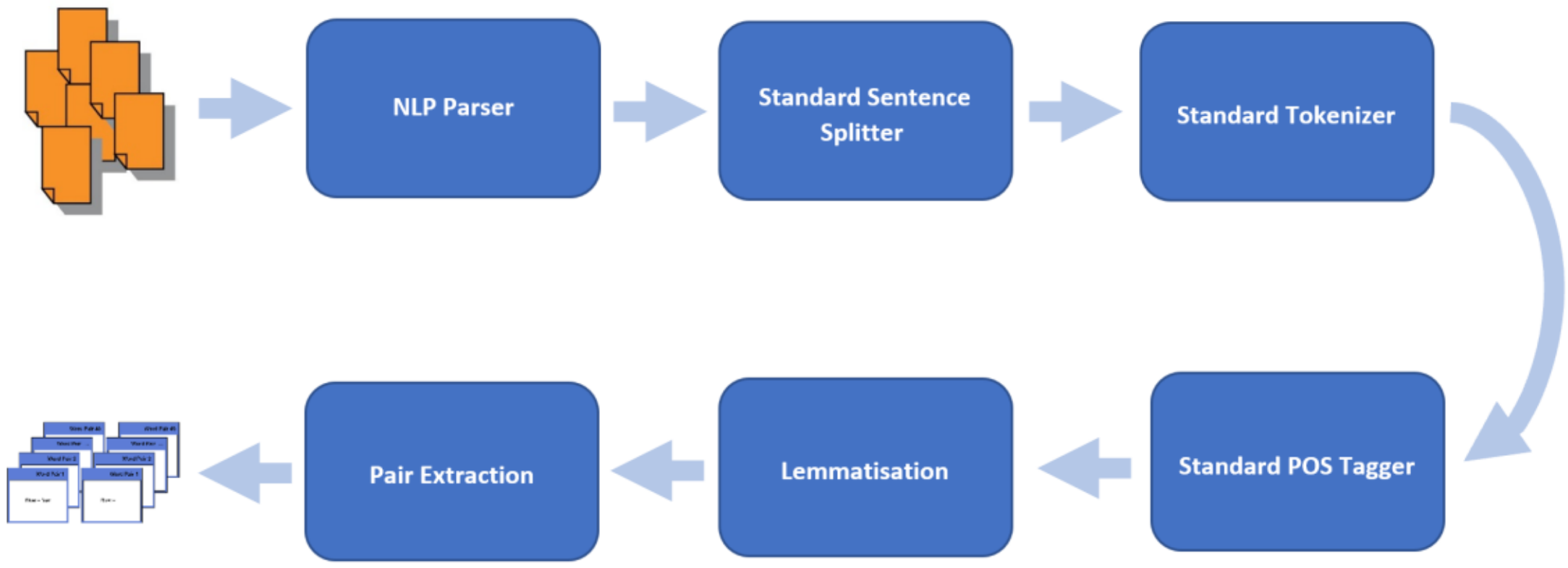

**Figure (3.2) : NLP components (extracted from** [14]**)**

Additionally, a pair collection is smoothed to address data sparseness - meaning that the frequency of pairs that do not appear in the corpus is estimated based on the frequency of other pairs. Finally, the pairs become a formal context to which Formal Concept Analysis is applied. The entire process of automatic construction of concept hierarchies from text corpora is elucidated in Figure (3.3).

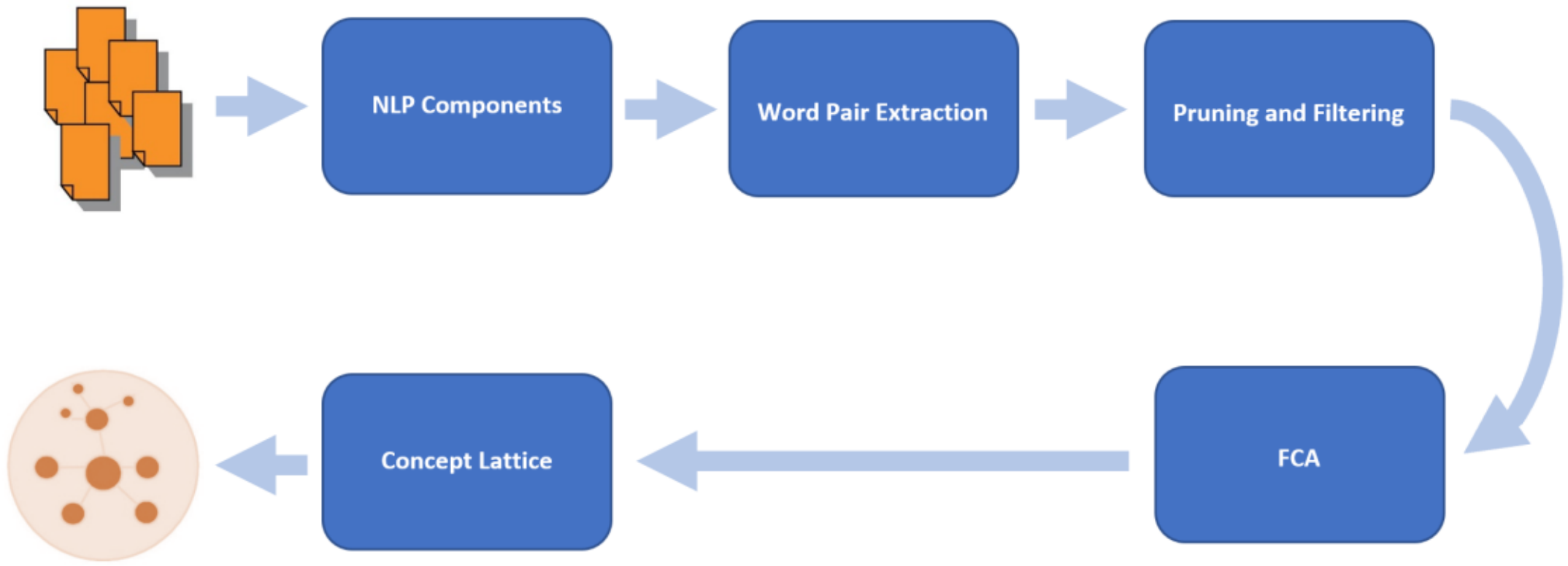

**Figure (3.3): A framework for constructing concept hierarchies (adapted from** [14]**)**

However, the extracted pairs may not always be exotic, and it might result in erroneous concept hierarchies to: derive the right and interesting pairs and reduce the size of a formal context. Reducing the size of formal context may result in removing uninterested and erroneous pairs, taking less time to extract the concept lattice.

Despite a number of techniques for concept lattice reduction reviewed in [108], there are molecular approaches known about the reduced structures of concept lattice that have been generated. The authors of the same review classified the techniques for reducing concept lattice in the literature into three groups:

**1. Redundant information removal:** Generally, the simplified concept lattice of redundant information removal approach has fewer characteristics and may be more suitable for applications involving direct interaction with the user, whether through simple visual interpretation of the lattice or manipulation of the ramifications of the current formal context.

**2. Simplification:** This approach has more applicability than previous ones by transforming the area of the definitions into a smaller one. This feature creates a situation where the consistency of the resulting definition lattice is in contrast with the original. Generally, the techniques are performed in the formal context and have a sophistication that helps to exploit extensive formal contexts. On the other hand, the resulting concept lattice may differ significantly from the original one and may, therefore, be unsatisfactory in quality.

**3. Selection:** This approach works by pruning the space of concepts using some parameters of importance, for example, by adding an objective function that removes paths, and seems meaningless on the concept lattice. The construction of iceberg concept lattices, suggested by [109], is a highly regarded method in this area of techniques. The drawback of this method is that important formal structures will inevitably be forgotten. Further, some techniques list all formal concepts and apply some criteria for selecting the relevant concepts [110]–[112]. A technique based on this method can be costly, as the entire search space is searched.

In the same study [108], a total of forty techniques were analysed and classified based on seven criteria, selected from among the major ones in the literature. The results were summarised in a formal context, and the analysis was carried out using FCA. All reduction techniques have been shown to alter the incidence relation and cause other degrees of descriptive loss. Half of the techniques continue from the set of concepts by considering only obsolete material removal techniques, and the other half from an extension of the FCA. In comparison, three - quarters of them use exact algorithms, and one-quarter use heuristic methods. Concerning the simplification techniques, most of them use the formal sense as a starting point, generate a non-isomorphic non-subset of the original definition lattice, and use exact algorithms. In comparison, some use historical knowledge to assist in the simplification process.

Finally, the techniques of selection start the process of that from the formal context, do not use background knowledge, and use exact algorithms. It is clear that the most considerable reductions, the ones that can significantly reduce the area to be accessed, are those that are achieved by simplification and selection techniques. However, simplification techniques can be considered relatively dangerous as they can change the set of formal concepts substantially. The process leading to such changes must be ensured that the essential parts of the model lattice are retained. Even though selection techniques are effective as they prune the concept space, the traversal of that space must be made in such a way as to ensure that the relevant concepts are reached.

Referring to the literature, few works on quality measures of reduced concept lattices have been reported such as the suggestion of [113]. It is noted that even though each reduction technique has specific characteristics and each approach has its priorities, there are limitations in the information represented in the final concept lattice that needs to be measured. The mechanism for measuring the loss of knowledge is also important when contrasting different methods. What is needed is some way to determine which information has been preserved, deleted, inserted, or altered so that the reduction quality can be evaluated and compared with different techniques. Such quality measures need to undergo further investigation. Thus, despite reviewing the current framework for concept hierarchy construction from text and concept lattice size reduction, this study focuses on proposing a framework for deriving concept hierarchies by taking advantage of deducing formal context size to produce meaningful lattice and concept hierarchy accordingly. The formal context could be significantly large because a large corpus created a formal context. Meanwhile, the formal context can lead to a time-consuming process and depends on the size and nature of the formal context data. As a fudge, using adaptive ECA *, a new framework is introduced to eliminate the wrong, uninteresting pairs in the formal context and to reduce the size of the formal context, and in turn, making the concept lattice less time-consuming. A subsequent experiment was carried out to analyse the effects of the reduced concept lattice against the original one. Resultantly, this thesis endeavours with substantial reduction concept lattice preserving the best quality of the resulting concept lattice. Presumably, the reduction of concept lattice for deriving concept hierarchies from corpora, like in the approach presented in this thesis, has not incidentally been applied before.

# Chapter Four: Methodology

This chapter is divided into six sections. In the first section, an operational framework of the main expansions of BSA and its principal implementations is proposed. In the second section, the performance of BSA is evaluated against its competitive algorithms via conducting experiments. In the third section, a multi-disciplinary evolutionary algorithm called ECA* is proposed for clustering heterogeneous datasets, followed by experiments to evaluate its performance compared to its counterpart techniques. In the fifth section, a framework is proposed using adaptive ECA* to reduce formal context size to derive concept hierarchies from corpora. The final section describes the experiment which is carried out to investigate the reduced formal context size.

## 4.1 Proposed Operational Framework of BSA

This section describes a proposed operational framework of the main expansions of BSA (hybridised BSA, modified BSA, and extensions of BSA) and the implementations of BSA that were suggested in the literature on BSA. The procedures of these three expansions from the basic BSA vary. The modification of BSA can be based on the combination of BSA with another meta-heuristic algorithm, control factor (F), historical information, routing problem, or some different cases such as binary BSA, hybridised MOBSA with a GA, and the use of BSA in the microprocessor. On the contrary, the procedure of hybridising BSA can be based on several approaches. Examples of these methods are control factor (F), guided and memetic BSA, neural network, fuzzy logic, local and global searches, meta-heuristic algorithms, and mathematical functions. Meanwhile, BSA has been extended to be used for binary optimisation, numerical optimisation, constrained optimisation, and multi-objective optimisation. These extensions are based on the combination of BSA with another meta-heuristic algorithm, three constraint handlers, network topology, new mutation, and crossover operations. A summary of the different expansions of BSA and their expansion procedures is presented in an operational framework in Figure (4.1).

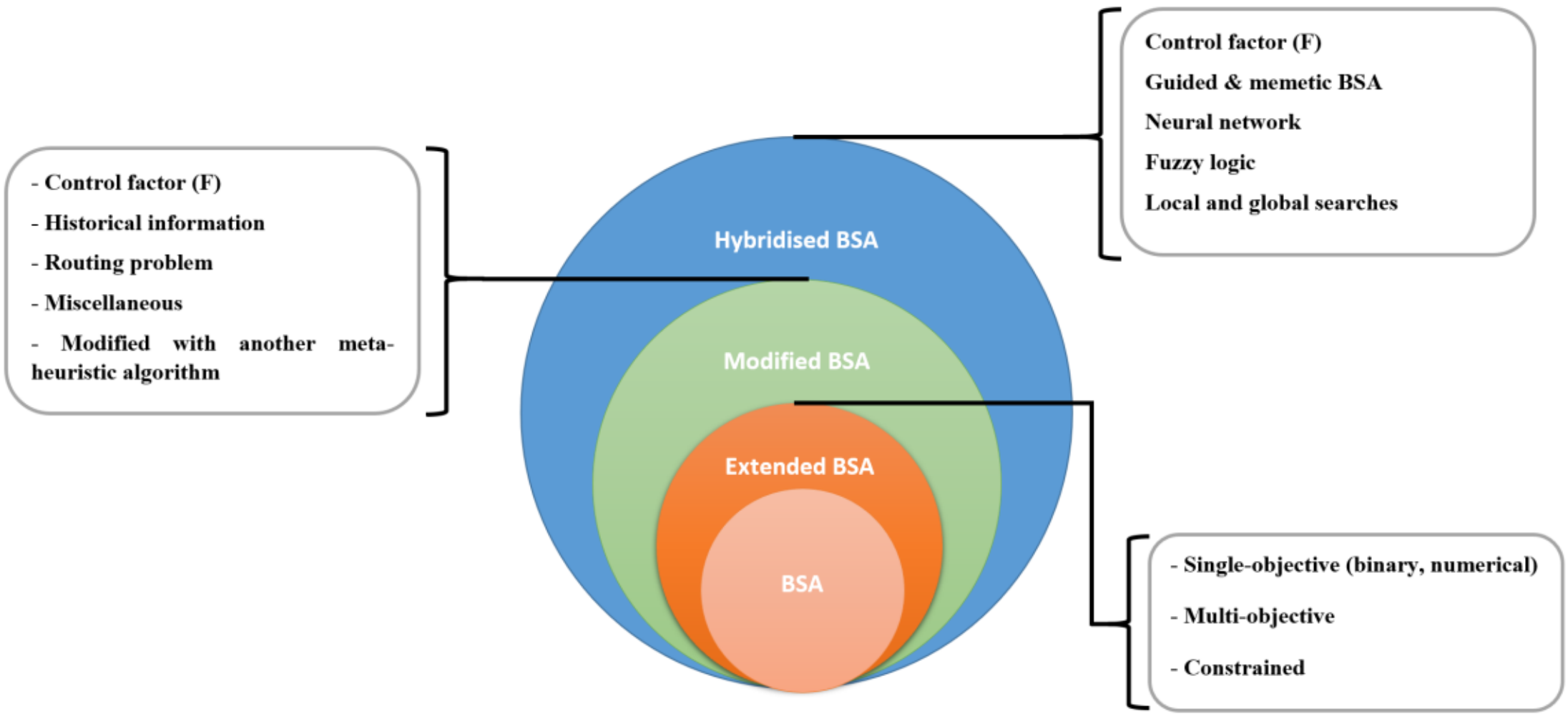

**Figure (4.1): The proposed operational framework of BSA**

## 4.2 Performance Evaluation of BSA

From the previous studies that were conducted to compare the nature of the performance of BSA with its competitors, a mean result, standard deviation, the worst and best result were extracted [16], [114], [115]. Nevertheless, an equitable and fair comparison of several metaheuristic algorithms in terms of performance is a difficult task owing to several reasons. Particularly, the selection and initialisation of the key parameters (e.g., the problem dimension, the problem's search space for each algorithm, and the number of iterations required to minimise the problem) are challenging. Another issue is the involvement of a randomness process in every algorithm. To weave out the effects of this randomisation and obtain an average result, the tests need to be conducted on the algorithms several times. The performance of the system used to implement the algorithms and the programming style of the algorithms may also affect the algorithms' classification performance. To mitigate this effect, a consistent code style for all the algorithms on the same system can be utilised. A further challenging task includes choosing the type of problems or benchmarking test functions that are used to evaluate the algorithms. For instance, an algorithm may perform satisfactory for a specific type of problem but not on other types of problems. Additionally, the performance of an algorithm can be affected by the size of a problem. A set of standard problems or benchmark test functions with different levels of difficulties and different cohorts need to be used to address the aforementioned issue, and thereby to test the algorithms on various levels of optimisation problems. Therefore, an experimental setup to fairly compare BSA with its competitive meta-heuristic algorithms requires the consideration of aspects such as initialisation of the control parameters, balance of the randomness process of the algorithms, computer performance used to implement the algorithms, programming style of the algorithms, and type of the tackled problems; this is owing to the lack of experimental study on BSA in the literature. Therefore, this section details the experimental setup, the list of benchmark functions used in the experiments with their control parameters, statistical analysis of the experiment, pairwise statistical testing tools, and the experimental results.

### 4.2.1 Experimental Setup

The experiment is to fairly compare BSA with PSO, ABC, FF, and DE by considering the initialisation of control parameters, such as the problem dimensions, problem search space, and a number of iterations. Setting up these initialisation parameters are required to minimise the problem, the performance of the system used to implement the algorithms, the programming style of the algorithms, achieving balance on the effect of randomisation, and the use of different types of optimisation problems in terms of hardness and cohorts. This experiment is about the unbiased comparison of BSA with PSO, ABC, FF, and DE on 16 benchmark problems with different levels of hardness [116]–[118] in three tests as follows:

1. Several iterations are needed to minimise a specific function with Nvar variables with the default search space for the population size of 30. Nvars take values of 10, 30, and 60. For each benchmark function, each algorithm is run for 30 times with 2000 iterations for Nvar values of 10, 30, and 60.

2. Several iterations are needed to minimise the functions with two variables for three different sized solution spaces for the population size of 30. For each benchmark function, each algorithm is run for 30 times with 2000 iterations for three different ranges (R1, R2, and R3), where

a) R1: [-5, 5]
b) R2: [-250, 250]
c) R3: [-500, 500]

3. Determining the ratio of successful minimisation of the functions for Tests 1 and 2 is needed to compare the success rate of BSA with its competitive algorithms.

Furthermore, Figure (4.2) depicts the framework for the experimental setup that includes the processes of Tests 1, 2, and 3.

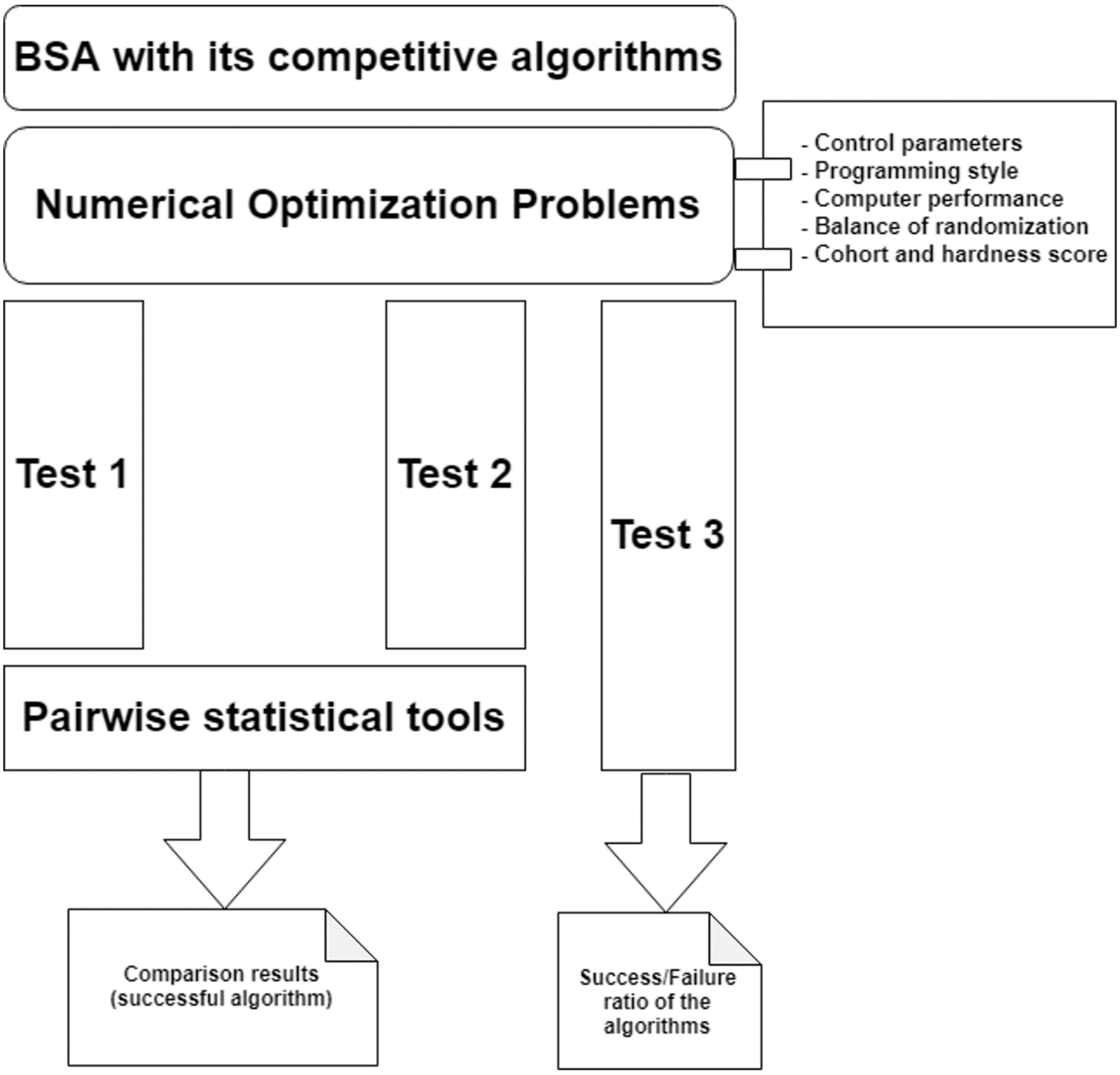

**Figure (4.2): The proposed framework for the experimental setup**

### 4.2.2 Benchmark Functions and Control Parameters

To examine the relative success of BSA compared to PSO, DE, ABC, and FF in solving a variety of optimisation problems based on their level of hardness and cohort [116]–[118], 16 benchmark problems are used in all the three tests. Regardless of the algorithm chosen to solve the test functions, some of the chosen problems are harder to minimise than the others. Table (4.1) lists the chosen optimisation algorithms with their search space, global minima, dimension, and percentage level of hardness. The percentage level of hardness ranges from 4.92 to 82.75%. Among the problems listed in the table, the Whitley function is the hardest, whereas the Sphere is the easiest.

**Table (4.1): Benchmark testing functions**

| ID | Function | Search space | Global min. | Dim. | Overall success (%) |
|---|---|---|---|---|---|
| F1 | Ackley | [-32, 32] | 0 | n | 48.25 |
| F2 | Alpine01 | [0, 10] | 0 | 2 | 65.17 |
| F3 | Bird | [-6.283185, | -106.76453 | 2 | 59.00 |
| F4 | Leon | [0, 10] | 0 | 2 | 41.17 |
| F5 | CrossInTray | [-10, 10] | -2.062611 | 2 | 74.08 |
| F6 | Easom | [-100, 100] | -1 | 2 | 26.08 |
| F7 | Whitley | [-10.24, | 0 | 2 | 4.92 |
| F8 | EggCrate | [-5, 5] | 0 | 2 | 64.92 |
| F9 | Griewank | [-600, 600] | 0 | n | 6.08 |
| F10 | HolderTable | [-10, 10] | -19.2085 | 2 | 80.08 |
| F11 | Rastrigin | [-5.12, 5.12] | 0 | n | 39.50 |
| F12 | Rosenbrock | [-5, 10] | 0 | n | 44.17 |
| F13 | Salomon | [-100, 100] | 0 | 2 | 10.33 |
| F14 | Sphere | [-1, 1] | 0 | 2 | 82.75 |
| F15 | StyblinskiTang | [-5, 5] | -39.1661 | n | 70.50 |
| F16 | Schwefel26 | [-500, 500] | 0 | 2 | 62.67 |

Additionally, for each benchmark function, each algorithm is run for 30 times with a maximum of 2000 iterations with a population size of 30 for all the three tests. Meanwhile, the search spaces and dimensions of the benchmark problems vary from Tests 1, and 2. For Test 1, the dimension of each benchmark function was set to default, and its search space is categorised as R1, R2, and R3. For Test 2, the search space of each benchmark function was set to two variables, and its dimension was categorised as Nvar1, Nvar2, and Nvar3. Finally, determining the ratio of successful minimisation of the functions for Tests 1 and 2 was needed to compare the success rate of BSA with its competitive algorithms. This initialisation of parameters is described in Table (4.2).

**Table (4.2): Control parameters of Test 1 and 2**

| Tests | Features of optimisation problems | | | Initial parameters | | |
|---|---|---|---|---|---|---|
| | Dimension | Search space | Hardness score | Maximum run | Iterations | Population size |
| Test 1 | Default dimension | R1: [-5, 5] | 4.92% - 82.75% | 30 | 2000 | 30 |
| | | R2: [-250, 250] | | | | |
| | | R3: [-500, 500] | | | | |
| Test 2 | Nvar1: 10 | 2 | | | | |
| | Nvar2: 30 | | | | | |
| | Nvar3: 60 | | | | | |

### 4.2.3 Statistical Analysis

Occasionally, EAs may provide the worst and best solutions from time to time for a specific problem, e.g., if an algorithm runs for two times on a specific problem, it may obtain the best solution initially and the worst solution later, and vice versa. Based on this, statistical tools were used in the literature [16], [17] to compare the success or failure of the problem-solving of BSA with the other EAs. In this study's experiment, the following seven statistical measures were used to solve the numerical optimisation problems: mean, standard deviation, best, worst, average computation time, number of successful minimisations, and number of failed minimisations.

### 4.2.4 Pairwise Statistical Tools

From Pairwise statistical testing tools, the Wilcoxon signed-rank test was utilised to compare two algorithms and determine which algorithm can solve a specific optimisation problem with higher statistical success [16], [119]. In the experiment, BSA was compared with other algorithms by using the Wilcoxon signed-rank test, where the significant statistical value (α) is considered as 0.05, and the null hypothesis (H0) for a specific benchmark problem is defined in Equation (4.1) [119] as follows:

$$Median\ (Algorithm\ A) = Median\ (Algorithm\ B) \tag{4.1}$$

To determine the algorithm that achieved a better solution in terms of statistics, R+, R-, and p-value are provided by the Wilcoxon signed-rank test to determine the ranking size. In the same experiment, GraphPad Prism is used to determine the R+ and R- values that ranged from 0 to 465. P-value is similar to the mathematical precision of current software and application development tools; it is 4-6. Generally, the precise value used for statistical Tests 1 and 2 was 6 because this precision level could be required in practical applications.

## 4.3 ECA*

Several new ideas were integrated to introduce a new and high-quality clustering algorithm. The first idea was to choose good quality of chromosomes instead of choosing them randomly. Using the notion of social class rankings can obtain a set of reasonably good chromosomes. Hence, ECA* uses percentile ranks for the genes of each chromosome. Based on their percentile ranks, each group of related chromosomes was allocated a cluster. Details will be presented below when discussing the ingredients of ECA* [120], [121]. The second notion is the novelty of selecting and re-defining the cluster centroids (*C*) and their historical cluster centroids ($C^{old}$) using quartile and intraCluster. The value of cluster centroids ($C^{old}$) was determined using the following two steps. Initially, the cluster centroids were generated using mean quartiles, while the old cluster centroids were randomly produced between the lower and upper quartiles to avoid misguidance by noise and outliers. At the initiation of each iteration, the $C^{old}$ of each cluster was re-defined based on the value of its intraCluster. Another intervention applied to ECA* was using the operators of evolutionary algorithms. Both mutation (Mutant) and crossover were operated on the cluster centroids to pull the clusters to reach a meaningful result. At first, each cluster centroids was mutated using control parameter (*F*) and learning knowledge from utilising the better historical cluster centroid information (*HI*). *HI* is defined using the differences between *C* and $C^{old}$ if intraCluster of C is higher than intraCluster of $\mathrm{C}^{\mathrm{old}}$ and vice versa. F is a random walk generator using LFO. Crossover ($C^{new}$) was computed using a uniform crossover between C and $\mathrm{C}^{\mathrm{old}}$. More significantly, these convergence operators were recombined to define new evolutionary operator called Mut-over. This new operation was used in ECA* as a greedy selection between mutation and crossover operators. Mut-over is the value of Mutant if the interCluster of $C^{old}$ is higher than the interCluster of C and vice versa. This recombination strategy pulls the cluster centroids and their related chromosomes towards a local maximum/minimum. Besides, this study's algorithm allows merging clusters based on a pre-defined cluster density threshold to combine less-density clusters with others and produce the optimum number of clusters accordingly. The average intraCluster and the minimum distance between two clusters are the main criteria to decide upon their combination. Above all, cluster cost is defined as the objective function. Since ECA* is an evolutionary algorithm for performing clustering, our objective function for every problem is cluster cost. The optimal cluster cost aims to minimise the interCluster distance value and to maximise the intraCluster distance value.

To calculate these distance measures, *A* and *B* are assumed to be generated clusters from a clustering algorithm. *D (x, y)* is a Euclidean distance between two observations *x* and *y* belonging to *A* and B respectively. *D (x, y)* was computed using the Euclidean distance. |*A*| and |*B*| are the number of observations in clusters *A* and *B*, respectively. There are many definitions for calculating the interCluster distance and intraCluster distance [122]. These two distance measures were used in this study: (i)

IntraCluster is the average distance between all the observations belonging to the same cluster. (ii) InterCluster is the distance between the cluster centroid and all the observations belonging to a different cluster.

IntraCluster was calculated, as shown in Equation (4.2) [122].

$$intraclass\ of\ A = \frac{1}{|A|.(|A|-1)} \sum_{\substack{x,y \in A \\ x \neq y}}^{n} \{d(x, y)\} \tag{4.2}$$

InterCluster was computed as expressed in Equation (4.3) [122].

$$\text{InterCluster of (A, B)} = \frac{1}{|A|+|B|} \{ \sum_{x \in A} d(x, v_b) + \sum_{y \in B} d(y, v_a) \} \tag{4.3}$$

Where:

$$v_a = \frac{1}{|A|} \sum_{x \in A} x \tag{4.4}$$

$$v_b = \frac{1}{|B|} \sum_{y \in B} y \tag{4.5}$$

As it is detailed in Figure (4.3), ECA* includes four principal ingredients: Initialisation, Clustering-I, Mut-over, Clustering-II, and Fitness evaluation.

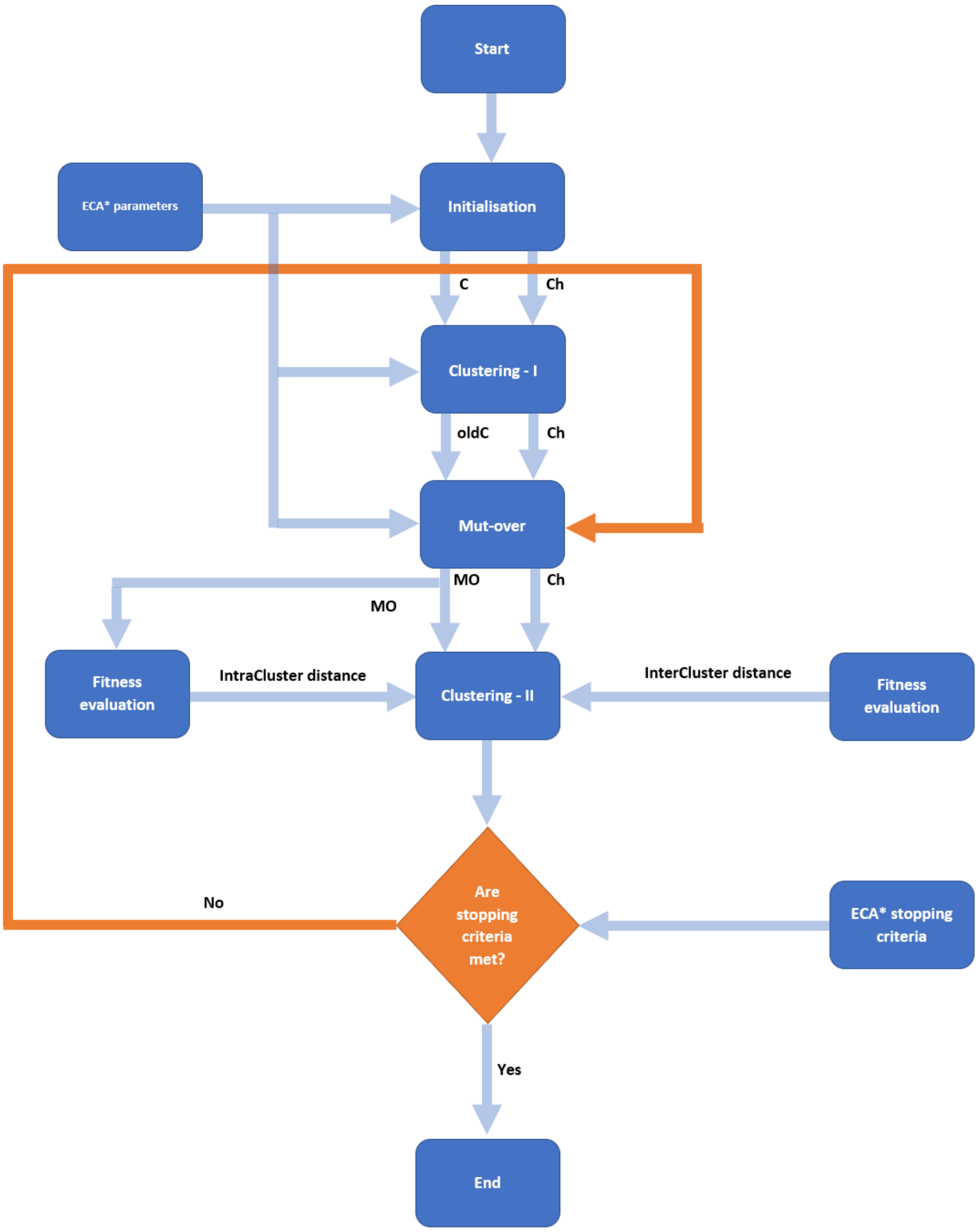

**Figure (4.3): A detailed flowchart of ECA***

**A. Initialisation.** The input dataset contains several records, and each record has several numerical or/and categorical attributes. It is assumed that the input data (Dataset) is represented as *N* chromosomes $Ch_i$, and each chromosome was composed of a set of genes ($G_{i0}, G_{i1}, G_{i2},...G_{ij}$). For instance, $Ch_{ij} = (G_{i0}, G_{i1},\ldots, G_{ij})$ has *j* number of genes.

For i=0, 1, 2,….., N and j=0, 1, 2,….., D Where N and D are the numbers of records and attributes respectively.

At this step, the below parameters were initialised.

1. Initialising the number of social class ranks (*S*).
2. After initialising the number of social ranks (S), the number of clusters (K) as an initial solution for the clustering problem was calculated from S power to number genes in each chromosome as shown in Equation (4.6) below:

$$K = S^j \quad (4.6)$$

Where:

j is the number of attributes in the dataset

3. Initialising the minimum cluster density threshold ($C^{dth}$).
4. Initialising the position of random walk (*F*).
5. Initialising the type of crossover ($C^{type}$).
6. Finding the percentile rank ($P_{ij}$) for each gene in a chromosome
7. Finding an average percentile rank ($P_i$) for each chromosome.
8. Each chromosome was allocated to a cluster (*K*) based on its percentile rank.

**B. Clustering-I.** In this component, two main steps were calculated. (i) The number of clusters ($K^{new}$) was calculated after removing the empty clusters. (ii) The cluster centroids were generated from the actual clusters. For each group of chromosomes, *Ci* was calculated as the cluster centroids as follows:

1. In addition to the number of new clusters ($K^{new}$), the empty clusters ($K^{empty}$), and the less-density clusters ($K^{dth}$) will be removed. The empty clusters are defined as those clusters that have no chromosomes. The actual number of clusters were computed as presented in Equation (4.7).

   $$K^{new} = K - K^{empty} - K^{dth} \quad (4.7)$$

   Besides, the less-density clusters are those clusters, in which their density is less than the minimum density cluster threshold. These clusters may be merged with their nearest neighbouring clusters. The cluster density for *i* was calculated as presented in Expression (4.8).

   $$CD_i = \frac{Number\ of\ chromosomes\ belong\ to\ i}{N} \quad (4.8)$$

   Where:

   *N* is the total number of populations.

2. Generate the initial cluster centroids for each cluster from its chromosomes from the quartile mean of each cluster. The initial cluster centroids were determined as depicted in Equation (4.9).

$$C_i = mean\ quartile\ (Ch_{ij}) \quad (4.9)$$

Where:

$Ch_{ij}$ is a set of chromosomes.

3. At the beginning of each iteration, the new clusters (old$C_i$) were generated randomly in between lower and upper quartiles, for each cluster of $i$, old$C_i$ is redefined by using Equation (4.10) as follows:

$$oldC_i \sim U\ (lQuartile, uQuartile) \quad (4.10)$$

4. At the initiation of each iteration, intraCluster and oldIntraCluster were calculated for each cluster. Likewise, interCluster and oldInterCluster were computed for the current clustering solution.
5. After determining the historical centroids, new cluster centroids were calculated as presented in Equation (4.11) as follows:

$$newC_i := \begin{cases} C_i & intraCluster_i < oldIntraCluster_i \\ oldC_i & otherwise \end{cases} \quad (4.11)$$

**C. Mut-over.** This strategy is also called a greedy operator, which consists of a recombination operator of mutation and crossover.

1. Mutation: Mutate each cluster centroid of *i* in order to move to the most density part of the clusters. Mutating each cluster was calculated according to Equation (4.12) as shown below:

$$Mutant_i := C_i + F\ (HI) \quad (4.12)$$

Where:

*F*: The initial position of random walk of levy flight optimisation.

*HI*: Learning knowledge from utilising the better historical information. Finding *HI* from historical information (*oldC*). *HI* was computed as presented in Equation (4.13) below:

$$HI := \begin{cases} oldC_i - C_i & If\ intraCluster_i < oldIntraCluster_i \\ C_i - oldC_i & Otherwise \end{cases} \quad (4.13)$$

2. Crossover: A new cluster centroid was generated from the current and old cluster centroids using a uniform crossover operator. For cluster i, the new cluster centroid (new$C_i$) was calculated as shown in Equation (4.14)

$$newC_i := oldC_i + C_i \quad (4.14)$$

3. Mut-over: It is a changeover operator between crossover and mutation of ECA* to generate the final trial of centroids between mutation and crossover based on objective functions as it is described in Equation (4.15).

$$MO_i := \begin{cases} Mutant_i \; if \; oldInterCluster_i > interCluster_i \\ newC_i \quad\quad\quad\quad\quad\quad\quad\quad Otherwsie \end{cases} \quad (4.15)$$

Some cluster centroids obtained at the end of ECA*'s crossover process may overflow their search space limitations as a consequence of ECA*'s mutation strategy. The Mut-over strategy was regenerated beyond the search space limitations using Algorithm (4.1). ECA*'s mut-over operator is similar to BSA's boundary control mechanism [16]. Plus, ECA*'s boundary control mechanism is effectual in generating population diversity that ensures effective searches for generating good clustering and cluster centroid results.

```
Input: K, MO, Mutant, C, InterClass, oldInterClass
Output: MO
for i ← 0 to K do
    if (oldInterClass_i > interClass_i) then
        MO_i := Mutant_i;
    end
    else
        MO_i := C_i;
    end
end
```

**Algorithm (4.1): Mut-over strategy of ECA***

**D. Clustering-II.** After the mut-over operator was generated, the clusters were merged based on their diversity, and then the cluster centroids were recalculated. In this step, merging between the close clusters was involved. In turn, a distance measurement between the closed clusters were taken into account. There are several widely used distance measurements between clusters [123]. The most commonly used methods are minimum and maximum distance methods, centroid distance method, and cluster-average method. The minimum distance method was used in this study [124]. On these grounds, the minimum distance method was adopted in this study, in which minimum the distance between clusters was considered as their merging criterion norm. Let $X$ and $Y$ be two clusters. Therefore, the minimum distance between them can be defined in Equation (4.16), as shown below:

$$D_{min}\,(X,Y) := \min\{d(A_x, A_y)\} \quad (4.16)$$

Where:

$A_x \in X, A_y \in Y$

The diversity between Ci and Cj is represented in Equation (4.17) as follows:

$$\sigma_{ij} = \min\left\{\left(D_{min}\left(C_i, C_j\right) - R(C_i)\right)\ \left(D_{min}\left(C_i, C_j\right) - R\left(C_j\right)\right)\right\} \quad (4.17)$$

Where:

R(ci) and R(Cj) are the average distance of the intra-cluster, $D_{min}\left(C_i, C_j\right)$ and $D_{min}\left(C_i, C_j\right)$ are the minimum distance between Ci and Cj

The merging criterion of two classes can be either: (1) If the result ($\sigma \leq 0$) is less than or equal to zero, it means these two classes are close to each other, they are interconnected to a large extent. Therefore, class Ci and Cj can be merged into one class (Cij). That is after less-density class could be an empty class. (2) If $\sigma > 0$, it means that the average distance of intra-class between these two classes is less than their respective shortest distance. This suggests that Ci and Cj persist as two different clusters. At last, the number of clusters were re-calculated to remove the empty clusters as they are produced from the less-density clusters.

**F. Fitness evaluation.** The clusters are measured by fitness evaluation. The inputs are interCluster distance and intraCluster distance of the generated clusters, and the output is the fitness. When the interCluster reaches the maximum and intraCluster reaches minimum (optimal), the algorithm stops. However, this is not realistic. The stopping criteria can be met if: (i). The algorithm reaches the initialised number of iterations. (ii). The value of interCluster and intraCluster remain un-changed during each iteration. That is, the interCluster value is not increased, and intraCluster value is not decreased during each iteration.

The overall pseudo-code of ECA* is represented in Algorithm (4.2). Meanwhile, The software codes in Java of ECA* can be found in [125].

```
Input: S, K, C^dth, F, C^type, maxcycle, P, N, D, Ch_ij, sig_ij
Output: A set of classes (MO), class centroids (C)
//1. INITIALISATION;
K = S^D for i ← 0 to N do
    for j ← 0 to D do
        Ch_ij = Dataset_ij
    end
end
for i ← 0 to N do
    for j ← 0 to D do
        P_ij = Percentilisation (Ch_ij)
    end
end
for iteration ← 1 to maxcycle do
    //2. CLUSTERING-I;
    K^new = K - K^empty ;
    for i ← 0 to N do
        if ((Ch_i belongs to P_i ) then
            C_i := MeanQuartilei(Ch_i);
        end
    end
    // 3. MUT-OVER;
    // MUTATION and CROSSOVER;
    for i ← 0 to N do
        if ((intraClass_i ¡ oldIntraClass_i ) then
            HI_i := oldC_i - C_i;
        end
        else
            HI_i := C_i - oldC_i;
        end
        Mutant = C_i + F (H_i);
        newC_i = uniformCrossover (oldC_i + C_i);
    end
    // MUT-OVER OPERATOR;
    // 4. CLUSTERING-II;
    for i ← 0 to K do
        j = i +1 ;
        Calculate sig_ij
        if ((sig_ij <= 0)) then
            Merge C_i with C_j
        end
        // 5. FITNESS EVALUATION;
        if (oldInterClass_i = interClass_i) AND (oldIntraClass_i = intraClass_i) then
            // Export the class centroids and their observations;
        end
    end
```

**Algorithm (4.2): The overall pseudo-code of ECA***

## 4.4 Experiments on ECA*

This experiment is divided into two sections. The first section presents 32 heterogenous and multi-featured datasets used in this experiment. In the second section, the experiment is setup.

### 4.4.1 Clustering Benchmark Datasets

This section presents heterogenous, multi-featured, and various types of datasets, which were used to evaluate the performance of ECA* compared to its counterpart algorithms. The benchmark datasets are challenging enough for most typical clustering algorithms to be solved, but simple enough for a proper clustering algorithm to find the right cluster centroids. The benchmarking data include the following two groups of well-known datasets, which are publicly available in [51].

- Two-dimensional datasets: include A-sets, S-sets, Birch, Un-balance, and Shape sets. The basic properties of these datasets are presented in Table (4.3).
- N-dimensional datasets: DIM (high), and G2 sets. The basic properties of these datasets are described in Table (4.4).

**Table (4.3): Two-dimensional clustering benchmark datasets** [51]

| **Dataset** | **Varying** | **Type** | **Number of observations** | **Number of clusters** |
|---|---|---|---|---|
| S | Overlap | S1 | 5000 | 15 |
| | | S2 | | |
| | | S3 | | |
| | | S4 | | |
| A | Number of clusters | A1 | 3000 | 20 |
| | | A2 | 5250 | 35 |
| | | A3 | 7500 | 50 |
| Birch | Structure | Birch1 | 100,000 | 100 |
| Un-balance | Both sparse and dense clusters | Un-balance | 6500 | 8 |
| Shape sets | Cluster shapes and cluster number | Aggregation | 788 | 7 |
| | | Compound | 399 | 6 |
| | | Path-based | 300 | 3 |
| | | D31 | 3100 | 31 |
| | | R15 | 600 | 15 |
| | | Jain | 373 | 2 |
| | | Flame | 240 | 2 |

**Table (4.4): Multi-dimensional clustering benchmark datasets** [51]

| Dataset | Varying | type | Dimension-variable | Number of observations | Number of clusters |
|---|---|---|---|---|---|
| DIM (high) | Well separated clusters (cluster structure) | Dim-32 | 32 | 1024 | 16 |
| | | Dim-64 | 64 | | |
| | | Dim-128 | 128 | | |
| | | Dim-256 | 256 | | |
| | | Dim-512 | 512 | | |
| | | Dim-1024 | 1024 | | |
| G2 sets: | Cluster dimensions and overlap | G2-16-10 | 10 | 2048 | 2 |
| | | G2-16-30 | 30 | | |
| | | G2-16-60 | 60 | | |
| | | G2-16-80 | 80 | | |
| | | G2-16-100 | 100 | | |
| | | G2-1024-10 | 10 | | |
| | | G2-1024-30 | 30 | | |
| | | G2-1024-60 | 60 | | |
| | | G2-1024-80 | 80 | | |
| | | G2-1024-100 | 100 | | |

### 4.4.2 Experimental Setup

The experiment was carried out to empirically study the performance of ECA* in comparison with its five competitive techniques (KM, KM++, EM, LVQ, and GENCLUST++). The main objectives of this experiment are three-fold: (1) studying the performance of ECA* against its counterpart algorithms on heterogeneous and multi-featured of datasets. (2) evaluating the performance of ECA* against its counterpart algorithms using internal and external measuring criteria (3) proposing a performance framework to investigate how sensitive the performance of these algorithms on different dataset features (cluster overlap, number of clusters, cluster dimensionality, cluster structure, and cluster shape). Because clustering solutions, produced by the algorithms, can differ between different runs, ECA* was run with its counterpart algorithms 30 times per benchmark dataset to record their cluster quality for each run. Weka 3.9 was used to tun the five competitive techniques of ECA* per each dataset problem. the average results were also recorded for the (30) times clustering solutions on each dataset problem for each technique. Additionally, three pre-defined parameters were initialised for all the techniques as presented in Table (4.5).

**Table (4.5): The pre-defined initial parameters**

| **Parameters** | **Initial assumption** |
|---|---|
| Cluster density threshold | 0.01 |
| Alpha (random walk) | 1.001 |
| Number of social class rank | 2-10 |
| Number of iterations | 50 |
| Number of runs | 30 |
| Type of crossover operator | Uniform crossover |

## 4.5 A Proposed Framework for Concept Hierarchy Construction

As mentioned earlier, this approach aims to reduce the size of formal context as possible as it can to remain the reduced concept lattice to the best quality. There are several methods for reducing the size of a formal context. The used technique in this research for concept lattice reduction are ones that aim at reducing the sophistication of a concept lattice, both in terms of interrelationships and magnitude, while retaining relevant information. Reducing the size of concept lattice leads to derive less sophisticated concept hierarchies. Figure (4.4) illustrates the flow of producing the concept hierarchies using adaptive ECA* with the aid of WordNet.

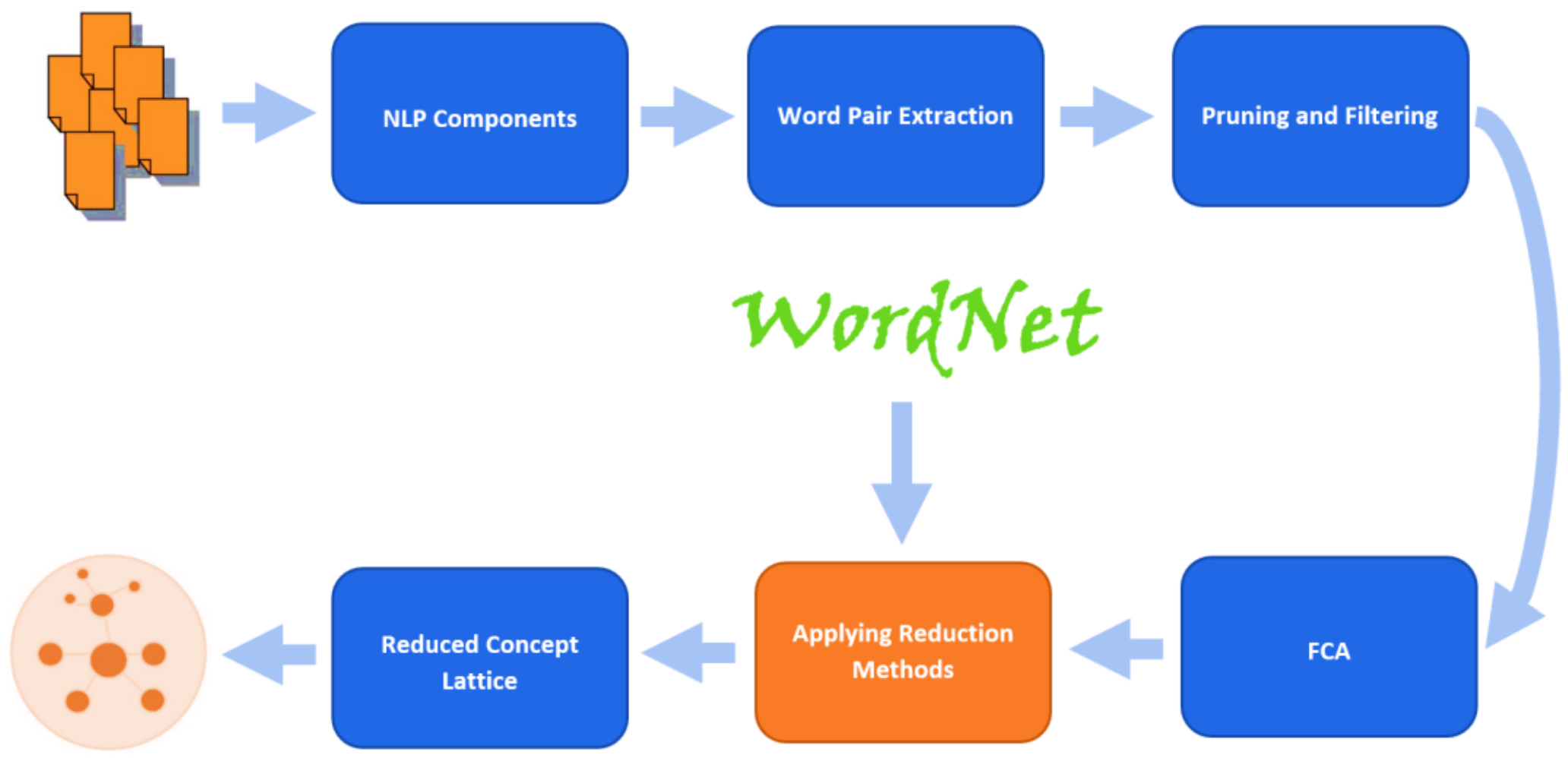

**Figure (4.4): A proposed framework of deriving concept hierarchies using adaptive ECA***

Since the introduced approach was based on multi-disciplinary techniques, it can, therefore, be highly effective [126]. ECA* consists of five components: Initialisation, Clustering I, Mut-over, Clustering II, and Evaluation. In this study, the operators of ECA* were adapted as follows: Initialisation, Construction I, Mut-over, Construction II, and Evaluation. The ingredients of adaptive ECA* are depicted in Figure (4.5).

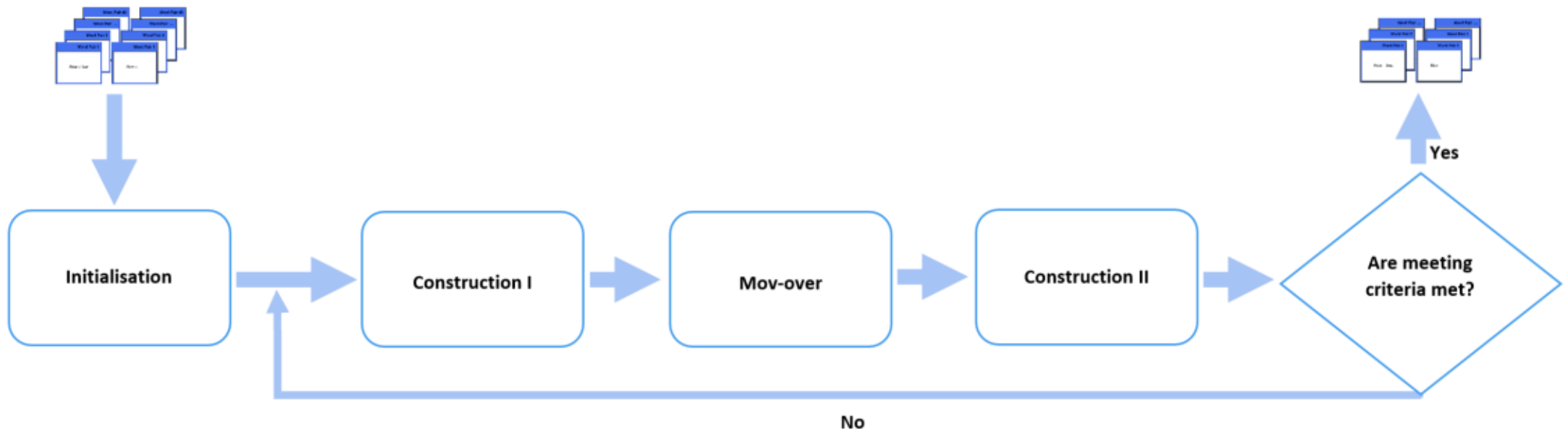

**Figure (4.5): The basic flowchart of adaptive ECA***

After FCA was applied to the pruned word pairs, a formal context was constructed. The formal context provides details about how attributes and objects should be related together.  That means, the formal context is a logical table that can be represented by a triplet *(O, A, I)*, in which *I*, has a binary relationship between *O* and *A*. Elements of *O* are called objects that match the table rows. At the same time, elements of A were named attributes and match the table columns. Moreover, for $o \in O$ and $a \in A$, $(o, a) \in I$ indicates that object *o* has attribute *y* whereas $(o, a) \notin I$ indicates that *o* does not have *a*. For example, Table (4.6) presents logical objects and their attributes. The corresponding triplet *(O, A, I)* is given by $O = \{o_1, o_2, o_3, \ldots, o_n\}$, $A = \{a_1, a_2, a_3, \ldots, a_m\}$, and we have $(o_1, a_2) \in I$, $(o_2, a_3) \notin I$.

**Table (4.6): A formal context example**

| | **$A_1$** | **$A_2$** | **$A_3$** | **. .** | **$A_m$** |
|---|---|---|---|---|---|
| **$O_1$** | x | x | x | . | |
| **$O_2$** | x | x | | . | |
| **$O_3$** | | x | x | . | |
| **. .** | . | . | . | . | |
| **$O_n$** | | | | | |

Adaptive ECA* aims to reduce to the size of this table by applying the mut-over operator on its objects and attributes. For example, if $A_1$ is similar to $A_3$, a crossover occurred on these attributes to generate $A_{13}$. The result of crossover can be a hypernym of both attributes based on WordNet. The similarity and relativity check were also applied to the objects *O*. Meanwhile, if $O_2$ and $O_3$ are related together, one of the attributes of both of them were mutated to find a common hypernym between them, and it can be presented as $O_{23}$. This relativity check was also applied to the pairs of attributes *A*. Table (4.7) depicts the result of Table (3.6) after applying adaptive ECA*.

**Table (4.7): Reduced formal context example after applying adaptive ECA***

| | **A13** | **A2** | **. .** | **Am** |
|---|---|---|---|---|
| **O1** | x | x | . | |
| **O23** | x | x | . | |
| **. .** | . | . | . | |
| **On** | | | | |

In FCA, tables are commonly represented with logical attributes by triplets. The table *(O, A, I)* was considered instead of the *triplet (O, A, I)*. The goal of FCA is to obtain two outputs from a given table. The first, called a concept lattice, is a partially ordered collection of an individual object and attribute clusters. The second one consists of formulas that describe specific attribute dependences true in the table. For this study, the second goal can be achieved using ready software solutions, such as Concept Explorer [127]. The components of adaptive ECA* are detailed as follows [128]:

**A. Initialisation:** The hypernym and hyponym depths were initialised. Moreover, the number of iterating the algorithm was initialised.

**B. Construction I:** Consider a formal context as *(O, A, I),* and each note structure of the context can be represented as *x*. Taking a set of the objects as $O = \{o_1, o_2, o_3, \ldots, o_n\}$, and a set of attributes as $A = \{a_1, a_2, a_3, \ldots., a_m\}$, objects can be presented as pairs $(o_i, o_{i+1})$, and also attributes can be represented as pairs $(a_j, a_{j+1})$. These representations are shown in Equation (4.18) and (4.19), respectively.

$$(o_i, o_j) = \sum_{i=1}^{n-1} \sum_{j=i+1}^{n} (o_i, o_j) \tag{4.18}$$

$$(a_i, a_j) = \sum_{i=1}^{m-1} \sum_{j=i+1}^{m} (a_i, a_j) \tag{4.19}$$

**C. Mut-over:** This operator consists of the recombination of crossover and mutation. The operation of mut-over is depicted in Equation (4.20).

$$(\mathrm{oi}, \mathrm{oj}) = \begin{cases} Crossover & o_i \text{ is similar to } o_j \\ Mutation & o_i \text{ relates to } o_j \\ Otherwise & \text{no operation} \end{cases} \tag{4.20}$$

1. Crossover: If two objects/attributes are similar to each other according to the WordNet, the crossover can be occurred on them to generate the most fitting general and hypernym synset for the close two objects/attributes in meaning. There are several methods to compute the similarities between the two terms. WordNet is one of the most common approaches for measuring knowledge-based similarities between terms. Referring to [129], WordNet is a winner approach for similarity measurement tasks as it uses lexical word alignment. Meantime, WordNet is used in this approach to determine how two terms are related together.

2. Mutation: If two objects/attributes are related to each other according to the WordNet, one or both of them can be mutated to find a common a hypernym between them. Their relationships can recognise the objects/attributes via the hypernym and hyponym depth.

3. If the objects/attributes are not either similar or relative to each other, no operations will be made on them.

**D. Construction II:** Deriving concept lattice from the formal context. After that, find the lattice invariants to check to what extent the resulting lattice is isomorphic with the original one.

**E. Evaluation:** evaluating the isomorphic level of the resulting lattice with the original one to terminate the algorithm or continue.

## 4.6 Experiments on Reducing Concept Hierarchy Construction

This section consists of two parts: the datasets used in the experiment, and the detailed setup of the experiment.

In order to ignore the uninteresting and some false pairs in the formed formal context, this study experimented to eliminate these word pairs and reduce the size of formal context accordingly. The aims of the experiment are to answer the following question in the lens of this experiment:

1. Can it reduce the size of the formal context without impacting the quality of the result?

2. Do we need an evolutionary algorithm with the aid of linguistic resources to reduce the size of the formal context keeping the quality of the resulting lattice?

3. Is adaptive ECA* with linguistic tools more effective in reducing the size of the formal context than the current framework for reducing the concept of lattice size?

### 4.6.1 Datasets

It is somewhat helpful to use well-understood and commonly used standard datasets so that the findings can be quickly evaluated. Nevertheless, most of the corpora datasets are prepared for NLP tasks. Wikipedia is a good source of well-organised written text corpora with a wide range of expertise. These articles are available online freely and conveniently. Based on the number of English articles that exist in Wikipedia, several articles were randomly chosen to be used as corpus datasets in this experiment. The sample size is determined from the confidence level, margin of error, population proportion, and population size. That means, 385 articles were needed to have a confidence level of 95% that the real value is within ±5% of the measured value. Table (4.8) presents the parameters related to the sample size of this study.

**Table (4.8): The parameters related to the sample of size of this experiment**

| **Parameters** | **Value** |
|---|---|
| Confidence level | 95% |
| Margin of error | 5% |
| Population proportion | 50% |
| Population size (number of English articles in Wikipedia) | 6090000 |

### 4.6.2 Experimental Setup

The overall steps of this study's proposed framework, including the experiment, are as follows:

**A. Input data:** 385 text corpora were taken from Wikipedia has different properties, such as length, and text cohort. The basic statistics of the free texts are calculated in Table (4.9).

**Table (4.9): Basic statistics of the corpus**

| **Basic statistics** | **Value** |
|---|---|
| Mean | 556.842 |
| Median | 490.000 |
| Sum | 214384.000 |
| Max | 2821.000 |
| Min | 172.000 |
| STDV P | 298.225 |
| STDV S | 298.613 |
| STDEVA | 298.613 |

**B. Word pair extraction:** The corpora were tagged in part-of-speech and then scanned to return a parsing tree per sentence. Next, the dependencies were derived from the parser trees. The pairs were subsequently lemmatised, the word pairs were suggested to be filtered and pruned, and then the formal context-dependent on the word pairs were constructed.

**C. FCA and concept lattice:** The formal contexts were formed from the word pairs to produce the concept lattices. Two categories of concept lattices were produced for evaluation as a result of implementing the experiment. The first concept of the lattice was formed directly from the text corpora without applying adaptive ECA*. In contrast, the latter one was constructed from the corpora after implementing the adaptive ECA* on it. These two concept lattices are labelled as follows:

1. Concept lattice 1: A concept lattice without applying any reduction method.
2. Concept lattice 2: A concept lattice after applying adaptive ECA*.

Figure (4.6) illustrates the mechanism used to derive the two above concept lattices, and concept hierarchies as well.

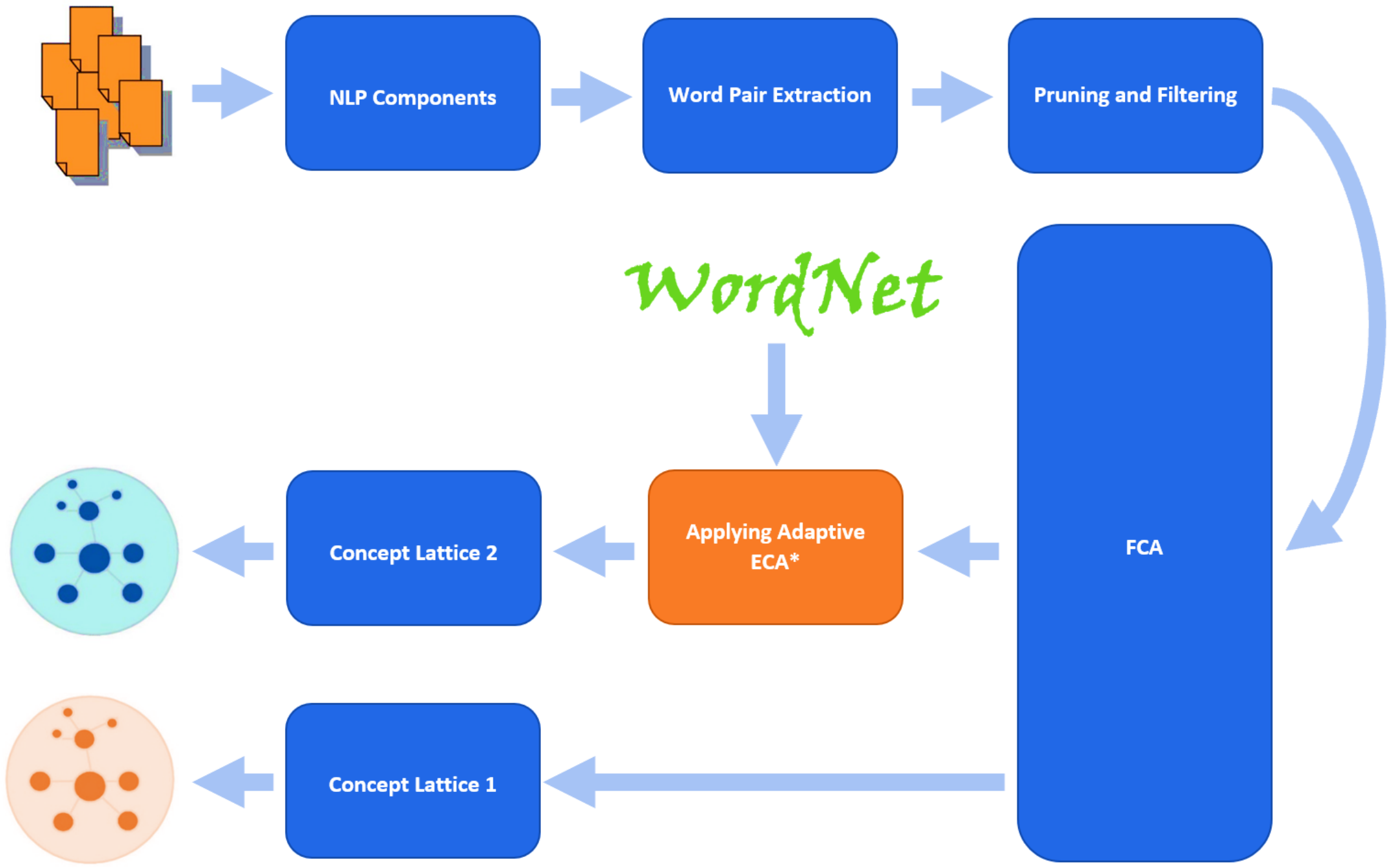

**Figure (4.6): A proposed framework for examining formal context size-reduction**

**D. Adaptive ECA*:** Once the concept lattices had been generated, adaptive ECA* was applied to them. The parameters of constructing the concept lattice 1, and 2 are presented in Table (4.10).

**Table (4.10): Parameters for deriving concept lattice 1, and 2**

| Concept lattice ID | Reduction method | Hypernym depth | Hyponym depth | Number of iterations |
|---|---|---|---|---|
| Concept lattice 1 | - | - | - | 1 |
| Concept lattice 2 | Adaptive ECA* | 4 | 4 | 30 |

**E. Concept lattice properties:** After deriving the concept lattices, some statistical data from each of them were recorded. These statistical data are some concepts, the number of edges, height, and width of lattices.

**F. Evaluating the results:** At the last step, the relationships between the two lattices were found via isomorphic and homeomorphism.

# Chapter Five: Evaluation Methods and Results Analysis

In this chapter, the results are presented, then directly analysed and discussed for readability purposes. Firstly, the results related to performance evaluation of BSA is argued, followed by the limitations and contributions of BSA. Secondly, the evaluation methods of the clustering results of ECA*, performance evaluation of the same algorithm, and a proposed performance ranking framework of ECA* are presented. Finally, the proposed algorithm applied to practical applications, such as concept hierarchies in ontology learning, are also used to reduce the size of formal context to construct concept hierarchies from free texts.

## 5.1 Result Analysis of BSA Performance Evaluation

This section discusses the results of the three tests described in the previous chapter on evaluating the performance of BSA against its counterparts.

In Test 1, all the algorithms successfully minimised the four optimisation problems: F3, F4, F5, and F8. In this test, the algorithms were run on the optimisation problems on their default search spaces and three different dimensions: Nvar1, 2, and 3. Nevertheless, the dimensions of the numerical optimisation problems did not affect the success rate of the algorithms to minimise the problems because the algorithms could minimise four problems in three variable dimensions. Moreover, the level of hardness of the minimised benchmark functions was relatively low. The minimum and maximum levels of the overall success of these problems were 41.17 and 59%, respectively. Consequently, none of the algorithms minimised the problems with high hardness scored in different variable dimensions. Meanwhile, F10 with an overall success score of 80.08% was not solved by any of the algorithms in any variable dimensions. This implies that a clear conclusion cannot be drawn regarding the success of BSA and other algorithms in minimising the optimisation problems based on their hardness score.

Therefore, there is no implicit correlation between the success of these algorithms and the difficulty level of minimising optimisation problems. A problem-based statistical comparison method was used to determine which of the algorithms used in the experiment could statistically solve the benchmark functions. This method used the computation time required by the algorithms to reach the global minimum as a result of 30 runs. In the experiment, BSA was compared with other algorithms by using the Wilcoxon signed-rank test by considering the significant statistical value (α) to be 0.05. Further, the null hypothesis (H0) for a specific benchmark problem defined in Equation (5.1) was considered. Tables (5.1), (5.2), and (5.3) list the algorithms that statistically gained better solutions compared to the other algorithms in Test 1 according to the Wilcoxon signed-rank test.

**Table (5.1): Statistical comparison to find the algorithm that provides the best solution for the solved benchmark function used in Test 1 (Nvar1) using two-sided Wilcoxon Signed-Rank Test (α=0.05)**

| **Problems** | **BSA vs DE** | | | | **BSA vs PSO** | | | | **BSA vs ABC** | | | | **BSA vs FF** | | | |
|---|---|---|---|---|---|---|---|---|---|---|---|---|---|---|---|---|
| | p-value | R+ | R- | Win | p-value | R+ | R- | Win | p-value | R+ | R- | Win | p-value | R+ | R- | Win |
| F3 | 0.0001 | 0 | 465 | + | 0.0004 | 49 | 329 | + | 0.0004 | 69 | 396 | + | 0.0001 | 463 | 2 | + |
| F4 | 0.0001 | 465 | 0 | + | 0.0003 | 399 | 66 | + | 0.6431 | 167 | 133 | - | <0.0001 | 465 | 0 | + |
| F5 | <0.0001 | 465 | 0 | + | 0.1094 | 311 | 154 | + | <0.0001 | 450 | 15 | + | <0.0001 | 463 | 2 | + |
| F8 | <0.0001 | 465 | 0 | + | <0.0001 | 465 | 0 | + | <0.0001 | 21 | 279 | + | 0.0006 | 347 | 59 | + |
| **+/=/-** | 4/0/0 | | | | 4/0/0 | | | | 3/0/1 | | | | 4/0/0 | | | |

**Table (5.2): Statistical comparison to find the algorithm that provides the best solution for the solved benchmark function used in Test 1 (Nvar2) using two-sided Wilcoxon Signed-Rank Test (α=0.05)**

| **Problems** | **BSA vs DE** | | | | **BSA vs PSO** | | | | **BSA vs ABC** | | | | **BSA vs FF** | | | |
|---|---|---|---|---|---|---|---|---|---|---|---|---|---|---|---|---|
| | p-value | R+ | R- | Win | p-value | R+ | R- | Win | p-value | R+ | R- | Win | p-value | R+ | R- | Win |
| F3 | <0.0001 | 450 | 15 | + | 0.0001 | 51 | 384 | + | <0.0001 | 0 | 465 | + | <0.0001 | 441 | 24 | + |
| F4 | <0.0001 | 465 | 0 | + | <0.0001 | 455 | 10 | + | <0.0001 | 465 | 0 | + | <0.0001 | 465 | 0 | + |
| F5 | <0.0001 | 463 | 2 | + | 0.3285 | 281 | 184 | + | 0.6263 | 208 | 257 | - | 0.0879 | 316 | 149 | + |
| F8 | <0.0001 | 465 | 0 | + | <0.0001 | 465 | 0 | + | <0.0001 | 3 | 432 | + | 0.0772 | 263 | 115 | + |
| +/=/1 | 4/0/0 | | | | 4/0/0 | | | | 3/0/1 | | | | 4/0/0 | | | |

**Table (5.3): Statistical comparison to find the algorithm that provides the best solution for the solved benchmark function used in Test 1 (Nvar3) using two-sided Wilcoxon Signed-Rank Test (α=0.05)**

| Problems | BSA vs DE | | | | BSA vs PSO | | | | BSA vs ABC | | | | BSA vs FF | | | |
|---|---|---|---|---|---|---|---|---|---|---|---|---|---|---|---|---|
| | p-value | R+ | R- | Win | p-value | R+ | R- | Win | p-value | R+ | R- | Win | p-value | R+ | R- | Win |
| F3 | <0.0001 | 447 | 18 | + | <0.0001 | 1 | 350 | + | <0.0001 | 1 | 464 | + | <0.0001 | 426 | 39 | + |
| F4 | <0.0001 | 465 | 0 | + | <0.0001 | 421 | 44 | + | 0.9906 | 190 | 188 | - | <0.0001 | 465 | 0 | + |
| F5 | 0.0879 | 316 | 149 | + | <0.0001 | 0 | 465 | + | <0.0001 | 0 | 465 | + | <0.0001 | 2 | 463 | + |
| F8 | <0.0001 | 465 | 0 | + | <0.0001 | 465 | 0 | + | <0.0001 | 0 | 406 | + | <0.0001 | 12 | 339 | + |
| +/=/- | 4/0/0 | | | | 4/0/0 | | | | 3/0/1 | | | | 4/0/0 | | | |

In contrast, all the algorithms in Test 2 solved some of the optimisation functions from the highest hardness score of difficulty to the lowest. In this test, the algorithms were run on the optimisation problems in two dimensions and three different search spaces: R1, 2, and 3. Unlike Test 1, the variety of the search spaces affected the success rates of the algorithms to minimise the problems because the algorithms could have minimised a different number of problems in each search space. For example, out of 16, the number of optimisation problems minimised by all the algorithms in R1, 2, and 3 was 11, 9, and 8, respectively. Further, the level of difficulty of the solved benchmark functions varied from high to low. For example, the minimum and maximum levels of the overall success of these problems in search space R3 were 6.08 and 82.75%, respectively.

Furthermore, none of the algorithms could minimise the problems with a high score of hardness in search space R3. F16 with overall success score of 62.67% could not be solved by any of the algorithms in search space R3. Similarly, F7 with overall success score of 4.92% could not be minimised by any of the algorithms in the same search space. This implies that a clear conclusion cannot be drawn on the success of BSA and other algorithms in minimising optimisation problems based on their hardness scores. Therefore, there is no implicit correlation between the success of these algorithms and the difficulty level of minimising optimisation problems. Similar to Test 1, Test 2 was about comparing BSA with other algorithms by using the Wilcoxon signed-rank test by considering the significant statistical value (α) as 0.05. Here, the considered null hypothesis (H0) for a specific benchmark problem is defined in Equation (5.1). Tables (5.4), (5.5), and (5.6) list the algorithms that statistically gained better solutions compared to the other algorithms in Test 2 according to the Wilcoxon signed-rank test.

**Table (5.4): Statistical comparison to find the algorithm that provides the best solution for the solved benchmark function used in Test 2 (R1) using two-sided Wilcoxon Signed-Rank Test (α=0.05)**

| **Problem** | **BSA vs DE** | | | | **BSA vs PSO** | | | | **BSA vs ABC** | | | | **BSA vs FF** | | | |
|---|---|---|---|---|---|---|---|---|---|---|---|---|---|---|---|---|
| | p-value | R+ | R- | Win | p-value | R+ | R- | Win | p-value | R+ | R- | Win | p-value | R+ | R- | Win |
| F2 | <0.0001 | 465 | 0 | + | <0.0001 | 406 | 0 | + | <0.0001 | 465 | 0 | + | <0.0001 | 465 | 0 | + |
| F4 | 0.0027 | 290 | 61 | + | <0.0001 | 465 | 0 | + | 0.1780 | 198 | 102 | - | 0.0004 | 44 | 307 | + |
| F5 | <0.0001 | 411 | 54 | + | <0.0001 | 433 | 32 | + | <0.0001 | 460 | 5 | + | <0.0001 | 450 | 5 | + |
| F6 | <0.0001 | 465 | 0 | + | 0.0137 | 114 | 351 | + | <0.0001 | 465 | 0 | + | 0.7611 | 217 | 248 | - |
| F8 | <0.0001 | 14 | 364 | + | <0.0001 | 465 | 0 | + | <0.0001 | 325 | 0 | + | <0.0001 | 371 | 7 | + |
| F9 | <0.0001 | 351 | 0 | + | 0.0001 | 105 | 0 | + | <0.0001 | 321 | 30 | + | <0.0001 | 435 | 0 | + |
| F11 | <0.0001 | 465 | 0 | + | <0.0001 | 462 | 3 | + | <0.0001 | 374 | 4 | + | <0.0001 | 322 | 3 | + |
| F12 | 0.7024 | 114 | 139 | - | <0.0001 | 465 | 0 | + | 0.0005 | 326 | 52 | + | 0.0179 | 68 | 232 | + |
| F13 | <0.0001 | 435 | 0 | + | <0.0001 | 325 | 0 | + | <0.0001 | 465 | 0 | + | <0.0001 | 377 | 1 | + |
| F14 | <0.0001 | 0 | 210 | + | <0.0001 | 465 | 0 | + | <0.0001 | 15 | 285 | + | <0.0001 | 0 | 325 | + |
| F15 | <0.0001 | 465 | 0 | + | <0.0001 | 465 | 0 | + | <0.0001 | 465 | 0 | + | <0.0001 | 465 | 0 | + |
| **+/=/-** | 10/0/1 | | | | 11/0/0 | | | | 10/0/1 | | | | 10/0/1 | | | |

**Table (5.5): Statistical comparison to find the algorithm that provides the best solution for the solved benchmark function used in Test 2 (R2) using two-sided Wilcoxon Signed-Rank Test (α=0.05)**

| **Problem** | **BSA vs DE** | | | | **BSA vs PSO** | | | | **BSA vs ABC** | | | | **BSA vs FF** | | | |
|---|---|---|---|---|---|---|---|---|---|---|---|---|---|---|---|---|
| | **p-val** | **R+** | **R-** | **Win** | **p-val** | **R+** | **R-** | **Win** | **p-val** | **R+** | **R-** | **Win** | **p-val** | **R+** | **R-** | **Win** |
| F2 | <0.0001 | 407 | 28 | + | <0.0001 | 377 | 1 | + | <0.0001 | 457 | 8 | + | <0.0001 | 378 | 0 | + |
| F4 | <0.0001 | 403 | 3 | + | <0.0001 | 231 | 0 | + | <0.0001 | 433 | 2 | + | <0.0001 | 325 | 0 | + |
| F8 | 0.9789 | 164 | 161 | - | <0.0001 | 465 | 0 | + | <0.0001 | 325 | 0 | + | <0.0001 | 465 | 0 | + |
| F9 | <0.0001 | 465 | 0 | + | <0.0001 | 267 | 0 | + | <0.0001 | 325 | 0 | + | <0.0001 | 435 | 0 | + |
| F11 | <0.0001 | 646 | 1 | + | <0.0001 | 432 | 3 | + | <0.0001 | 351 | 0 | + | <0.0001 | 378 | 0 | + |
| F12 | <0.0001 | 378 | 0 | + | <0.0001 | 465 | 0 | + | <0.0001 | 153 | 0 | + | <0.0001 | 374 | 4 | + |
| F13 | <0.0001 | 464 | 1 | + | <0.0001 | 378 | 0 | + | <0.0001 | 378 | 0 | + | <0.0001 | 465 | 0 | + |
| F14 | <0.0001 | 13 | 365 | + | <0.0001 | 465 | 0 | + | <0.0001 | 299 | 1 | + | <0.0001 | 405 | 1 | + |
| F15 | <0.0001 | 465 | 0 | + | <0.0001 | 465 | 0 | + | <0.0001 | 465 | 0 | + | <0.0001 | 465 | 0 | + |
| +/=/- | 8/0/1 | | | | 9/0/0 | | | | 9/0/0 | | | | 9/0/0 | | | |

**Table (5.6): Statistical comparison to find the algorithm that provides the best solution for the solved benchmark function used in Test 2 (R3) using two-sided Wilcoxon Signed-Rank Test (α=0.05)**

| Problem | BSA vs DE | | | | BSA vs PSO | | | | BSA vs ABC | | | | BSA vs FF | | | |
|---|---|---|---|---|---|---|---|---|---|---|---|---|---|---|---|---|
| | p-value | R+ | R- | Win | p-value | R+ | R- | Win | p-value | R+ | R- | Win | p-value | R+ | R- | Win |
| F2 | <0.0001 | 433 | 32 | + | <0.0001 | 406 | 0 | + | <0.0001 | 450 | 15 | + | <0.0001 | 465 | 0 | + |
| F4 | <0.0001 | 406 | 0 | + | <0.0001 | 351 | 0 | + | <0.0001 | 378 | 0 | + | <0.0001 | 351 | 0 | + |
| F8 | 0.0027 | 290 | 61 | + | <0.0001 | 465 | 0 | + | <0.0001 | 325 | 0 | + | <0.0001 | 276 | 0 | + |
| F9 | <0.0001 | 6 | 345 | + | <0.0001 | 300 | 0 | + | <0.0001 | 325 | 0 | + | <0.0001 | 276 | 0 | + |
| F11 | <0.0001 | 422 | 43 | + | 0.2286 | 292 | 173 | + | <0.0001 | 604 | 0 | + | <0.0001 | 435 | 0 | + |
| F12 | <0.0001 | 350 | 1 | + | <0.0001 | 465 | 0 | + | 0.0156 | 28 | 0 | + | <0.0001 | 405 | 1 | + |
| F13 | <0.0001 | 377 | 1 | + | <0.0001 | 406 | 0 | + | <0.0001 | 351 | 0 | + | <0.0001 | 465 | 0 | + |
| F14 | 0.0118 | 94 | 312 | + | <0.0001 | 465 | 0 | + | <0.0001 | 431 | 4 | + | <0.0001 | 406 | 0 | + |
| +/=/- | 8/0/0 | | | | 8/0/0 | | | | 8/0/0 | | | | 8/0/0 | | | |

In Tables (5.1) to (5.6), three sings are used as described below:

- '-' represents the cases of rejecting the null hypothesis and displaying BSA as statistically inferior performance in the statistical comparison of the problems;
- '+' represents that the cases of rejecting the null hypothesis and displaying BSA as statistically superior performance;
- '=' represents the cases with no statistical difference between the two comparison algorithms in the evaluation of the success of minimising the problems. The last rows of Tables (4.1) to (4.6) present the summation in the (+/=/-) format for these statistically significant cases: '+', '=', and '-' in the pair-wise problem-based statistical comparisons of the algorithms. Based on the examination of the (+/=/-) values in Tests 1 and 2, it can be concluded that BSA showed superior performance statistically than the other comparison algorithms in minimising numerical optimisation problems.

Overall, the results obtained from Tests 1, and 2 reveal that BSA is relatively more successful in solving numerical optimisation problems with different levels of hardness, variable size, and search spaces. However, none of the algorithms could minimise all the 16 benchmark functions successfully.

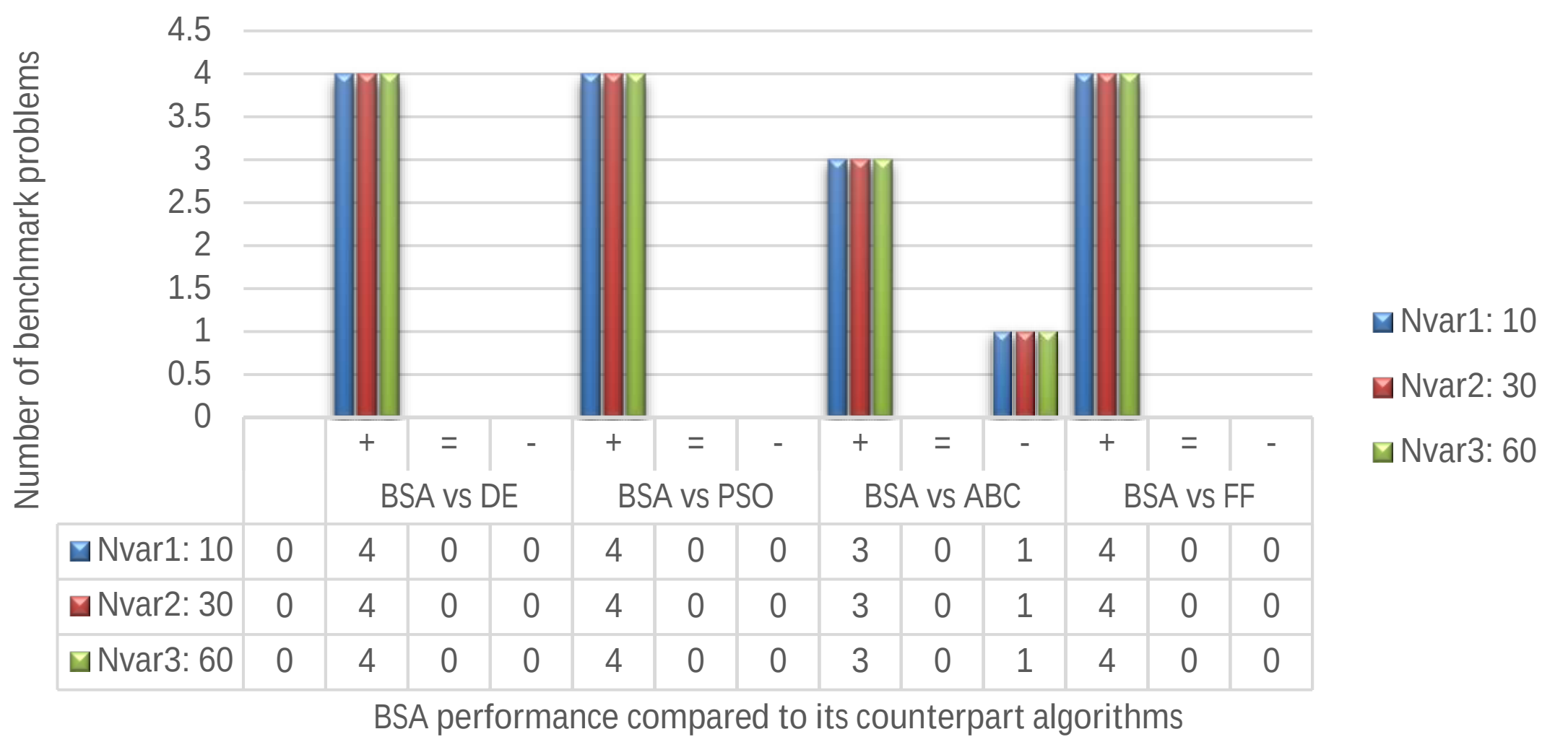

| | | BSA vs DE | | | BSA vs PSO | | | BSA vs ABC | | | BSA vs FF | | |
|---|---|---|---|---|---|---|---|---|---|---|---|---|---|
| | | + | = | - | + | = | - | + | = | - | + | = | - |
| Nvar1: 10 | 0 | 4 | 0 | 0 | 4 | 0 | 0 | 3 | 0 | 1 | 4 | 0 | 0 |
| Nvar2: 30 | 0 | 4 | 0 | 0 | 4 | 0 | 0 | 3 | 0 | 1 | 4 | 0 | 0 |
| Nvar3: 60 | 0 | 4 | 0 | 0 | 4 | 0 | 0 | 3 | 0 | 1 | 4 | 0 | 0 |

**Figure (5.1): A graphical representation of Test 1 results**

The results listed in Tables (5.1), (5.2), and (5.3) are graphically represented in Figure (5.1). Likewise, the results in Tables (5.4), (5.5), and (5.6) are graphically represented in Figure (5.2).

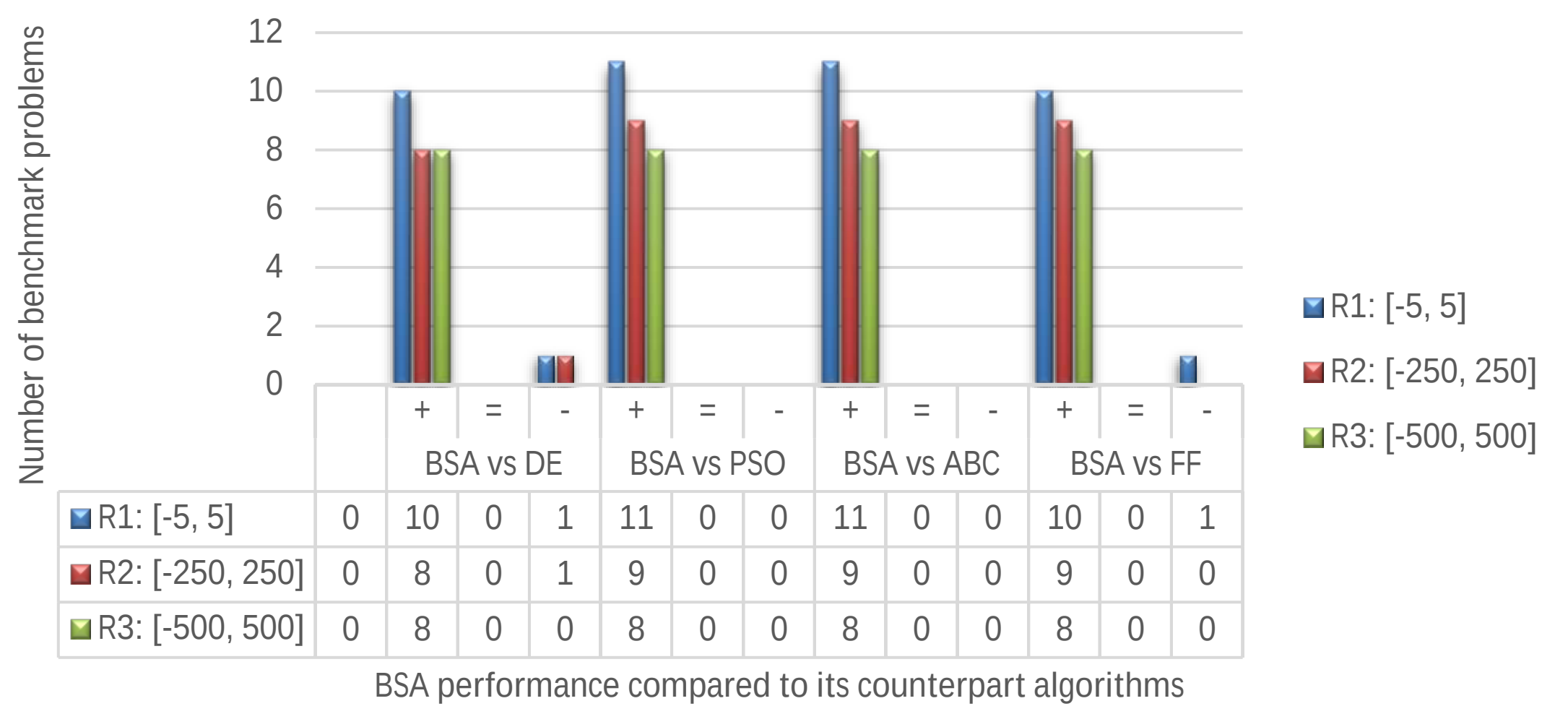

| | | BSA vs DE | | | BSA vs PSO | | | BSA vs ABC | | | BSA vs FF | | |
|---|---|---|---|---|---|---|---|---|---|---|---|---|---|
| | | + | = | - | + | = | - | + | = | - | + | = | - |
| R1: [-5, 5] | 0 | 10 | 0 | 1 | 11 | 0 | 0 | 11 | 0 | 0 | 10 | 0 | 1 |
| R2: [-250, 250] | 0 | 8 | 0 | 1 | 9 | 0 | 0 | 9 | 0 | 0 | 9 | 0 | 0 |
| R3: [-500, 500] | 0 | 8 | 0 | 0 | 8 | 0 | 0 | 8 | 0 | 0 | 8 | 0 | 0 |

**Figure (5.2): A graphical representation of Test 2 results**

Referring to Test 1 and 2, none of the algorithms was successful in solving all benchmark problems. Concerning Test 3, the ratio of successful minimisation of all sixteen benchmark functions varies in Nvar 1, 2, and 3 with default search space, and two variable dimensions with three different search spaces (R1, R2, and R3). As a more illustration, Figure (5.3) depicts the success and failure ratio for minimising the sixteen benchmark functions in Test 1.

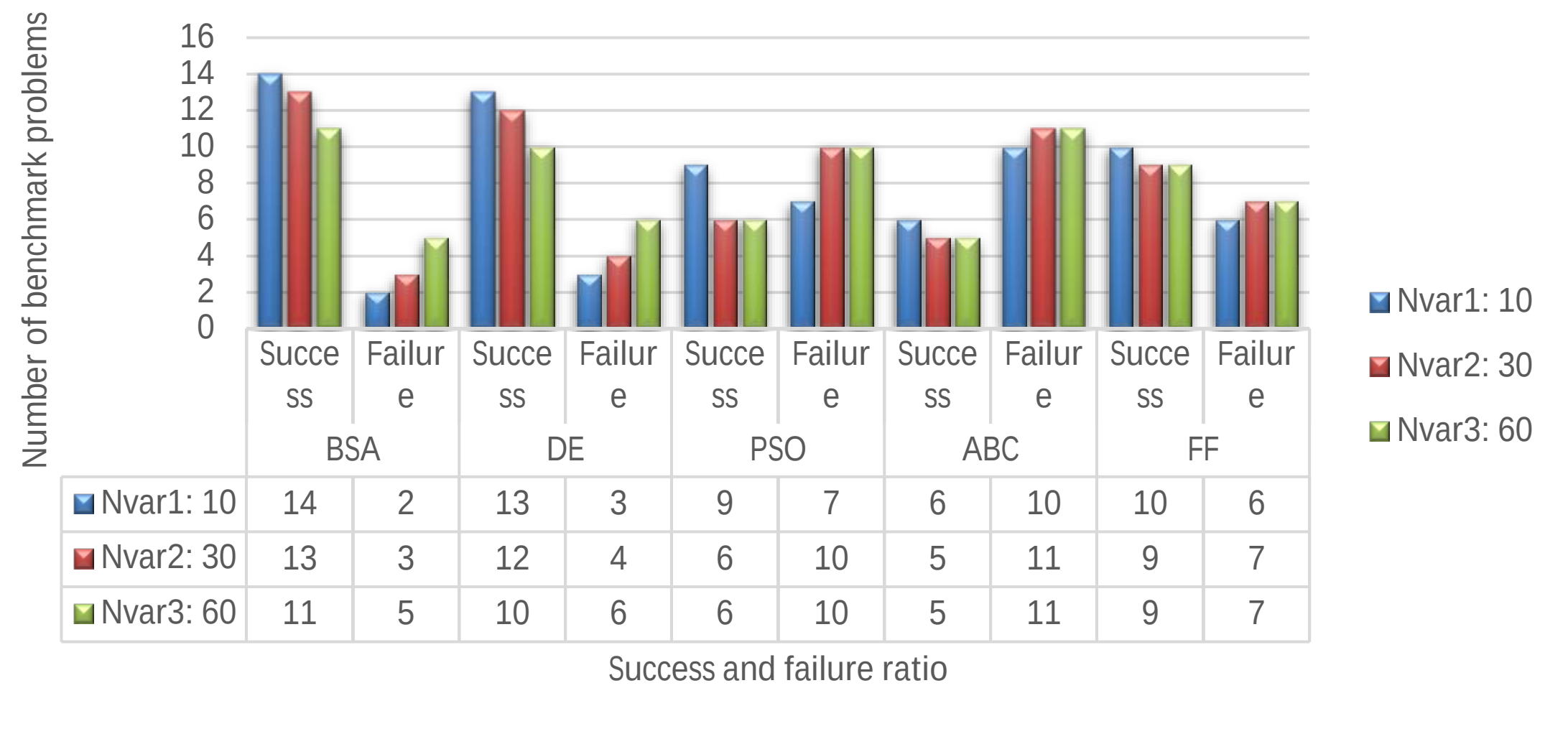

**Figure (5.3): The success and failure ratio for minimising the sixteen benchmark functions in Test 1**

Regarding Tests 1 and 2, it was seen that none of the algorithms was successful in solving all the benchmark problems. Regarding Test 3, the ratio of successful minimisation of all sixteen benchmark functions varied in Nvar1, 2, and 3 with the default search space, and two variable dimensions with three different search spaces (R1, R2, and R3). Figure (5.4) depicts the success and failure ratios in minimising the 16 benchmark functions in Tests 2.

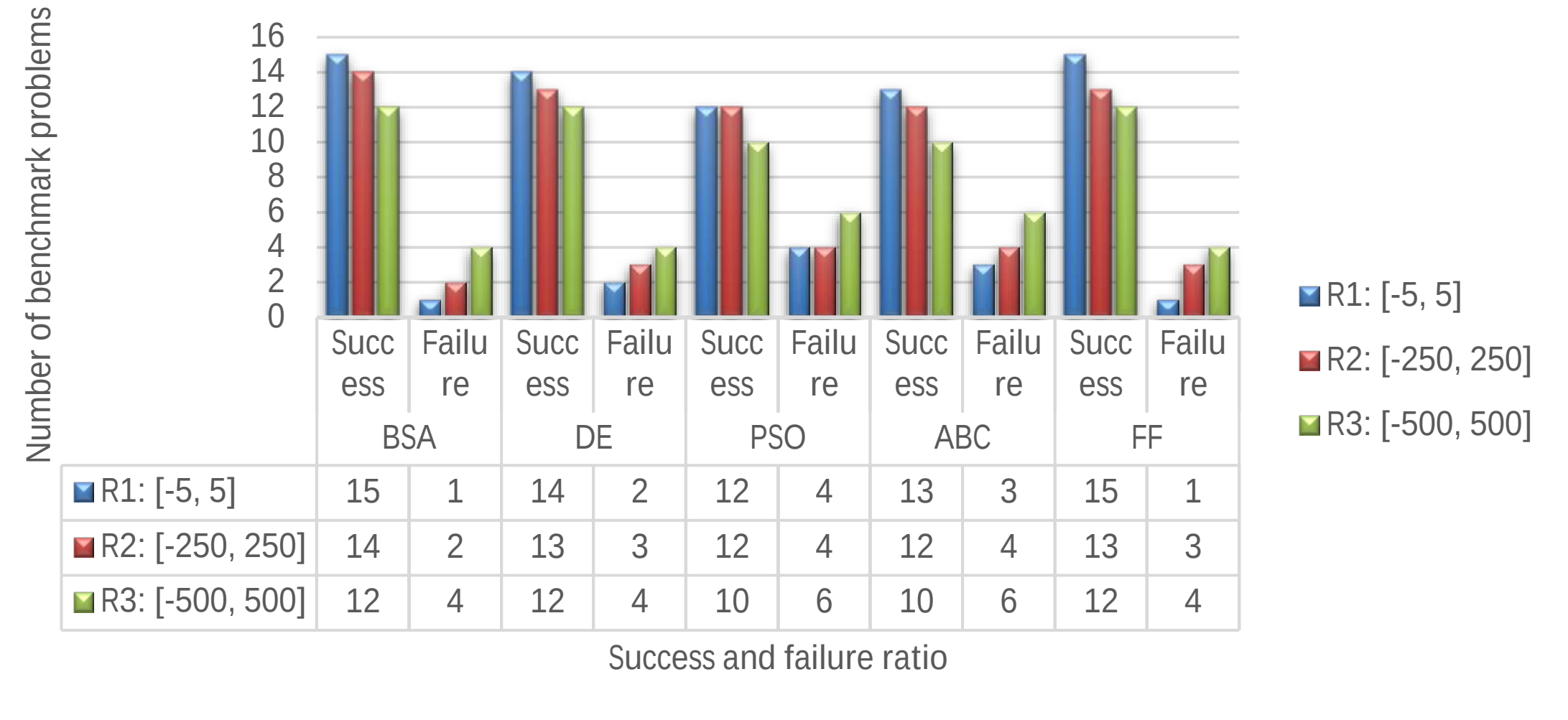

**Figure (5.4): The success and failure ratio in minimising the sixteen benchmark functions in Test 2**

Test 3 revealed that BSA was marginally the most successful algorithm in minimizing the maximum number of the given 16 benchmark functions, whereas ABC was the least successful in minimising the same. It also revealed that BSA could minimise the maximum number of optimisation problems; however, the minimum ratio of successful minimisation of F1-F16 functions in Nvar1, 2, 3 variables with the default search space were seen for ABC, followed by PSO. Whereas the maximum ratio of successful minimisation of these optimisation problems with variable size of two with three different search spaces (R1, 2, and 3) was seen for both ABC and PSO, followed by DE.

## 5.2 Evaluation of BSA

The literature discusses the advantages and disadvantages of BSA along with those of the other meta-heuristic algorithms. Certain factors, which could be responsible for the higher success of BSA compared to other EAs, are as follows:

1. The trial populations are produced very efficiently in each generation by using BSA's crossover and mutation operators.
2. BSA balances the relationship between exploration (global search) and exploitation (local search) appropriately owing to the control parameter, *F*. This parameter controls the direction of the local and global searches in an efficient and balanced manner. This balance can produce a wide range of numerical values necessary for the local and global searches. An extensive range of numerical values is produced for the global search, whereas a small range is produced for the local search. This helps BSA to solve a variety of problems.

3. The search direction of BSA is calculated by using the historical population *(oldP)*. In other words, the historical populations that are used in more capable individuals are relative to those that are used in older generations. This operation helps BSA to generate the trial individuals efficiently.

4. BSA's crossover is non-uniform and complex structure process compared to DE's crossover. In each generation, the crossover operator in BSA creates new trial individuals to enhance BSA's ability to deal with the problems.

5. Population diversity can be effectively achieved in BSA by a boundary control mechanism to ensure efficient searches.

6. Unlike other meta-heuristic algorithms, BSA uses historical and current populations; therefore, it is called a dual-population algorithm.

7. The simplicity of BSA may facilitate its usage by researchers and professionals.

Presumably, no clear and precise drawbacks of BSA were explicitly addressed in the literature. Nonetheless, scholars have attempted to modify and hybridise the standard version of BSA as a compromise to further improve its efficiency and effectiveness to maximise its application domain. These attempts imply that researchers implicitly pointed out some drawbacks of BSA. Similar to the other meta-heuristic algorithms, BSA has some drawbacks that are listed below:

1. BSA requires memory space for storing historical populations. It is a dual-population algorithm because it utilises both historical and current populations. In each iteration, the old populations are stored on the stack for use in the next iterations. The notion of employing a lexical database as the historical population of BSA is entirely supported to store its historical population. This lexical database is beneficial to reduce the memory size.

2. Setting the initial population in BSA entirely depends on the randomisation. One could simply mitigate the un-typical random nature of BSA by taking an average run of the algorithm. Alternatively, this study could effectively deal with the un-typical random efforts of BSA by using the initial population of another nature-inspired algorithm. BSA and cuckoo search (CS) are both population-based algorithms [130]; therefore, the initial population of BSA can be set using CS. This helps the new proposed algorithm to reach the convergence faster than the standard BSA.

3. In the literature, it is stated that BSA could be improved further [26], [131]–[137]. The improvements can be related to efficiency, effectiveness, computing time, speed of convergence, and improvement of local and global searches. The reasons behind these issues are complex. One possible reason can be the control parameters of BSA. Another reason could be related to the reality that BSA may appropriately need to find the best relationship between the local (exploitation) and global searches (exploration) that will suit each type of application to converge the best solution quickly.
4. This study [131] claims three shortcomings of BSA: difficulty in adjusting control parameter (F), lack of learning from optimal individuals, and easily falling into local optima. The same research states that BSA is effective in exploration (global optima), whereas it is relatively weak in exploitation (local optima). On this basis, the convergence speed may be affected. Based on our knowledge, it is suggested that modifying BSA to improve its convergence speed and other drawbacks by redesigning the amplitude control factor (F) based on Levy flights and dynamic penalty to balance the local and global optima. This balance decreases the number of iterations and speeds up the convergence of BSA. In a recent study [58], F is designed to balance the exploration and exploitation of the algorithm. It is believed that learning knowledge from the optimal individuals in BSA is not an intricate task. Recently, a new algorithm based on BSA was proposed; namely, each individual learns knowledge by utilising the global and local best information and the better historical information [58].

Comparing with other meta-heuristic algorithms such as GA and PSO, BSA has not been used many applications and a variety of fields. BSA has mostly been used in engineering and information technology but not in the fields of text mining, natural language processing, Semantic Web, health, and medical sciences. The limited popularity of BSA compared with GA and PSO is not related to any weaknesses of BSA. Instead, it can probably be attributed to the newness of BSA. Time may be a factor in revealing the uses and expansions of BSA.

## 5.3 Result Analysis of ECA*

This section presents the evaluation methods of the clustering results of ECA*, performance evaluation of the same algorithm, and a proposed performance ranking framework of ECA*.

### 5.3.1 Evaluation of Clustering Results of ECA*

This thesis used three internal measures and three external measures to evaluate ECA* against its counterpart techniques. For internal measures, the sum of squared error (SSE), mean squared error (MSE), and approximation ratio (ε- ratio) were used. For external measures, the centroid index (CI), centroid similarity index (CSI), and normalised mutual information (NMI) were utilised.

**1. Internal measures:** The measures of internal performance evaluation rely only on the datasets themselves. These measures are all different from each other for the same objective function. The sum of squared error (SSE) was used to measure the squared differences between each observation with its cluster centroid and the variation with a cluster. If all cases with a cluster are similar, the SEE would be equal to 0. That is, the less the value of SSE represents the better work of the algorithms. For example, if one algorithm gives SSE=5.54 and another SSE=9.08, then it can be presumed that the former algorithm works better. The formula of SSE is shown in Equation (5.1) below [51]:

$$\sum_{i=1}^{N}(x_i - c_i)^2 \quad (5.1)$$

Where $x_i$ is the observation data and $c_j$ is its nearest cluster centroid.

Besides, mean squared error (MSE) was used to measure the average of squared error, and the average squared difference between the actual value and the estimated values. As it is used in [51], a normalised mean squared error (nMSE) was used in this study as it is shown in Equation (5.2):

$$nMSE = \frac{SSE}{N.D} \quad (5.2)$$

Where:

*SSE*: the sum of squared error.

*N*: number of populations.

*D*: number of attributes in the dataset,

Nevertheless, the use of objective function values, such as SSE and nMSE could not say much about the results. Alternatively, the approximation ratio (*ε- ratio*) was used to evaluate the results with the theoretical results achieved for approximation techniques. The calculation of *ε- ratio* is presented in Equation (5.3) [51].

$$\varepsilon - ratio = \frac{SSE - SSE_{opt}}{SSE_{opt}} \tag{5.3}$$

To calculate $SSE_{opt}$, ground centroids could be used, but their locations can vary from the optimal locations used for minimising SSE. Meanwhile, it is hard to find the $SSE_{opt}$ for A, S, Birch, Shape, G2, and dim datasets in the literature to be used in this study. The results of an algorithm on a few datasets were used in [138], but it is not sufficient to be used in this research. It is claimed that SSE should be as close as zero [51]. Minimising SSE to zero is hard in both two and multi-dimensional datasets [139], yet it is typically not necessary to find the optimal value of SSE. For this reason, it is assumed that the optimal value of SSE is 0.001.

**2. External measures:** The measures of external performance evaluation reply on the data that was previously obtained. three main external measures were used in this evaluation. As our primary measure of success, this study firstly used the centroid index (CI) to count how many clusters are missing their centroids, or how many clusters have more than one centroid. The higher value of CI is the less number of correct cluster centroids [140]. For example, if CI= 0, the result of clusters is correct, and the algorithms could solve the problem correctly. If CI > 0, the algorithm could not solve the problem, either some clusters are missing, or a cluster may have more than one centroid. Since CI measures only the cluster level differences, the centroid similarity index (CSI) was also used to measure point level differences in the matched clusters and calculate the proportional number of identical points between the matched clusters. Accordingly, this measure provides a more precise result at the cost of interpreting the loss of value intuitively. Normalised mutual information (NMI) was also used to give more details on the point level differences between the matched clusters. In other words, NMI is a normalisation of the mutual information between the cluster centroids and ground truth centroids of the same matched clusters. NMI scales the results between 1 (perfect correlation between the matched clusters), and 0 (no mutual information between the matched clusters). MATLAB implementation of MNI can be found in [141].

### 5.3.2 Performance Evaluation of ECA*

The results of ECA* are summarised in Table (5.7). It shows the ECA* implementation over 32 datasets. The evaluation measures are based on cluster quality (CI, CSI, and NMI), an objective function (SSE, nMSE, and ε- ratio). In most of the cases, ECA* is the sole success algorithm among its counterparts. Table (5.7) shows the cluster quality and objective function measures for ECA* for 30 run average.

**Table (5.7): Cluster quality and objective function measures for ECA* for 30 run average**

| Datasets | Cluster quality | | | Objective function | | |
|---|---|---|---|---|---|---|
| | CI | CSI | NMI | SSE | nMSE | ε- ratio |
| S1 | 0 | 0.994 | 1.000 | 9.093E+12 | 9.093E+08 | 9.093E+15 |
| S2 | 0 | 0.977 | 1.000 | 1.420E+13 | 1.420E+09 | 1.420E+16 |
| S3 | 0 | 0.9809 | 1.000 | 1.254E+13 | 1.254E+09 | 1.254E+16 |
| S4 | 0 | 0.872 | 1.000 | 9.103E+12 | 9.103E+08 | 9.103E+15 |
| A1 | 0 | 1 | 1.000 | 1.215E+10 | 2.026E+06 | 1.215E+13 |
| A2 | 0 | 0.9785 | 1.000 | 7.103E+09 | 1.184E+06 | 7.103E+12 |
| A3 | 0 | 0.9995 | 1.000 | 2.949E+10 | 4.915E+06 | 2.949E+13 |
| Birch1 | 0 | 0.999 | 0.989 | 7.007E+09 | 3.504E+04 | 7.007E+12 |
| Un-balance | 0 | 1 | 1.000 | 2.145E+11 | 1.650E+07 | 2.145E+14 |
| Aggregation | 0 | 1 | 0.967 | 1.024E+04 | 6.499E+00 | 1.024E+07 |
| Compound | 0 | 1 | 1.000 | 4.197E+03 | 5.259E+00 | 4.197E+06 |
| Path-based | 0 | 1 | 0.923 | 4.615E+03 | 7.691E+00 | 4.615E+06 |
| D31 | 0 | 1 | 0.838 | 3.495E+03 | 5.637E-01 | 3.495E+06 |
| R15 | 0 | 1 | 0.653 | 1.092E+02 | 9.099E-02 | 1.092E+05 |
| Jain | 0 | 1 | 1.000 | 1.493E+04 | 2.001E+01 | 1.493E+07 |
| Flame | 0 | 1 | 0.965 | 3.302E+03 | 6.880E+00 | 3.302E+06 |
| Dim-32 | 0 | 0.999 | 0.807 | 1.618E+05 | 4.938E+00 | 1.618E+08 |
| Dim-64 | 0 | 0.915 | 0.708 | 1.814E+06 | 4.324E-01 | 2.834E+07 |
| Dim-128 | 0 | 0.907 | 0.613 | 1.958E+04 | 1.494E-01 | 1.958E+07 |
| Dim-256 | 0 | 0.974 | 0.475 | 1.255E+04 | 4.789E-02 | 1.255E+07 |
| Dim-512 | 0 | 0.871 | 0.330 | 2.969E+05 | 5.663E-01 | 2.969E+08 |
| Dim-1024 | 0 | 0.934 | 0.283 | 1.992E+05 | 1.900E-01 | 1.992E+08 |
| G2-16-10 | 0 | 1 | 0.715 | 2.048E+05 | 6.250E+00 | 3.259E+11 |
| G2-16-30 | 0 | 1 | 0.613 | 1.825E+06 | 5.569E+01 | 1.825E+09 |
| G2-16-60 | 0 | 0.998 | 0.571 | 1.045E+07 | 3.190E+02 | 1.045E+10 |
| G2-16-80 | 0 | 0.997 | 0.505 | 1.701E+07 | 5.192E+02 | 1.701E+10 |
| G2-16-100 | 0 | 0.999 | 0.566 | 3.259E+08 | 9.947E+03 | 3.879E+11 |
| G2-1024-10 | 0 | 1 | 0.702 | 2.097E+07 | 1.000E+01 | 2.097E+10 |
| G2-1024-30 | 0 | 0.999 | 0.584 | 2.086E+08 | 9.947E+01 | 2.086E+11 |
| G2-1024-60 | 0 | 0.998 | 0.527 | 8.074E+08 | 3.850E+02 | 8.074E+11 |
| G2-1024-80 | 0 | 0.997 | 0.506 | 3.040E+09 | 1.450E+03 | 3.040E+12 |
| G2-1024-100 | 0 | 0.996 | 0.498 | 2.221E+09 | 1.059E+04 | 2.221E+13 |

Both Figures (5.5) and (5.6) elucidate the results of Table (5.7). Figure (5.5) illustrates the cluster quality measured by external measures (CI, CSI, and NMI) for ECA* compared with its counterpart algorithms for 30 run average. Differently, Figure (5.6) demonstrates the cluster quality measured by internal measures (SSE, nMSE, and ε-ratio) for ECA* compared with its counterpart algorithms for 30 run average.

Extensively, Figure (5.5) shows the average values, not the scores, of the external evaluation criteria (CI, CSI, and NMI) for 32 numerical benchmark datasets. The clustering quality of ECA*, KM, KM++, EM, LVQ, and GENCLUST++ were evaluated to assess the contribution of ECA*. For CI measure. All the algorithms, except GENCLUST++, were successful in determining the right number of clusters for the used 32 datasets. In the meantime, GENCLUST++ is successful in most of the cases. Exceptionally, it was not successful for finding the right number of centroids indexes in A3, Compound, Dim-32, Dim-128, Dim-256, Dim-512, Dim-1024, and G2-1024-80. GENCLUST++ was more successful than in two-dimensional clustering benchmark datasets in comparison with multi-dimensional clustering benchmark datasets. In the comparison of CSI measure for ECA* with its counterpart algorithms. After ECA*, KM, EM, KM++, LQV, and GENGLUST++. In a few cases, the algorithms measured 1 or approximately 1 for CSI value.

Overall, ECA* was successful in having the optimal value of CSI for all the datasets, except S4. Alongside with its success, ECA*'s counterpart algorithms were successful in some datasets. Mainly, KM is successful in almost of shape datasets, including Aggregation, Compound, Path-based, D31, R15, Jain, and Flame, and three of the multi-dimensional clustering benchmark datasets, including G2-16-10, G2-16-30, and G2-1024-100. After that, EM is the third successful technique, in which a winner in six datasets (S2, S4, Compound, R15, Jain, and G2-1024-100), followed by KM++ which is the fourth winner in four datasets (R15, Dim-256, Dim-256, and G2-1024-100). Meanwhile, LVQ and GENCLUST++ are the least successful winner in only two datasets. Lastly, for the cluster quality measured by NMI for ECA* and its competitive algorithms for all the used datasets, ECA* cluster centroid results have the best correlation between its cluster centroids and ground truth centroids of the same matched clusters. Afterward, KM has relatively good relations of its cluster centroids with the ground truth centroids of the same matched clusters, particularly, for the two-dimensional benchmark datasets. In terms of this correlation, KM is followed by KM++, EM, and LVQ. It needs to mention that GENCLUST++ records the least correlation for almost all the current datasets.

Generally, the results show that ECA* mostly overcomes its existing algorithms for cluster quality evaluation criteria. Correctly, ECA*, KM, KM++, EM, and LVQ perform well for determining the correct number of centroid indexes for all the 32 datasets, but GENCLUS++ does not find the right number of centroid indexes in all cases. Regarding CSI, ECA* outperformed its counterpart techniques. After that, EM performed better than the other four algorithms, followed by LVQ, KM++, KM, and GENCLUST++, respectively. For evaluating these techniques based on normalised mutual information (NMI), both ECA* and KM performed pretty much the same, but ECA* slightly outperformed KM for finding a better quality of clusters. After these two techniques, KM++, EM, and LVQ perform quite similar in providing better cluster quality. Finally, GENCLUST++ was the least performing algorithm among its counterparts.

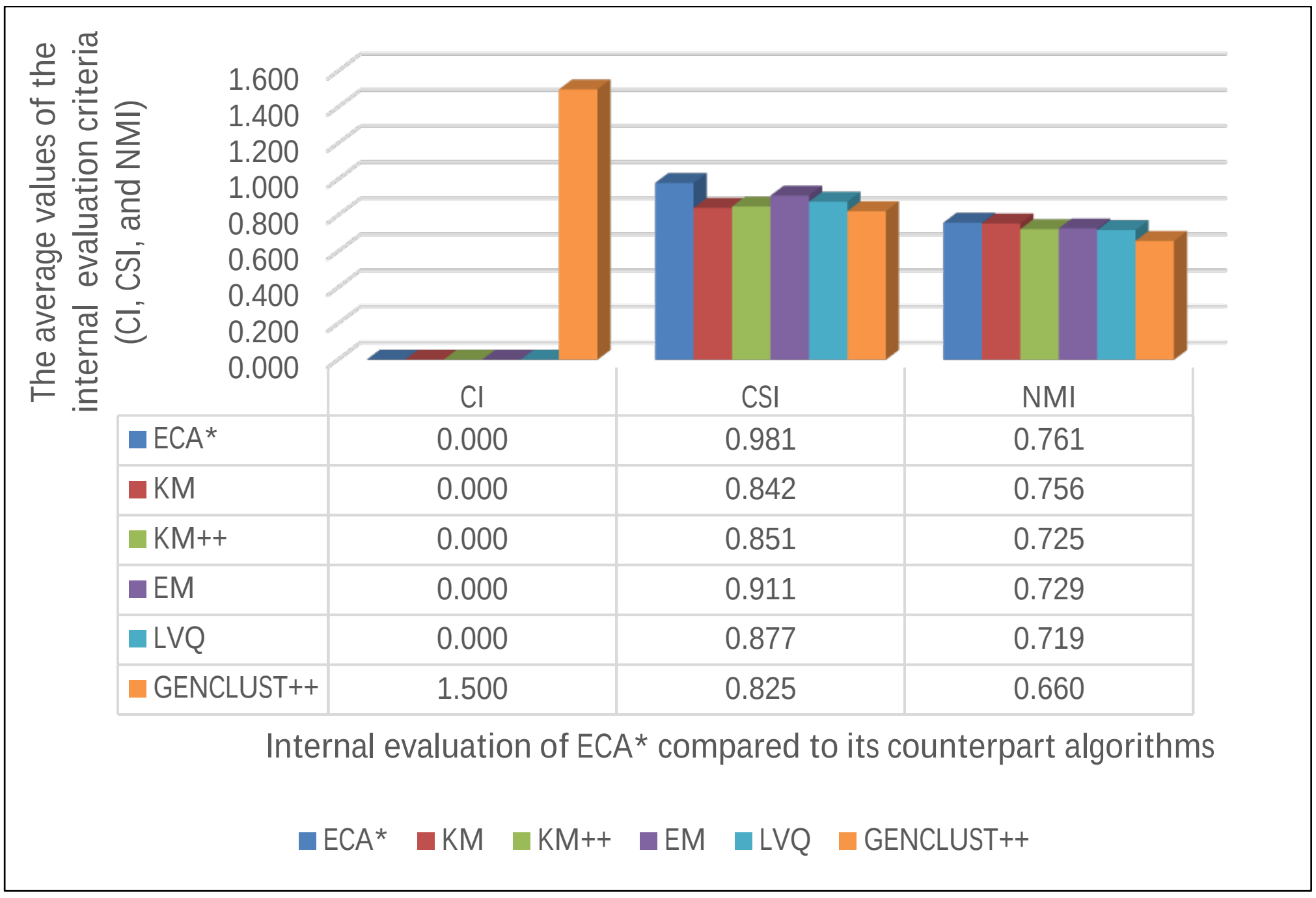

| | CI | CSI | NMI |
|---|---|---|---|
| ECA* | 0.000 | 0.981 | 0.761 |
| KM | 0.000 | 0.842 | 0.756 |
| KM++ | 0.000 | 0.851 | 0.725 |
| EM | 0.000 | 0.911 | 0.729 |
| LVQ | 0.000 | 0.877 | 0.719 |
| GENCLUST++ | 1.500 | 0.825 | 0.660 |

**Figure (5.5): The average values of the external evaluation criteria (CI, CSI, and NMI) for six algorithms in 32 numerical datasets**

Furthermore, Figure (5.6) presents the average values of the internal evaluation criteria (SSE, nMSE, and ε- ratio) for 32 numerical datasets. SSE is used to measure the squared differences between each data observation with its cluster centroid. In this comparison, ECA* performs well in most of the datasets, and it is the winner in 25 datasets. On the other hand, the counterpart algorithms of ECA* are the winner in a few numbers of datasets. Therefore, ECA* performs well compared to its competitive algorithms in the current datasets. In addition to SSE, the comparison of external cluster evaluation measured by nMSE for ECA* and its counterpart algorithms, ECA* performs well compared to its counterpart techniques in most of the datasets. After that, GENCLUS++ is the second winner that has good results in four datasets (S2, Aggregation, Flame, and G2-16-80). On the other hand, KM is not a winner in all cases, whereas KM++ is the winner in two datasets (A3, and G2-1024-100). It is necessary to mention that KM++, EM, and LVQ are the winners for the G2-1024-100 dataset. Also, EM is the winner for S1, and LVQ is the winner for two other datasets (Dim-512, and G2-16-100).

Finally, the comparison of internal cluster evaluation measured by ε- ratio for ECA* and its counterpart algorithms, ECA* performs well compared to its counterpart techniques in most of the datasets. After that, EM is the second winner that has good results in three datasets (S3, R15, and G2-16-100). Meanwhile, KM++, LVQ, and GENCLUS++ are the winner in only one case, whereas KM does not win in any of the 32 datasets. Overall, the results indicate that ECA* overcomes its competitive algorithms in all the mentioned external evaluation measures. ECA* records the minimum value of SSE followed by EM, KM++, KM, GENCLUS++, and LVQ. For nMSE, ECA* outperforms other techniques, while EM marginally performs the same as ECA*. Succeeded by these two algorithms, these techniques GENCLUS++, KM++, and KM perform relatively similar in nMSE. Lastly, LVQ has the least performed algorithm on the list. Once more, ECA* records the minimum value for the ε- ratio, followed by EM. That is, ECA* and EM have the best cluster quality among their counterpart techniques once their cluster results are compared with the theoretical results achieved for approximation techniques. KM++, GENCLUS++, and KM are in the second position for cluster quality measured by ε- ratio. Incidentally, LVQ records the worst cluster quality measured by the ε- ratio among its counterpart algorithms.

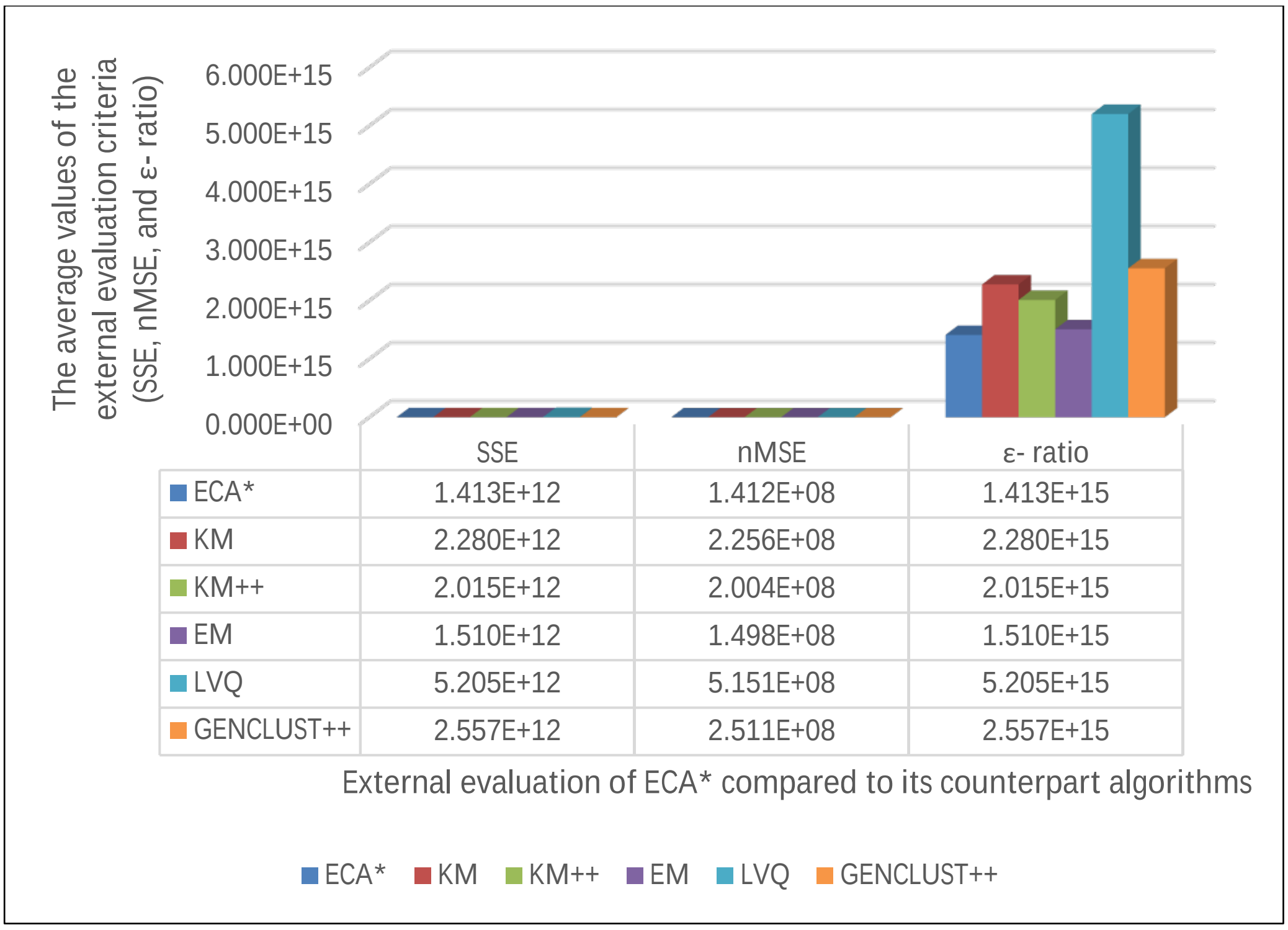

| | SSE | nMSE | ε- ratio |
|---|---|---|---|
| ECA* | 1.413E+12 | 1.412E+08 | 1.413E+15 |
| KM | 2.280E+12 | 2.256E+08 | 2.280E+15 |
| KM++ | 2.015E+12 | 2.004E+08 | 2.015E+15 |
| EM | 1.510E+12 | 1.498E+08 | 1.510E+15 |
| LVQ | 5.205E+12 | 5.151E+08 | 5.205E+15 |
| GENCLUST++ | 2.557E+12 | 2.511E+08 | 2.557E+15 |

**Figure (5.6): The average values of the internal evaluation criteria (SSE, nMSE, and ε- ratio) for six algorithms in 32 numerical datasets**

### 5.3.3 Performance Ranking Framework of ECA*

In the previous section, the performance of ECA* was studied comparing with its counterpart algorithms using 32 benchmark datasets. This section empirically measures how much the performance of these techniques depending on five factors: (i) Cluster overlap. (ii) The number of clusters. (ii) Cluster dimensionality. (iv) Well-structured clusters (structure). (v) Cluster shape. The measure of ECA*'s overall performance with its competitive techniques in terms of the above factors is presented in the form of a framework. These factors have been also studied by [51] for evaluating different clustering algorithms. Figure (5.7) presents the overall average performance rank of ECA* with its counterpart algorithms. In this figure, these techniques are ranked based on the above factors using internal measures and external measures. For both internal and external measures, rank 1 represents the best performing algorithm to a current factor, whereas rank 5 refers to the worst-performing algorithms to a current factor.

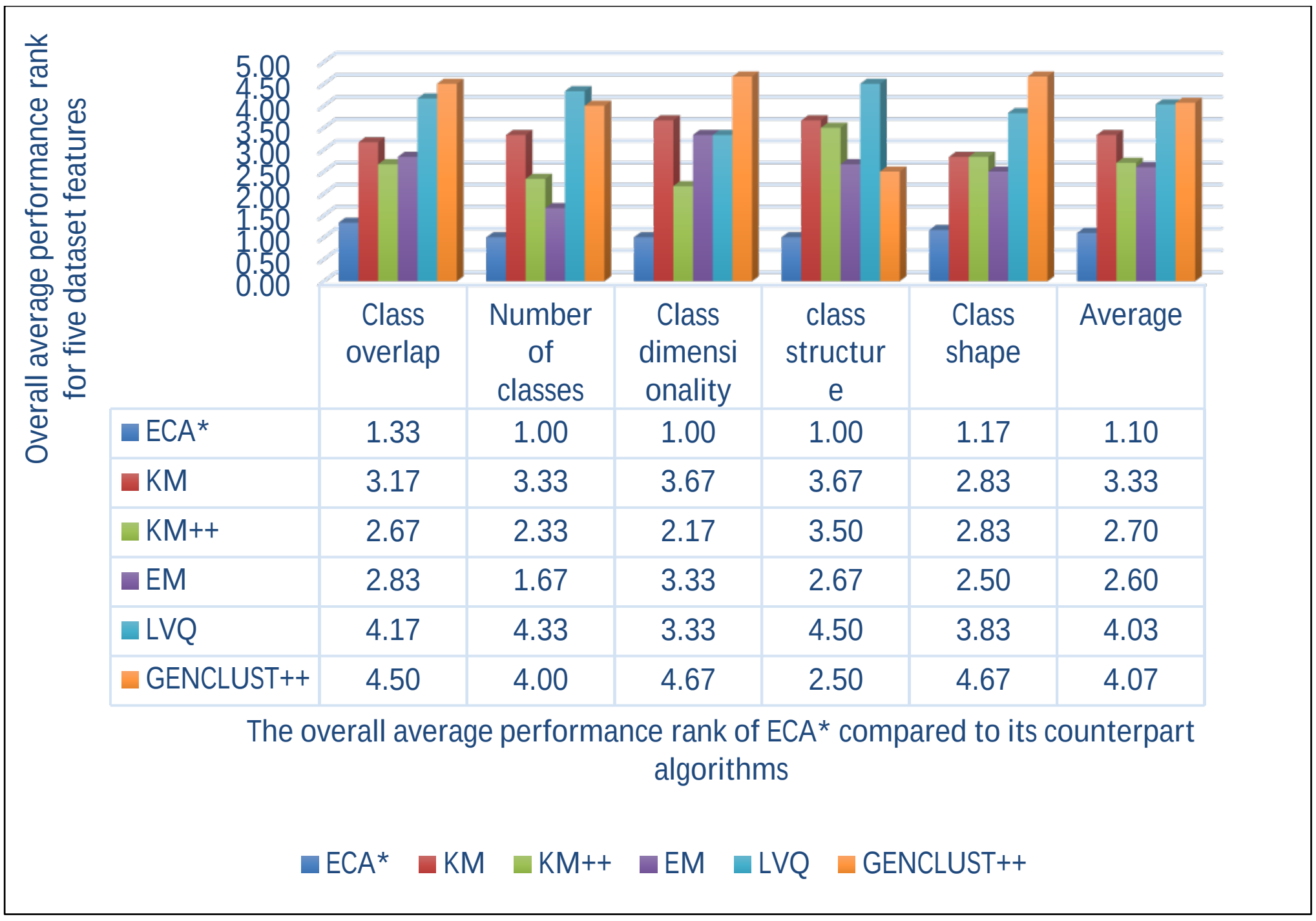

| | Class overlap | Number of classes | Class dimensionality | class structure | Class shape | Average |
|---|---|---|---|---|---|---|
| ECA* | 1.33 | 1.00 | 1.00 | 1.00 | 1.17 | 1.10 |
| KM | 3.17 | 3.33 | 3.67 | 3.67 | 2.83 | 3.33 |
| KM++ | 2.67 | 2.33 | 2.17 | 3.50 | 2.83 | 2.70 |
| EM | 2.83 | 1.67 | 3.33 | 2.67 | 2.50 | 2.60 |
| LVQ | 4.17 | 4.33 | 3.33 | 4.50 | 3.83 | 4.03 |
| GENCLUST++ | 4.50 | 4.00 | 4.67 | 2.50 | 4.67 | 4.07 |

**Figure (5.7): Overall average performance rank of ECA* compared to its counterpart algorithms for 32 datasets according to five dataset features**

The results show that ECA* outperformed its competitive techniques in all the mentioned factors. After ECA*, three different techniques scored the second successful technique according to the five mentioned factors: (1) EM is considered as a useful technique to solve the dataset problems that have cluster overlap and cluster shape. (2) KM++ works well for the datasets that have several clusters, and cluster dimensionality. (3) GENCLUST++ does well for the datasets that have a good cluster structure. Figure (5.8) depicts a performance ranking framework of ECA* compared with its counterpart algorithms for 32 multi-featured datasets according to these features (cluster overlap, cluster number, cluster dimensionality, cluster structure, and cluster shape). In this Figure, the dark colours show that the algorithm is well performed (ranked as 1) for a specified dataset features, whereas the light colours indicate that the technique is poorly performed (ranked as 6) for a specified dataset property. The colours in between dark and light are ranked as 2, 3, 4, and 5, respectively. Overall, the rank of the algorithm for all the five dataset features is ECA*, EM, KM++, KM, LVQ, and GENCLUST++.

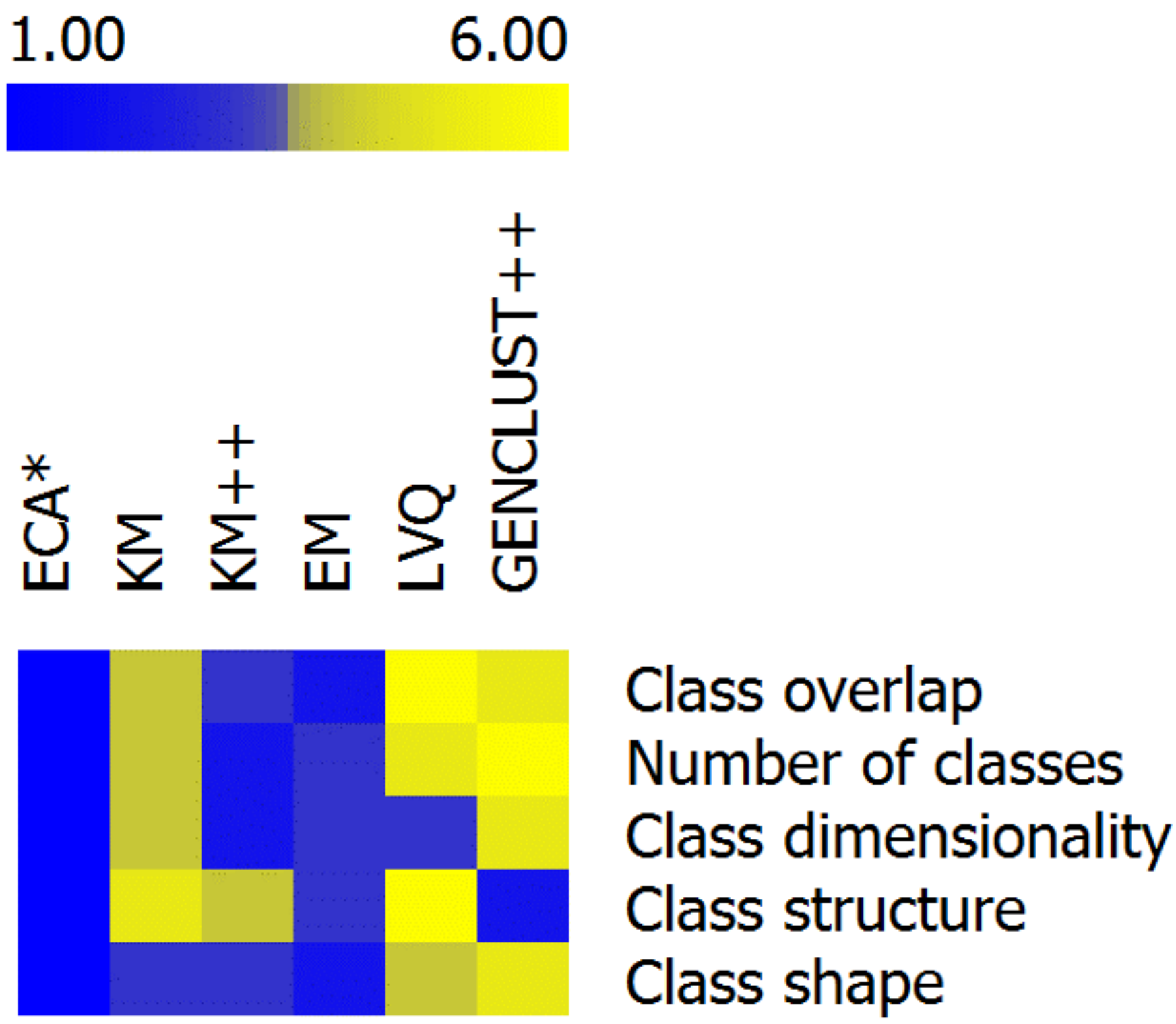

**Figure (5.8): A performance ranking framework of ECA* compared to its counterpart algorithms for 32 multi-featured datasets**

## 5.4 Result Analysis of Formal Context Reduction

This section consists of two sub-sections: the evaluation method, and result analysis.

### 5.4.1 Evaluation Method

To evaluate this study's proposed framework comparing with the standard framework for deriving concept hierarchies, there is a need to assess how well the concept lattices and concept hierarchies represent the given domain. Several methods are available to analyse lattice graphs. One possible method is to measure the similarity between the newly generated concept hierarchies to a given hierarchy for a specified domain. Nevertheless, the difficultly is how to define the similarity and similarity measures between the hierarchy of concepts. Although a few studies are dealing with calculating the similarities between concept lattices, simple graphs, and concept graphs proposed by [142] and [143], it is not clear how to measure these similarities and also translate them into concept hierarchies. Another exciting research was introduced by [144] in which ontologies can be evaluated based on different levels. However, these levels are the measure used to lexical overlap and taxonomic between two ontologies. Accordingly, ontologies are evaluated with the defined similarity measures, and thus the agreement on the task of modelling ontology is yielded from different subjects. Because concept lattice is a particular type of homomorphism of the structure, evaluating the concept lattices by finding a relationship between the lattice graphs is another evaluation method. In graph theory, if there is an isomorphism from the subdivision of the first graph to some subdivision of the second one, these two graphs are considered as homeomorphisms [145]. For example, if there is an isomorphism from the subdivision of graph *X* to some subdivision of graph *Y*, graph *X* and *Y* are homeomorphisms. Furthermore, a subdivision of a graph in the theory of graphs can be defined as a graph resulting from the subdivision of edges in itself. Moreover, the subdivision of some edge a with endpoints {x, y} creates a graph containing one new vertex *z* and replacing a with two new edges, {*y*, *z*} and {*z*, *y*}. This approach of evaluation seems to be more systematic and formal because finding two graphs is theoretically precise whether homeomorphism is or not. To determine whether the original concept lattice with the results are isomorphic or not, common characteristics need to be found out between them. Such characteristics are called concept lattice-invariant, which is maintained by isomorphism. The concept lattice-invariant, applied in several studies in the literature [108], including the number of concepts, the number of edges, degrees of the concepts, and length of cycle. These invariants are also adapted for use in this thesis.

### 5.4.2 Result Analysis

To analyse the results of the experiment, the two concept lattices were evaluated using the 385 text corpora based on concept lattice-invariant as it is represented by the number of concepts, the number of edges, lattice height, and estimation of lattice width. Figure (5.9) presents the concept counts for concept lattice 1, and 2 for the used corpora in this study. It indicated that the number of concepts of lattice 2 is less than the ones in lattice 2. The mean impact of applying adaptive ECA* on the second concept lattice compared with its original one is the decrease of concept counts by 16%.

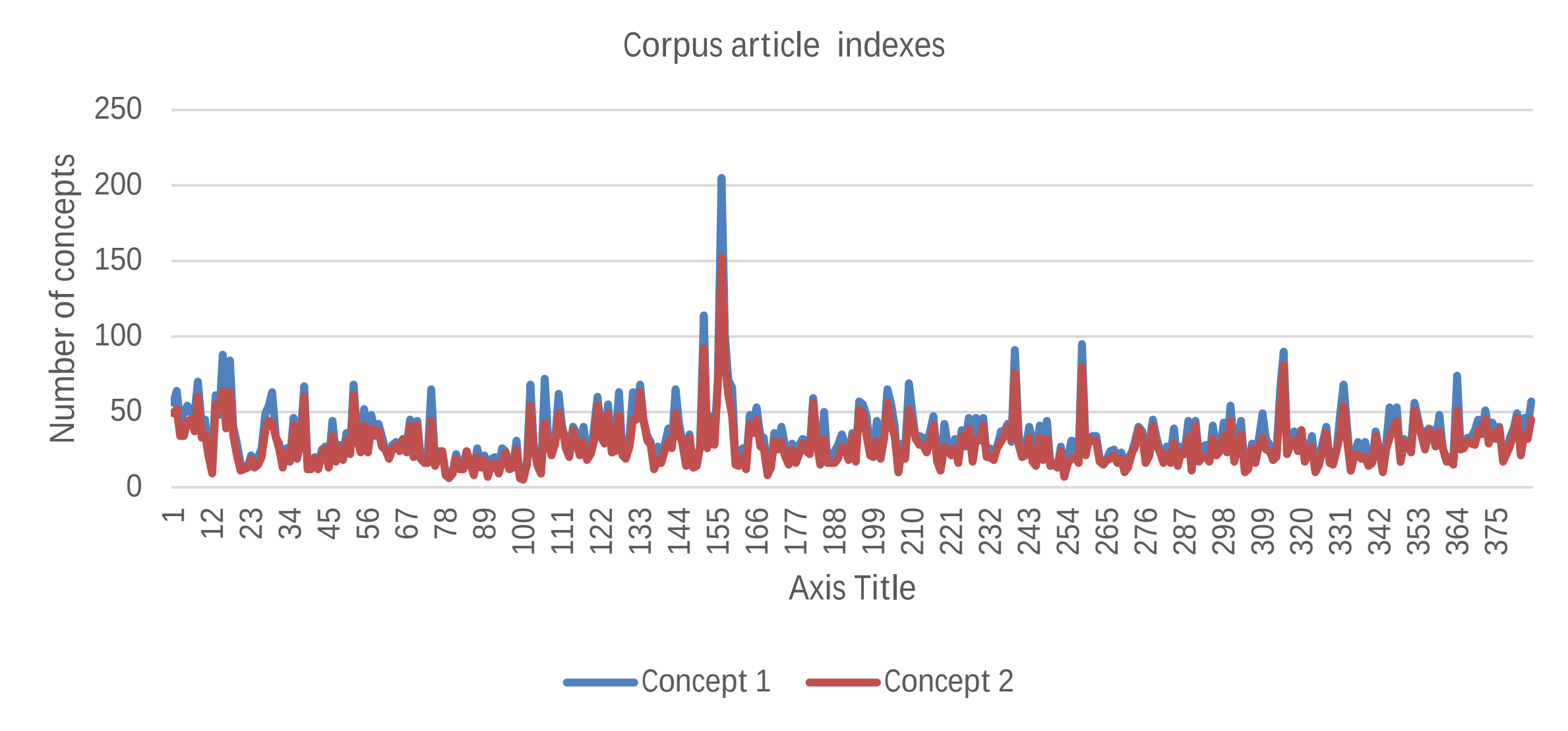

**Figure (5.9): The concept counts for concept lattice 1, and 2 for the 385 corpus articles**

Moreover, Figure (5.10) shows the fluctuation of the number of edges for concept lattice 1 and 2 for the used corpora. The number of edges of lattice 2 is less than or equal to the ones in lattice 2 from corpus articles 1 to 385. On average, the concept lattice 2 is simplified by 16% in terms of their edges.

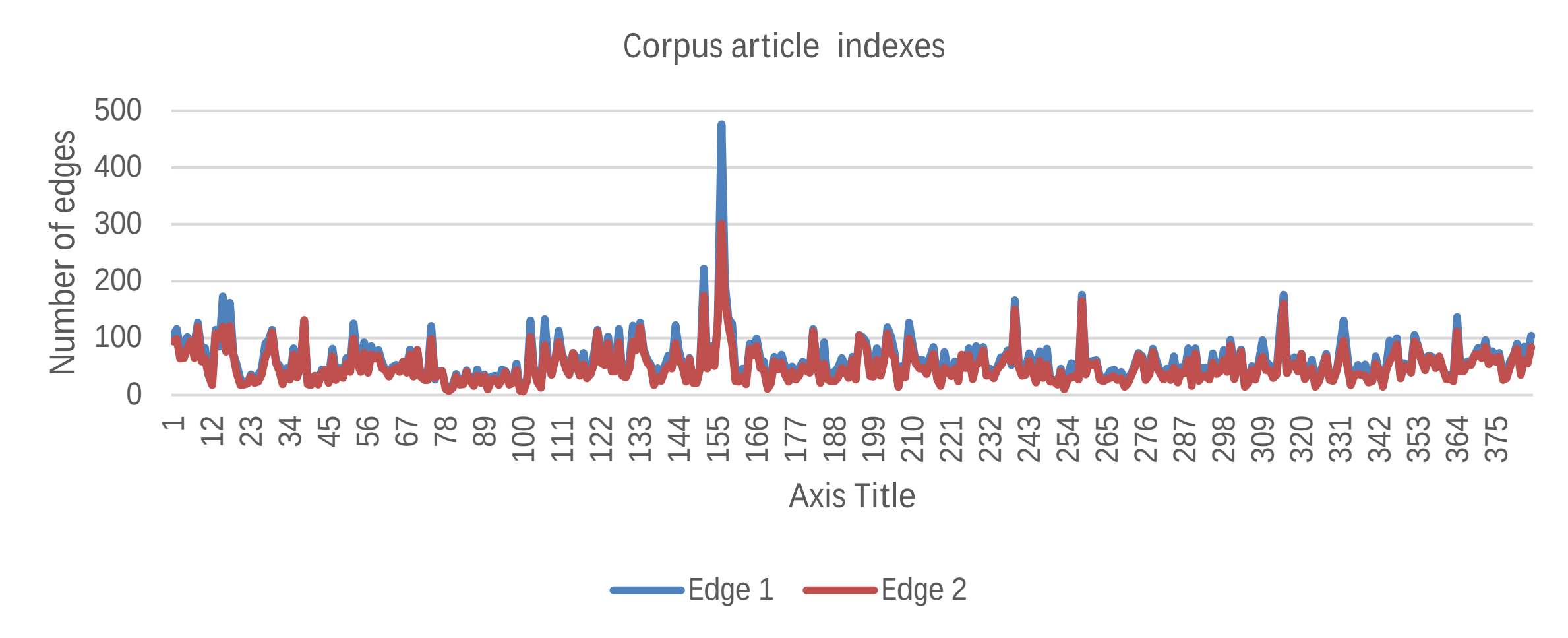

**Figure (5.10): The fluctuation of edge counts for concept lattice 1, and 2 for the 385 corpus articles**

For the heights of concept lattice 1, and 2, they have relatively the same lattice heights if the average heights of both lattices are taken. Nevertheless, the heights of the resulting lattice (lattice 2) were changing from one text corpus into another one. In most cases, the heights of resulting lattice were equal to the original lattice. The heights of concept lattice 1, and 2 are depicted in Figure (5.11).

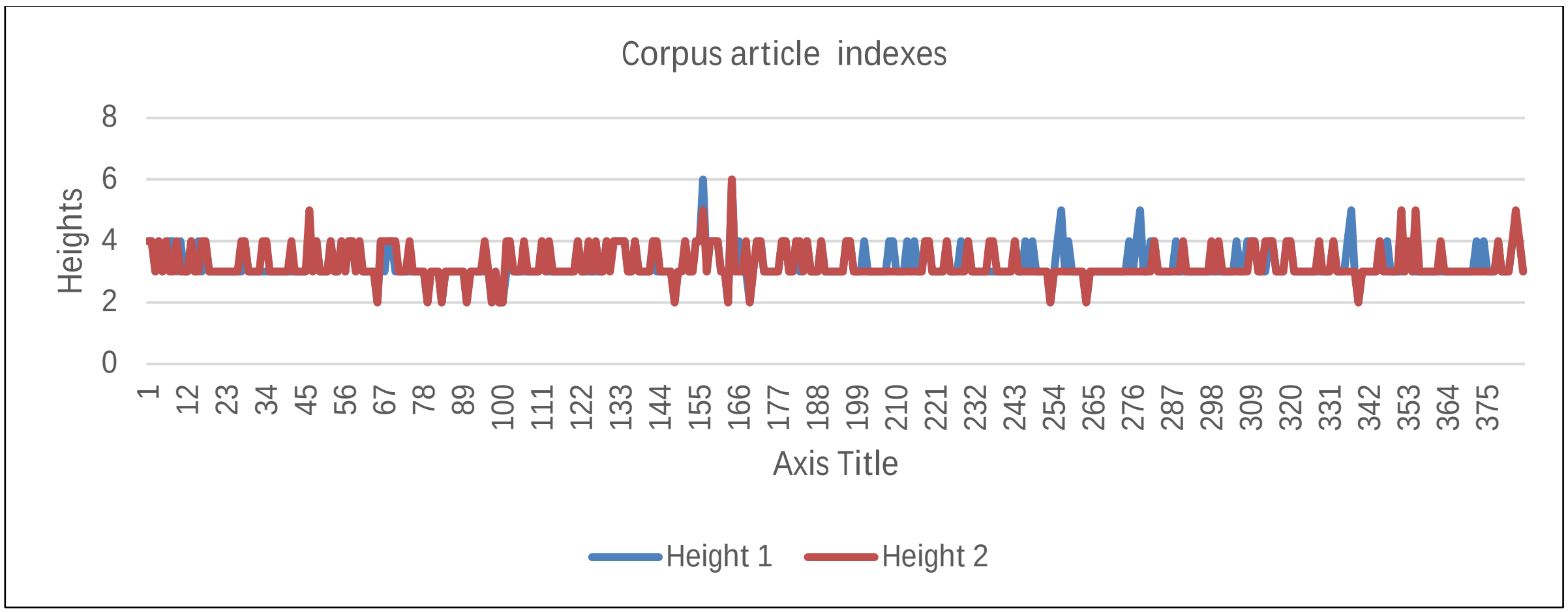

**Figure (5.11): The heights of lattice 1, and 2 for the 385 corpus articles**

Moreover, for the width of the original and resulting lattice, their widths were considered equal in several cases. In contrast, in some instances, there is a sharp fall in the resulting lattice's width. On the whole, the resulting lattice's width was reduced by one-third in comparison to the underlying lattice.

To further illustrate the results, the results of both concept lattices applied to the 385 datasets are aggregated in Table (5.8). According to the aggregated values given in the table, the concept count, edge count, height, and width substantially decreased in the resulting lattice compared with the concept lattice 1. For example, the average number of concepts and edges in the resulting lattice decreased by 16% with similarly the same height of lattices. Meanwhile, the width of reduced concept lattices was considerably reduced compared with the original one. However, three factors might implicitly or explicitly affect the results of reduced concept lattices: the depth of hypernym and hyponym of the WordNet, the cohort, and length of text corpora.

**Table (5.8): Aggregated results of concepts and edges of concept lattice 1, and 2**

| Aggregated functions | Concept 1 | Concept 2 | Edge 1 | Edge 2 | Height 1 | Height 2 | Width 1 | Width 2 |
|---|---|---|---|---|---|---|---|---|
| Mean | 33.306 | 27.930 | 60.026 | 50.226 | 3.221 | 3.221 | [22.51,30.01] | [12.64, 24.61] |
| Median | 30.000 | 25.000 | 52.000 | 43 | 3.000 | 3.000 | [21,26] | [12, 21] |
| Sum | 12823.000 | 10753.000 | 23110 | 19337 | 1240 | 1240.000 | [8666,11554] | [4865, 9475] |
| Max | 205.000 | 152.000 | 476.000 | 301 | 6.000 | 6.000 | [86,200] | [51, 217] |
| Min | 7.000 | 5.000 | 10.000 | 6 | 2.000 | 2.000 | [4,5] | [3, 3] |
| STDV P | 18.697 | 14.994 | 38.559 | 30.882 | 0.535 | 0.530 | [11.26,18.47] | [5.64, 17.65] |
| STDV S | 18.722 | 15.014 | 38.609 | 30.922 | 0.536 | 0.531 | [11.27,18.5] | [5.65, 17.68] |
| STDEVA | 18.722 | 15.014 | 38.609 | 30.922 | 0.536 | 0.531 | [11.27,18.5] | [5.65, 17.68] |

Additionally, it can be seen that the resulting concept lattice is still isomorphic from the meanings of the basic concept lattice, and the most simplified concepts were used in their formal context. However, there are some invariants that do not exist in the concept lattice 2. In the reduced concept lattice, the objects and attributes of these non-existent concepts were the least used or may be uninteresting. Consequently, thanks to the common invariants between them, it can be seen that the reduced concept lattices have relatively the same quality of results as the concept lattice 1. As a result, the original lattice is homeomorphism to the resulting lattice by 89%. That means, there is about 11% quality and meaning loss between the two lattices. This loss leads to the quality loss of the resulting concept hierarchy in contract with the original one.

# Chapter Six: Conclusions and Future Works

## 6.1 Conclusions

Firstly, the statistical experiments conducted on BSA in the literature were analysed, and the performance of BSA was evaluated against its counterpart algorithms for different types of real-world problems. Further, a general and an operational framework of the main expansions and implementations of BSA were proposed. Further, experiments were set up to evaluate BSA with other competitive algorithms statistically. It was found from these frameworks that BSA can be improved by amending its parameters, combining it with some other techniques or meta-heuristic algorithms to improve its performance, and adapting it to minimise different optimisation problems. BSA was not able to minimise all the 16 benchmark functions used in the experiment in this thesis. This failure can probably be attributed to four factors: different hardness scores of the problem and the variety of dimensions, search spaces, and cohorts of the problems. A clear conclusion cannot be drawn to determine the success and failure ratios of BSA in solving different types of problems with different hardness scores, problem dimensions, and search spaces.

Secondly, this thesis has proposed a new evolutionary clustering algorithm. The evidence of the experimental studies has indicated two primary results. (i) None of its counterpart algorithms is superior to ECA* concerning their capacity to recognise right clusters of functionally related observations in the given 32 benchmarking datasets. (ii) The proposed performance framework indicates that ECA* is a more competitive algorithm compared with all other techniques, and it is well suited for complex and multi-featured benchmarking problems. Further, these experimental results have inferred to discuss the strengths and weaknesses of ECA*. On the one hand, ECA* has several strengths in comparison with its counterpart algorithms:

1. ECA* is useful for understanding clustering results for diverse and multi-featured benchmarking problems.
2. The proposed ECA* method was observed to increases the performance due to its mut-over strategy.
3. The clustering method is an unsupervised learning algorithm that seems useful in identifying excellent results in the absence of pre-defined and prior knowledge.

4. Another remarkable achievement of ECA* is the use of a dynamic clustering strategy to find the right number of clusters.
5. The use of statistical techniques in ECA* helps to adjust the clustering centroids towards its objective function and to result in excellent clustering centroids accordingly.

Thirdly, two forms of framing the concept hierarchy derivation were proposed: (i) The state-of-art framework was reviewed for forming concept hierarchies from free text using FCA; (ii) A novel framework was proposed using adaptive ECA* to remove the erroneous and uninteresting pairs in the formal context and to reduce the size of the formal context, resulting in less time consuming when concept lattice derives from this. Later, an experiment was conducted to examine the results of the reduced concept lattice against the original lattice. Through the lens of the experiment and result analysis, the resulting lattice is homeomorphism to the original one with preserving the structural relationship between the two concept lattices. This similarity between the two lattices preserves the quality of resulting concept hierarchies by 89% in contrast to the basic one. The quality of resulting concept hierarchies is significantly promising, and there is information loss by 11% between the two concept hierarchies. Indeed, the introduced framework is not out of shortcomings. First, assigning an adaptive value for hypernym and hyponym depths sometimes can improve the quality of the produced concept lattices. Second, polysemy problems also remain with this framework because it does not have the potential to include signals or indications with many related meanings. As the framework cannot sign the specific meaning of several related meanings, a word is usually regarded as different from homonymy, where the multiple meanings of a word can be unrelated or unrelated. Last but not least, classification as a way of identifying text bodies into categories or subcategories has been ignored as this can contribute to a better quality of results and facilitate a decision for the assignment of a proper value of hypernyms and hyponyms depths. We have noticed that in recent years, evolutionary clustering algorithm is still considered one of the most promising algorithms in terms of being integrated with other technologies. This means that the changes and developments in it are continuing to come up with new methods that can be more effective and capable of yielding more satisfying results. For future reading, the authors advise the reader could optionally read the following research works [2], [22], [149]–[154], [23], [61], [120], [121], [128], [146]–

[148].

## 6.2 Future Works

1. The first part of this thesis has a four-fold contribution in the direction of future research and development.

A. The BSA frameworks proposed in this thesis can guide researchers who are working on the improvement of BSA.

B. It was seen that BSA was mostly used in the engineering fields [10], particularly in the areas of electrical and mechanical engineering and information and communication engineering.

C. The experiments conducted in this thesis have a role of revealing the sensitivity of BSA towards the hardness score of the optimisation problem, problem dimension and type, and search space of the problem to a certain extent.

D. It is recommended that conducting more studies in the area of tackling the limitations of BSA is necessary.

2. Contrarily, the experimental results conducted in this thesis recognise some weaknesses of ECA* that could be addressed in the future:

A. Initialising the optimal value for the pre-defined variables in ECA*, such as the number of social class rank, and cluster density threshold is a difficult task.

B. Choosing a different value of social class ranks might affect how to define the cluster threshold density.

C. The merging process of clusters is not always accurate.

D. Since ECA* is considered as an integrated multidisciplinary algorithm, this may sometimes bring a reluctant conclusion that multidisciplinary may not facilitate the use of ECA* by single discipline scholars and professionals.

E. Exploiting and adapting ECA* for real-world applications is a vital possibility for future research, such as ontology learning in the Semantic Web [61], engineering applications [2], e-government services [146], [147], intelligent systems [155], and web science [149].

3. Further researches could be carried out on reducing the size of formal context in deriving concept hierarchies from corpora in the future to yield a better quality of the resulting lattice with less loss of meaning.

# List of Publications

The following list of publications/manuscripts are extracted from this thesis:

- **B. A. Hassan** and T. A. Rashid, "Datasets on statistical analysis and performance evaluation of backtracking search optimisation algorithm compared with its counterpart algorithms," Data Br., vol. 28, p. 105046, 2020
- **B. A. Hassan** and T. A. Rashid, "Datasets on evaluating ECA* compared to its counterpart algorithms," Data Br., 2020
- **B. A. Hassan** and T. A. Rashid, "Dataset Results on Formal Context Reduction in Deriving Concept Hierarchies from Corpora Using Adaptive Evolutionary Clustering Algorithm" Data Br., 2021
- **B. A. Hassan** and T. A. Rashid, "Data for: statistical analysis and performance evaluation of backtracking search optimisation algorithm compared with its competitive algorithms," Mendeley Data, v3, 2020, doi: 10.17632/hx8xbyjmf5.3
- **B. A. Hassan** and T. A. Rashid, "Performance Evaluation Results of evolutionary clustering algorithm star for clustering heterogeneous datasets", Mendeley Data, v1, 2019, doi: 10.17632/bsn4vh3zv7.1
- **B. A. Hassan** and T. A. Rashid, "Data for: Formal Context Reduction in Deriving Concept Hierarchies from Corpora Using Adaptive Evolutionary Clustering Algorithm", Mendeley Data, v1, 2021

# APPENDIX A

# Benchmark Functions

The dataset contains the simulation data of BSA compared to four competitive algorithms (DE, PSO, ABS, and FF) applied to minimise sixteen benchmark functions in three tests. The optimisation benchmark problems used in these tests are presented in Table (A.1).

**Table (A.1): Benchmark functions**

| ID | Function name | Matehmatical function |
|---|---|---|
| F1 | Ackley | $f(x,y) = -200e^{-0.2\sqrt{x^2+y^2}} + 5e^{cos(3x)+sin(3y)}$ |
| F2 | Alpine01 | $f(\mathrm{x}) = f(x_1,...,x_n) = \sum_{i=1}^{n} \lvert x_i sin(x_i) + 0.1x_i \rvert$ |
| F3 | Bird | $f(x,y) = sin(x)e^{(1-cos(y))^2} + cos(y)e^{(1-sin(x))^2} + (x-y)^2$ |
| F4 | Leon | $f(x,y) = 100(y-x^3)^2 + (1-x)^2$ |
| F5 | CrossInTray | $f(x,y) = -0.0001(\lvert sin(x)sin(y)exp(\lvert 100 - \frac{\sqrt{x^2+y^2}}{\pi} \rvert)\rvert + 1)^{0.1}$ |
| F6 | Easom | $f(x,y) = -cos(x_1)cos(x_2)exp(-(x-\pi)^2 - (y-\pi)^2)$ |
| F7 | Whitley | $f(x) = \sum_{i=1}^{n}\sum_{j=1}^{n}(\frac{(100(x_i^2 - x_j)^2 + (1 - x_j)^2)^2}{4000} - cos\ (100(x_i^2 - x_j)^2 + (1 - x_j)^2) + 1)$ |
| F8 | EggCrate | $f(x,y) = x^2 + y^2 + 25(sin^2(x) + sin^2(y))$ |
| F9 | Griewank | $f(\mathrm{x}) = f(x_1,...,x_n) = 1 + \sum_{i=1}^{n} \frac{x_i^2}{4000} - \prod_{i=1}^{n} cos(\frac{x_i}{\sqrt{i}})$ |
| F10 | HolderTable | $f(x,y) = -\lvert sin(x)cos(y)exp(\lvert 1 - \frac{\sqrt{x^2+y^2}}{\pi} \rvert)\rvert$ |
| F11 | Rastrigin | $f(x,y) = 10n + \sum_{i=1}^{n} (x_i^2 - 10cos(2\pi x_i))$ |
| F12 | Rosenbrock | $f(x,y) = \sum_{i=1}^{n} [b(x_{i+1} - x_i^2)^2 + (a - x_i)^2]$ |
| F13 | Salomon | $f(\mathbf{x}) = f(x_1,...,x_n) = 1 - cos(2\pi\sqrt{\sum_{i=1}^{D} x_i^2}) + 0.1\sqrt{\sum_{i=1}^{D} x_i^2}$ |
| F14 | Sphere | $f(\mathrm{x}) = f(x_1,x_2,...,x_n) = \sum_{i=1}^{n} x_i^2.$ |
| F15 | StyblinskiTang | $f(\mathrm{x}) = f(x_1,...,x_n) = \frac{1}{2}\sum_{i=1}^{n} (x_i^4 - 16x_i^2 + 5x_i)$ |
| F16 | Schwefel26 | $f(\mathrm{x}) = f(x_1,x_2,...,x_n) = 418.9829d - \sum_{i=1}^{n} x_i sin(\sqrt{\lvert x_i \rvert}).$ |

# APPENDIX B

## Datasets of Backtracking Search Optimisation Algorithm

The datafiles (XLSX format) are the raw data results from Tests 1, 2, and 3 used to evaluate the performance of BSA compare to its counterpart algorithms (DE, PSO, ABS, and FF) applied to minimise sixteen benchmark problem in these three tests. Additionally, Information about the Tests 1, 2, and 3 are depicted in the tables as follows: (i) Tables (B.1), (B.2), and (B.3) present basic statistics of 30-solutions obtained by the algorithms in Test 1. (ii) Tables (B.4), (B.5), and (B.6) present basic statistics of 30-solutions obtained by the algorithms in Test 2. (iii) Tables (B.7) and (B.8) presents the ratio of successful minimisation of the functions for Tests 1 and 2.

1. Tables (B.1), (B.2), and (B.3) present basic statistics (Mean: mean-solution, S.D.: standard-deviation of mean-solution, Best: the best solution, Worst: the worst solution, Exec. Time: mean runtime in seconds, No. of succeeds: number of successful minimisation, and No. of Failure: number of failed minimisation) of 30-solutions obtained by BSA, DE, PSO, ABC, and FF in Test 1 to minimise F1-F16 functions with default search space with number of variables 10, 30, and 60 respectively.

**Table (B.1): Basic statistics of the 30-solutions obtained by BSA, DE, PSO, ABC, and FF in Test 1 (default search space with Nvar1 dimensions)**

| Functions | Statistics | BSA | DE | PSO | ABC | FF |
|---|---|---|---|---|---|---|
| F1 | Mean | 688.3666667 | 209.8 | NC | NC | 323.3333333 |
| | S.D. | 61.19808614 | 5.761944116 | NC | NC | 7.716231586 |
| | Best | 584 | 193 | NC | NC | 301 |
| | Worst | 830 | 220 | NC | NC | 334 |
| | Exec. time | 0.3129113 | 0.767031967 | NC | NC | 8.876311833 |
| | No. of Succeeds | 30 | 30 | 0 | 0 | 30 |
| | No. of Failures | 0 | 0 | 30 | 30 | 0 |
| F2 | Mean | 78.96666667 | 305.1666667 | NC | NC | 93.43333333 |
| | S.D. | 11.21723838 | 22.1672068 | NC | NC | 78.09882481 |
| | Best | 63 | 245 | NC | NC | 12 |
| | Worst | 109 | 348 | NC | NC | 248 |
| | Exec. time | 0.079013967 | 0.613814533 | NC | NC | 1.1518529 |
| | No. of Succeeds | 30 | 30 | 0 | 0 | 30 |
| | No. of Failures | 0 | 0 | 30 | 30 | 0 |
| F3 | Mean | 288.2 | 128.7666667 | 54.59259259 | 35.33333333 | 32.5 |
| | S.D. | 57.68487521 | 40.0548618 | 4.901014489 | 7.169539654 | 14.28708122 |
| | Best | 193 | 72 | 44 | 23 | 6 |
| | Worst | 387 | 201 | 62 | 53 | 63 |
| | Exec. time | 0.121238167 | 0.392393267 | 0.097268 | 0.087277867 | 0.656926867 |
| | No. of Succeeds | 30 | 30 | 27 | 30 | 30 |
| | No. of Failures | 0 | 0 | 3 | 0 | 0 |
| F4 | Mean | 387.7 | 3960.133333 | 857 | 128.3333333 | 1480.666667 |
| | S.D. | 127.0707514 | 244.6563717 | 675.7609249 | 136.9530756 | 34.20963096 |
| | Best | 47 | 3546 | 5 | 26 | 1388 |
| | Worst | 588 | 4472 | 1747 | 641 | 1520 |
| | Exec. time | 0.142008467 | 10.51661843 | 1.519789933 | 0.217805708 | 22.88727067 |
| | No. of Succeeds | 30 | 30 | 30 | 24 | 30 |
| | No. of Failures | 0 | 0 | 0 | 6 | 0 |
| F5 | Mean | 121.1666667 | 38.26666667 | 27.23333333 | 14.76666667 | 3.766666667 |
| | S.D. | 34.72362143 | 8.337424283 | 9.814427569 | 4.031627834 | 0.678910554 |
| | Best | 45 | 23 | 6 | 7 | 2 |
| | Worst | 175 | 52 | 40 | 24 | 5 |
| | Exec. time | 0.027849683 | 0.095730833 | 0.033622767 | 0.042479367 | 0.057955067 |
| | No. of Succeeds | 30 | 30 | 30 | 30 | 30 |
| | No. of Failures | 0 | 0 | 0 | 0 | 0 |
| F6 | Mean | 760.4333333 | 208.5 | 63.83333333 | NC | 128.1666667 |
| | S.D. | 63.97665702 | 33.34847357 | 3.939922398 | NC | 20.77977022 |
| | Best | 617 | 162 | 54 | NC | 91 |
| | Worst | 880 | 326 | 71 | NC | 166 |
| | Exec. time | 0.251127833 | 0.5434189 | 0.094766533 | NC | 1.785906667 |
| | No. of Succeeds | 30 | 30 | 30 | 0 | 30 |
| | No. of Failures | 0 | 0 | 0 | 30 | 0 |
| F7 | Mean | 972 | 4809.6 | NC | NC | NC |
| | S.D. | 349.46797 | 1851.094463 | NC | NC | NC |
| | Best | 555 | 1616 | NC | NC | NC |
| | Worst | 1926 | 8339 | NC | NC | NC |
| | Exec. time | 0.505736867 | 8.1352712 | NC | NC | NC |
| | No. of Succeeds | 30 | 10 | 0 | 0 | 0 |
| | No. of Failures | 0 | 20 | 30 | 30 | 30 |
| F8 | Mean | 119.1 | 2897.133333 | 637.5666667 | 15.875 | 5.071428571 |
| | S.D. | 30.35121994 | 22.93278534 | 4.636313438 | 2.997281377 | 1.741190981 |
| | Best | 63 | 2855 | 627 | 11 | 3 |
| | Worst | 169 | 2944 | 650 | 21 | 10 |
| | Exec. time | 0.044835 | 4.779958067 | 1.223019467 | 0.029835208 | 0.066557357 |
| | No. of Succeeds | 30 | 30 | 30 | 24 | 28 |
| | No. of Failures | 0 | 0 | 0 | 6 | 2 |
| F9 | Mean | 373.1 | 805.2666667 | NC | NC | NC |
| | S.D. | 82.92345868 | 100.8679346 | NC | NC | NC |
| | Best | 285 | 631 | NC | NC | NC |
| | Worst | 611 | 1081 | NC | NC | NC |
| | Exec. time | 0.172796067 | 1.5184459 | NC | NC | NC |
| | No. of Succeeds | 30 | 30 | 0 | 0 | 0 |
| | No. of Failures | 0 | 0 | 30 | 30 | 30 |
| F10 | Mean | 314.1333333 | 72.83333333 | 54.26666667 | NC | 20.83333333 |
| | S.D. | 61.64119387 | 14.48681525 | 7.315233672 | NC | 11.21293383 |
| | Best | 209 | 43 | 40 | NC | 5 |
| | Worst | 418 | 111 | 65 | NC | 44 |
| | Exec. time | 0.1276912 | 0.181376167 | 0.088567367 | NC | 0.2915058 |
| | No. of Succeeds | 30 | 30 | 30 | 0 | 30 |
| | No. of Failures | 0 | 0 | 0 | 30 | 0 |
| F11 | Mean | 1038.933333 | 562.3666667 | NC | NC | NC |
| | S.D. | 128.4778079 | 28.48046316 | NC | NC | NC |
| | Best | 811 | 508 | NC | NC | NC |
| | Worst | 1293 | 634 | NC | NC | NC |
| | Exec. time | 0.379444367 | 0.744079967 | NC | NC | NC |
| | No. of Succeeds | 30 | 30 | 0 | 0 | 0 |
| | No. of Failures | 0 | 0 | 30 | 30 | 30 |
| F12 | Mean | 1119.566667 | NC | NC | NC | NC |
| | S.D. | 307.9534989 | NC | NC | NC | NC |
| | Best | 578 | NC | NC | NC | NC |
| | Worst | 2001 | NC | NC | NC | NC |
| | Exec. time | 0.3752452 | NC | NC | NC | NC |
| | No. of Succeeds | 30 | 0 | 0 | 0 | 0 |
| | No. of Failures | 0 | 30 | 30 | 30 | 30 |
| F13 | Mean | NC | NC | NC | NC | NC |
| | S.D. | NC | NC | NC | NC | NC |
| | Best | NC | NC | NC | NC | NC |
| | Worst | NC | NC | NC | NC | NC |
| | Exec. time | NC | NC | NC | NC | NC |
| | No. of Succeeds | 0 | 0 | 0 | 0 | 0 |
| | No. of Failures | 30 | 30 | 30 | 30 | 30 |
| F14 | Mean | 151.9 | 154.6 | 1967.9 | NC | 12.53333333 |
| | S.D. | 31.17840765 | 4.123941871 | 517.0026579 | NC | 1.008013866 |
| | Best | 107 | 148 | 1782 | NC | 11 |
| | Worst | 208 | 162 | 4691 | NC | 15 |
| | Exec. time | 0.092171037 | 0.463599767 | 3.315528667 | NC | 0.156159333 |
| | No. of Succeeds | 30 | 30 | 30 | 0 | 30 |
| | No. of Failures | 0 | 0 | 0 | 30 | 0 |
| F15 | Mean | 490.7666667 | NC | 100 | 659.6666667 | 245.8461538 |
| | S.D. | 56.80073053 | NC | 5.253570215 | 108.7443485 | 4.017589531 |
| | Best | 345 | NC | 94 | 516 | 240 |
| | Worst | 598 | NC | 109 | 949 | 253 |
| | Exec. time | 0.2717145 | NC | 0.188506 | 1.422079033 | 3.067869077 |
| | No. of Succeeds | 30 | 0 | 11 | 30 | 13 |
| | No. of Failures | 0 | 30 | 19 | 0 | 17 |
| F16 | Mean | NC | 254.9615385 | NC | NC | NC |
| | S.D. | NC | 14.32614608 | NC | NC | NC |
| | Best | NC | 231 | NC | NC | NC |
| | Worst | NC | 289 | NC | NC | NC |
| | Exec. time | NC | 0.572146308 | NC | NC | NC |
| | No. of Succeeds | 0 | 26 | 0 | 0 | 0 |
| | No. of Failures | 30 | 4 | 30 | 30 | 30 |

**Table (B.2): Basic statistics of the 30-solutions obtained by BSA, DE, PSO, ABC, and FF in Test 1 (default search space with Nvar2 dimensions)**

| Functions | Statistics | BSA | DE | PSO | ABC | FF |
|---|---|---|---|---|---|---|
| F1 | Mean | 1391.866667 | 600.8 | NC | NC | 373.0333333 |
| | S.D. | 240.1736058 | 11.12747454 | NC | NC | 4.31903033 |
| | Best | 949 | 572 | NC | NC | 362 |
| | Worst | 1899 | 626 | NC | NC | 381 |
| | Exec. time | 0.546482767 | 2.1215337 | NC | NC | 9.566863367 |
| | No. of Succeeds | 30 | 30 | 0 | 0 | 30 |
| | No. of Failures | 0 | 0 | 30 | 30 | 0 |
| F2 | Mean | 467.4333333 | 2653.133333 | NC | NC | 346.0333333 |
| | S.D. | 54.7487574 | 311.8812477 | NC | NC | 34.27121375 |
| | Best | 397 | 2136 | NC | NC | 227 |
| | Worst | 601 | 3367 | NC | NC | 408 |
| | Exec. time | 0.2806565 | 6.5514224 | NC | NC | 6.9452777 |
| | No. of Succeeds | 30 | 30 | 0 | 0 | 30 |
| | No. of Failures | 0 | 0 | 30 | 30 | 0 |
| F3 | Mean | 346.6333333 | 170.8666667 | 60.79310345 | 35.6 | 32.7 |
| | S.D. | 84.07528783 | 38.91346131 | 28.86592073 | 6.003447286 | 16.26748396 |
| | Best | 182 | 99 | 44 | 26 | 5 |
| | Worst | 505 | 236 | 209 | 45 | 67 |
| | Exec. time | 0.190157233 | 0.5601642 | 0.115230897 | 0.0631293 | 0.644730733 |
| | No. of Succeeds | 30 | 30 | 29 | 30 | 30 |
| | No. of Failures | 0 | 0 | 1 | 0 | 0 |
| F4 | Mean | 471.6333333 | 5530.1 | 1152.866667 | 91.71428571 | 1492 |
| | S.D. | 139.2421566 | 266.6691544 | 496.0978583 | 41.26953602 | 38.38821658 |
| | Best | 180 | 4762 | 4 | 23 | 1323 |
| | Worst | 862 | 6071 | 1644 | 160 | 1533 |
| | Exec. time | 0.193297133 | 14.77711263 | 2.1152108 | 0.253110071 | 22.1456451 |
| | No. of Succeeds | 30 | 30 | 30 | 28 | 30 |
| | No. of Failures | 0 | 0 | 0 | 2 | 0 |
| F5 | Mean | 169.1 | 41.76666667 | 29.86666667 | 13.8 | 3.733333333 |
| | S.D. | 49.42800407 | 11.49717607 | 11.64335931 | 4.604345773 | 0.868344971 |
| | Best | 70 | 14 | 2 | 2 | 2 |
| | Worst | 248 | 62 | 54 | 22 | 5 |
| | Exec. time | 0.044610567 | 0.106829767 | 0.046657733 | 0.044076667 | 0.051693733 |
| | No. of Succeeds | 30 | 30 | 30 | 30 | 30 |
| | No. of Failures | 0 | 0 | 0 | 0 | 0 |
| F6 | Mean | 912.9 | 267.5666667 | 64.8 | NC | 121.7333333 |
| | S.D. | 97.77713928 | 41.86487568 | 4.574290976 | NC | 34.65985786 |
| | Best | 678 | 194 | 55 | NC | 19 |
| | Worst | 1045 | 386 | 75 | NC | 173 |
| | Exec. time | 0.318349367 | 0.724312633 | 0.127859867 | NC | 1.7146897 |
| | No. of Succeeds | 30 | 30 | 30 | 0 | 30 |
| | No. of Failures | 0 | 0 | 0 | 30 | 0 |
| F7 | Mean | 4764.566667 | NC | NC | NC | NC |
| | S.D. | 624.9610572 | NC | NC | NC | NC |
| | Best | 3887 | NC | NC | NC | NC |
| | Worst | 6810 | NC | NC | NC | NC |
| | Exec. time | 3.934701977 | NC | NC | NC | NC |
| | No. of Succeeds | 30 | 0 | 0 | 0 | 0 |
| | No. of Failures | 0 | 30 | 30 | 30 | 30 |
| F8 | Mean | 133.1666667 | 3542.366667 | 637.3333333 | 15.55172414 | 4.666666667 |
| | S.D. | 35.99337423 | 39.65409344 | 5.441433212 | 2.599260979 | 1.109400392 |
| | Best | 35 | 3436 | 625 | 9 | 3 |
| | Worst | 203 | 3610 | 651 | 21 | 8 |
| | Exec. time | 0.063654233 | 6.444122133 | 1.30194 | 0.030338172 | 0.077771407 |
| | No. of Succeeds | 30 | 30 | 30 | 29 | 27 |
| | No. of Failures | 0 | 0 | 0 | 1 | 3 |
| F9 | Mean | 587.0333333 | 1297 | NC | NC | NC |
| | S.D. | 80.61123535 | 45.5290737 | NC | NC | NC |
| | Best | 469 | 1234 | NC | NC | NC |
| | Worst | 841 | 1423 | NC | NC | NC |
| | Exec. time | 0.295489567 | 2.409856833 | NC | NC | NC |
| | No. of Succeeds | 30 | 30 | 0 | 0 | 0 |
| | No. of Failures | 0 | 0 | 30 | 30 | 30 |
| F10 | Mean | 366.8666667 | 91.63333333 | 199.1666667 | NC | 19.8 |
| | S.D. | 42.6790212 | 14.87994098 | 554.8413929 | NC | 10.03579799 |
| | Best | 301 | 63 | 36 | NC | 5 |
| | Worst | 457 | 125 | 2450 | NC | 45 |
| | Exec. time | 0.187796133 | 0.2636237 | 0.357409 | NC | 0.275910433 |
| | No. of Succeeds | 30 | 30 | 30 | 0 | 30 |
| | No. of Failures | 0 | 0 | 0 | 30 | 0 |
| F11 | Mean | 3761.9 | 4211.533333 | NC | NC | NC |
| | S.D. | 503.0939687 | 835.0827642 | NC | NC | NC |
| | Best | 2890 | 3339 | NC | NC | NC |
| | Worst | 4679 | 7001 | NC | NC | NC |
| | Exec. time | 1.841760067 | 6.437364733 | NC | NC | NC |
| | No. of Succeeds | 30 | 30 | 0 | 0 | 0 |
| | No. of Failures | 0 | 0 | 30 | 30 | 30 |
| F12 | Mean | 1518.7 | NC | NC | NC | NC |
| | S.D. | 520.5891987 | NC | NC | NC | NC |
| | Best | 871 | NC | NC | NC | NC |
| | Worst | 2901 | NC | NC | NC | NC |
| | Exec. time | 0.457080213 | NC | NC | NC | NC |
| | No. of Succeeds | 30 | 0 | 0 | 0 | 0 |
| | No. of Failures | 0 | 30 | 30 | 30 | 30 |
| F13 | Mean | NC | NC | NC | NC | NC |
| | S.D. | NC | NC | NC | NC | NC |
| | Best | NC | NC | NC | NC | NC |
| | Worst | NC | NC | NC | NC | NC |
| | Exec. time | NC | NC | NC | NC | NC |
| | No. of Succeeds | 0 | 0 | 0 | 0 | 0 |
| | No. of Failures | 30 | 30 | 30 | 30 | 30 |
| F14 | Mean | 417 | 484.4333333 | NC | 714.3 | 54.36666667 |
| | S.D. | 56.51914537 | 7.219147592 | NC | 77.84739714 | 2.85854235 |
| | Best | 303 | 469 | NC | 568 | 50 |
| | Worst | 531 | 497 | NC | 834 | 63 |
| | Exec. time | 0.2572861 | 0.84180583 3 | NC | 2.2013179 | 0.5315005 |
| | No. of Succeeds | 30 | 30 | 0 | 30 | 30 |
| | No. of Failures | 0 | 0 | 30 | 0 | 0 |
| F15 | Mean | NC | NC | NC | NC | NC |
| | S.D. | NC | NC | NC | NC | NC |
| | Best | NC | NC | NC | NC | NC |
| | Worst | NC | NC | NC | NC | NC |
| | Exec. time | NC | NC | NC | NC | NC |
| | No. of Succeeds | 0 | 0 | 0 | 0 | 0 |
| | No. of Failures | 30 | 30 | 30 | 30 | 30 |
| F16 | Mean | NC | 991 | NC | NC | NC |
| | S.D. | NC | 1078.5 | NC | NC | NC |
| | Best | NC | 52.5988247 4 | NC | NC | NC |
| | Worst | NC | 991 | NC | NC | NC |
| | Exec. time | NC | 1200 | NC | NC | NC |
| | No. of Succeeds | 0 | 12 | 0 | 0 | 0 |
| | No. of Failures | 30 | 18 | 30 | 30 | 30 |

**Table (B.3): Basic statistics of the 30-solutions obtained by BSA, DE, PSO, ABC, and FF in Test 1 (default search space with Nvar3 dimensions)**

| **Functions** | **Statistics** | **BSA** | **DE** | **PSO** | **ABC** | **FF** |
|---|---|---|---|---|---|---|
| F1 | Mean | 2212.03333 3 | 1265.4 | NC | NC | 409.8 |
| | S.D. | 285.810340 2 | 17.2938099 4 | NC | NC | 2.57842507 4 |
| | Best | 1789 | 1231 | NC | NC | 404 |
| | Worst | 2705 | 1297 | NC | NC | 414 |
| | Exec. time | 1.2222052 | 4.16114533 3 | NC | NC | 6.67840653 3 |
| | No. of Succeeds | 30 | 30 | 0 | 0 | 30 |
| | No. of Failures | 0 | 0 | 30 | 30 | 0 |
| F2 | Mean | 1437.03333 3 | 1233.33333 3 | NC | NC | 453.466666 7 |
| | S.D. | 192.166501 7 | 188.052144 9 | NC | NC | 26.0738121 1 |
| | Best | 1129 | 907 | NC | NC | 411 |
| | Worst | 1982 | 1619 | NC | NC | 542 |
| | Exec. time | 0.94669963 3 | 2.58951793 3 | NC | NC | 9.55394106 7 |
| | No. of Succeeds | 30 | 30 | 0 | 0 | 30 |
| | No. of Failures | 0 | 0 | 30 | 30 | 0 |
| F3 | Mean | 364.7666667 | 186.8666667 | 53.80769231 | 38.33333333 | 28.26666667 |
| | S.D. | 73.52535726 | 51.15475731 | 5.592990118 | 8.326663998 | 19.5711493 |
| | Best | 242 | 124 | 39 | 21 | 4 |
| | Worst | 525 | 284 | 61 | 56 | 81 |
| | Exec. time | 0.217162567 | 0.4063774 | 0.105839692 | 0.0705598 | 0.603975867 |
| | No. of Succeeds | 30 | 30 | 26 | 30 | 30 |
| | No. of Failures | 0 | 0 | 4 | 0 | 0 |
| F4 | Mean | 525.7 | 6225.6 | 909.4 | 84.7037037 | 1489.633333 |
| | S.D. | 218.5027657 | 461.136198 | 641.2112245 | 43.79481425 | 32.82606548 |
| | Best | 284 | 5442 | 6 | 23 | 1409 |
| | Worst | 1269 | 7563 | 1837 | 186 | 1531 |
| | Exec. time | 0.2490285 | 14.56781393 | 2.5651712 | 0.266280556 | 20.34113953 |
| | No. of Succeeds | 30 | 30 | 30 | 27 | 30 |
| | No. of Failures | 0 | 0 | 0 | 3 | 0 |
| F5 | Mean | 173.5333333 | 45.3 | 24.2 | 14.53333333 | 3.666666667 |
| | S.D. | 39.28299916 | 10.16807038 | 10.69772776 | 4.904138529 | 0.884086645 |
| | Best | 82 | 25 | 6 | 2 | 2 |
| | Worst | 234 | 64 | 41 | 23 | 5 |
| | Exec. time | 0.109724933 | 0.132266967 | 0.043759533 | 0.045434933 | 0.051644167 |
| | No. of Succeeds | 30 | 30 | 30 | 30 | 30 |
| | No. of Failures | 0 | 0 | 0 | 0 | 0 |
| F6 | Mean | 996.1 | 296.5666667 | 64.16666667 | NC | 113.7333333 |
| | S.D. | 88.20757259 | 47.62691454 | 4.609460536 | NC | 26.92509832 |
| | Best | 861 | 209 | 55 | NC | 56 |
| | Worst | 1197 | 408 | 76 | NC | 168 |
| | Exec. time | 0.4946741 | 0.879912567 | 0.118392067 | NC | 1.790483167 |
| | No. of Succeeds | 30 | 30 | 30 | 0 | 30 |
| | No. of Failures | 0 | 0 | 0 | 30 | 0 |
| F7 | Mean | NC | NC | NC | NC | NC |
| | S.D. | NC | NC | NC | NC | NC |
| | Best | NC | NC | NC | NC | NC |
| | Worst | NC | NC | NC | NC | NC |
| | Exec. time | NC | NC | NC | NC | NC |
| | No. of Succeeds | 0 | 0 | 0 | 0 | 0 |
| | No. of Failures | 30 | 30 | 30 | 30 | 30 |
| F8 | Mean | 118.2 | 3750.9 | 637.0333333 | 15.03571429 | 4.5 |
| | S.D. | 27.9425519 | 44.63055239 | 6.392254652 | 2.486737307 | 0.989949494 |
| | Best | 78 | 3643 | 624 | 10 | 2 |
| | Worst | 190 | 3811 | 649 | 20 | 6 |
| | Exec. time | 0.169004103 | 6.867129 | 1.4425986 | 0.029832 | 0.073269769 |
| | No. of Succeeds | 30 | 30 | 30 | 28 | 26 |
| | No. of Failures | 0 | 0 | 0 | 2 | 4 |
| F9 | Mean | 965.9666667 | 2588.033333 | NC | NC | NC |
| | S.D. | 76.88974791 | 72.99715787 | NC | NC | NC |
| | Best | 855 | 2514 | NC | NC | NC |
| | Worst | 1150 | 2898 | NC | NC | NC |
| | Exec. time | 0.582512467 | 5.4244718 | NC | NC | NC |
| | No. of Succeeds | 30 | 30 | 0 | 0 | 0 |
| | No. of Failures | 0 | 0 | 30 | 30 | 30 |
| F10 | Mean | 397.2333333 | 92.83333333 | 56.66666667 | NC | 20.83333333 |
| | S.D. | 43.98172295 | 15.20681184 | 5.168427583 | NC | 10.19493902 |
| | Best | 309 | 41 | 49 | NC | 7 |
| | Worst | 471 | 114 | 67 | NC | 38 |
| | Exec. time | 9565.148219 | 0.241654933 | 0.099469833 | NC | 0.284334333 |
| | No. of Succeeds | 30 | 30 | 30 | 0 | 30 |
| | No. of Failures | 0 | 0 | 0 | 30 | 0 |
| F11 | Mean | NC | NC | NC | NC | NC |
| | S.D. | NC | NC | NC | NC | NC |
| | Best | NC | NC | NC | NC | NC |
| | Worst | NC | NC | NC | NC | NC |
| | Exec. time | NC | NC | NC | NC | NC |
| | No. of Succeeds | 0 | 0 | 0 | 0 | 0 |
| | No. of Failures | 30 | 30 | 30 | 30 | 30 |
| F12 | Mean | 16054.2 | NC | NC | NC | NC |
| | S.D. | 22085.31046 | NC | NC | NC | NC |
| | Best | 9065 | NC | NC | NC | NC |
| | Worst | 132801 | NC | NC | NC | NC |
| | Exec. time | 6.478484033 | NC | NC | NC | NC |
| | No. of Succeeds | 30 | 0 | 0 | 0 | 0 |
| | No. of Failures | 0 | 30 | 30 | 30 | 30 |
| F13 | Mean | NC | NC | NC | NC | NC |
| | S.D. | NC | NC | NC | NC | NC |
| | Best | NC | NC | NC | NC | NC |
| | Worst | NC | NC | NC | NC | NC |
| | Exec. time | NC | NC | NC | NC | NC |
| | No. of Succeeds | 0 | 0 | 0 | 0 | 0 |
| | No. of Failures | 30 | 30 | 30 | 30 | 30 |
| F14 | Mean | 943.9 | 1060 | NC | 6468.933333 | 113.1333333 |
| | S.D. | 84.9061754 | 15.59840841 | NC | 311.951912 | 3.329422994 |
| | Best | 798 | 1034 | NC | 5878 | 106 |
| | Worst | 1109 | 1097 | NC | 7116 | 120 |
| | Exec. time | 0.621603966 | 3.001068333 | NC | 19.45626283 | 1.025718967 |
| | No. of Succeeds | 30 | 30 | 0 | 30 | 30 |
| | No. of Failures | 0 | 0 | 30 | 0 | 0 |
| F15 | Mean | NC | NC | NC | NC | NC |
| | S.D. | NC | NC | NC | NC | NC |
| | Best | NC | NC | NC | NC | NC |
| | Worst | NC | NC | NC | NC | NC |
| | Exec. time | NC | NC | NC | NC | NC |
| | No. of Succeeds | 0 | 0 | 0 | 0 | 0 |
| | No. of Failures | 30 | 30 | 30 | 30 | 30 |
| F16 | Mean | NC | NC | NC | NC | NC |
| | S.D. | NC | NC | NC | NC | NC |
| | Best | NC | NC | NC | NC | NC |
| | Worst | NC | NC | NC | NC | NC |
| | Exec. time | NC | NC | NC | NC | NC |
| | No. of Succeeds | 0 | 0 | 0 | 0 | 0 |
| | No. of Failures | 30 | 30 | 30 | 30 | 30 |

In addition, Tables (B.4), (B.5), and (B.6) present basic statistics (Mean: mean-solution, S.D.: standard-deviation of mean-solution, Best: the best solution, Worst: the worst solution, Exec. Time: mean runtime in seconds, No. of succeeds: number of successful minimization, and No. of Failure: number of failed minimization) of 30-solutions obtained by BSA, DE, PSO, ABC, and FF in Test 2 to minimize F1-F16 functions with default number of variables in three different search spaces R1, R2, and R3 respectively.

**Table (B.4): Basic statistics of the 30-solutions obtained by BSA, DE, PSO, ABC, and FF in Test 2 (two-variable dimensions with R1)**

| Functions | Statistics | BSA | DE | PSO | ABC | FF |
|---|---|---|---|---|---|---|
| F1 | Mean | 72.33333333 | NC | 0 | 19.33333333 | 60.36666667 |
| | S.D. | 16.00287331 | NC | NC | 2.368374 | 23.48217715 |
| | Best | 45 | NC | NC | 14 | 5 |
| | Worst | 101 | NC | NC | 23 | 95 |
| | Exec. time | 0.028775833 | NC | NC | 0.159118567 | 1.193958333 |
| | No. of Succeeds | 30 | 0 | NC | 30 | 30 |
| | No. of Failures | 0 | 30 | 0 | 0 | 0 |
| F2 | Mean | 39.5 | 15.26666667 | 794.25 | 9.133333333 | 4.033333333 |
| | S.D. | 11.70838191 | 2.935318033 | 379.0831654 | 1.907034769 | 0.808716878 |
| | Best | 22 | 10 | 154 | 4 | 2 |
| | Worst | 76 | 22 | 1146 | 13 | 5 |
| | Exec. time | 0.00711624 | 0.045706 | 1.445961536 | 0.0771137 | 0.064563133 |
| | No. of Succeeds | 30 | 30 | 28 | 30 | 30 |
| | No. of Failures | 0 | 0 | 2 | 0 | 0 |
| F3 | Mean | 139.5 | 53.33333333 | 53.75 | NC | 31.63333333 |
| | S.D. | 28.07717198 | 14.21347911 | 9.777619948 | NC | 12.64224809 |
| | Best | 73 | 32 | 40 | NC | 9 |
| | Worst | 206 | 79 | 97 | NC | 59 |
| | Exec. time | 0.076975133 | 0.084473933 | 0.092407821 | NC | 0.493045367 |
| | No. of Succeeds | 30 | 30 | 28 | 0 | 30 |
| | No. of Failures | 0 | 0 | 2 | 30 | 0 |
| F4 | Mean | 377.1333333 | 89.69230769 | 1213.866667 | 16 | 9.269230769 |
| | S.D. | 68.4693404 | 33.18405549 | 159.6013856 | 47.125 | 3.053623321 |
| | Best | 286 | 28 | 926 | 33.18369492 | 5 |
| | Worst | 649 | 191 | 1550 | 6 | 19 |
| | Exec. time | 0.200669233 | 0.293420846 | 2.0382361 | 147 | 0.126300154 |
| | No. of Succeeds | 30 | 26 | 30 | 0.281062708 | 26 |
| | No. of Failures | 0 | 4 | 0 | 24 | 4 |
| F5 | Mean | 45.56666667 | 15.03333333 | 21.8 | 7.433333333 | 3.133333333 |
| | S.D. | 17.02469471 | 5.18275212 | 9.459168149 | 3.339247627 | 0.730296743 |
| | Best | 13 | 5 | 3 | 2 | 2 |
| | Worst | 76 | 24 | 37 | 13 | 4 |
| | Exec. time | 0.019755067 | 0.032688133 | 0.039887567 | 0.077219 | 0.0421937 |
| | No. of Succeeds | 30 | 30 | 30 | 30 | 30 |
| | No. of Failures | 0 | 0 | 0 | 0 | 0 |
| F6 | Mean | 248.2666667 | 30.86666667 | 794.25 | 20.3 | 11.96666667 |
| | S.D. | 31.22325982 | 4.108303898 | 379.0831654 | 1.600646421 | 7.716901886 |
| | Best | 202 | 25 | 154 | 17 | 4 |
| | Worst | 311 | 45 | 1146 | 24 | 34 |
| | Exec. time | 0.0888098 | 0.051433367 | 0.0724958 | 0.214975967 | 0.094171033 |
| | No. of Succeeds | 30 | 30 | 28 | 30 | 30 |
| | No. of Failures | 0 | 0 | 2 | 0 | 0 |
| F7 | Mean | 64.36666667 | 139.5714286 | NC | 27.6 | 249.9473684 |
| | S.D. | 12.86396553 | 16.33997274 | NC | 22.41568999 | 26.50885321 |
| | Best | 41 | 103 | NC | 9 | 193 |
| | Worst | 91 | 170 | NC | 89 | 282 |
| | Exec. time | 0.038450533 | 0.427799464 | NC | 0.25814615 | 3.344931158 |
| | No. of Succeeds | 30 | 28 | 0 | 20 | 19 |
| | No. of Failures | 0 | 2 | 30 | 10 | 11 |
| F8 | Mean | 248.5666667 | 18.48148148 | 636.9 | 12.24 | 4.851851852 |
| | S.D. | 23.91894839 | 2.359438825 | 5.984750737 | 2.241279397 | 0.948833442 |
| | Best | 206 | 15 | 627 | 8 | 4 |
| | Worst | 300 | 23 | 651 | 16 | 7 |
| | Exec. time | 0.042607433 | 0.032870519 | 0.713972767 | 0.09773984 | 0.068419926 |
| | No. of Succeeds | 30 | 27 | 30 | 25 | 27 |
| | No. of Failures | 0 | 3 | 0 | 5 | 3 |
| F9 | Mean | 25.23333333 | 166.5769231 | 100.7142857 | 9.076923077 | 539.3793103 |
| | S.D. | 7.833100999 | 110.8112532 | 5.13552591 | 5.483962632 | 45.94671898 |
| | Best | 14 | 89 | 92 | 2 | 413 |
| | Worst | 47 | 691 | 109 | 19 | 597 |
| | Exec. time | 0.016005467 | 0.554092731 | 0.162314571 | 0.046853615 | 6.89237431 |
| | No. of Succeeds | 30 | 26 | 14 | 26 | 29 |
| | No. of Failures | 0 | 4 | 16 | 4 | 1 |
| F10 | Mean | 36.9 | 60.15 | NC | NC | 5.333333333 |
| | S.D. | 17.7906407 | 179.1662783 | NC | NC | 1.688364508 |
| | Best | 16 | 14 | NC | NC | 3 |
| | Worst | 79 | 821 | NC | NC | 10 |
| | Exec. time | 0.029989533 | 0.17157865 | NC | NC | 0.084701533 |
| | No. of Succeeds | 30 | 20 | 0 | 0 | 30 |
| | No. of Failures | 0 | 10 | 30 | 30 | 0 |
| F11 | Mean | 111.3666667 | 81.86666667 | 105.5 | 29.22222222 | 9.96 |
| | S.D. | 16.08486831 | 3.559671947 | 4.289120157 | 6.405126152 | 3.813135193 |
| | Best | 79 | 76 | 93 | 16 | 4 |
| | Worst | 147 | 89 | 117 | 45 | 20 |
| | Exec. time | 0.034564033 | 0.277437967 | 0.1759911 | 0.152298889 | 0.14458784 |
| | No. of Succeeds | 30 | 30 | 30 | 27 | 25 |
| | No. of Failures | 0 | 0 | 0 | 3 | 5 |
| F12 | Mean | 157.9333333 | 60.77272727 | 722.8 | 35.88888889 | 8.083333333 |
| | S.D. | 35.35332334 | 24.92069673 | 104.1181027 | 24.95585847 | 2.019829237 |
| | Best | 98 | 8 | 414 | 5 | 5 |
| | Worst | 236 | 103 | 857 | 94 | 13 |
| | Exec. time | 0.145922333 | 0.145608045 | 1.219939367 | 0.322523741 | 0.111710708 |
| | No. of Succeeds | 30 | 22 | 30 | 27 | 24 |
| | No. of Failures | 0 | 8 | 0 | 3 | 6 |
| F13 | Mean | 271.8 | 147.137931 | 646.36 | 133.7666667 | 10.92592593 |
| | S.D. | 40.4086027 | 52.0286488 | 5.321967055 | 100.097774 | 2.840990157 |
| | Best | 201 | 57 | 634 | 27 | 6 |
| | Worst | 344 | 260 | 656 | 349 | 17 |
| | Exec. time | 0.051252967 | 0.229098862 | 1.19053768 | 1.2913116 | 0.168615333 |
| | No. of Succeeds | 30 | 29 | 25 | 30 | 27 |
| | No. of Failures | 0 | 1 | 5 | 0 | 3 |
| F14 | Mean | 232.7666667 | 10.2 | 637.5 | 6 | 2.68 |
| | S.D. | 20.00979645 | 2.483630619 | 5.544490897 | 1.560378995 | 0.476095229 |
| | Best | 190 | 6 | 626 | 3 | 2 |
| | Worst | 275 | 14 | 654 | 9 | 3 |
| | Exec. time | 0.077608533 | 0.034451 | 1.2173071 | 0.05095925 | 0.033408 |
| | No. of Succeeds | 30 | 20 | 30 | 24 | 25 |
| | No. of Failures | 0 | 10 | 0 | 6 | 5 |
| F15 | Mean | 52.06666667 | 22.43333333 | 44.16666667 | 19.1 | 11.7 |
| | S.D. | 21.2244708 | 2.51455533 | 6.454527058 | 3.262641198 | 5.408486307 |
| | Best | 21 | 18 | 27 | 10 | 3 |
| | Worst | 90 | 27 | 52 | 25 | 23 |
| | Exec. time | 0.031657633 | 0.0592526 | 0.0887543 | 0.138674967 | 0.151984633 |
| | No. of Succeeds | 30 | 30 | 30 | 30 | 30 |
| | No. of Failures | 0 | 0 | 0 | 0 | 0 |
| F16 | Mean | NC | NC | NC | NC | NC |
| | S.D. | NC | NC | NC | NC | NC |
| | Best | NC | NC | NC | NC | NC |
| | Worst | NC | NC | NC | NC | NC |
| | Exec. time | NC | NC | NC | NC | NC |
| | No. of Succeeds | 0 | 0 | 0 | 0 | 0 |
| | No. of Failures | 30 | 30 | 30 | 30 | 30 |

**Table (B.5): Basic statistics of the 30-solutions obtained by BSA, DE, PSO, ABC, and FF in Test 2 (two variable dimensions with R2)**

| **Functions** | **Statistics** | **BSA** | **DE** | **PSO** | **ABC** | **FF** |
|---|---|---|---|---|---|---|
| F1 | Mean | 425.7333333 | NC | NC | NC | 249.4137931 |
| | S.D. | 65.63111822 | NC | NC | NC | 24.0735736 |
| | Best | 338 | NC | NC | NC | 192 |
| | Worst | 535 | NC | NC | NC | 295 |
| | Exec. time | 0.129019233 | NC | NC | NC | 5.009911103 |
| | No. of Succeeds | 30 | 0 | 0 | 0 | 29 |
| | No. of Failures | 0 | 30 | 30 | 30 | 1 |
| F2 | Mean | 156.0333333 | 39.10344828 | 732.0740741 | 38.33333333 | 61 |
| | S.D. | 16.09558374 | 5.1084302 | 407.9909431 | 6.199740447 | 22.28486896 |
| | Best | 126 | 26 | 166 | 27 | 11 |
| | Worst | 188 | 49 | 1158 | 49 | 93 |
| | Exec. time | 0.039772367 | 0.112101655 | 1.457569407 | 0.197538433 | 0.884010333 |
| | No. of Succeeds | 30 | 29 | 27 | 30 | 27 |
| | No. of Failures | 0 | 1 | 3 | 0 | 3 |
| F3 | Mean | 1087.666667 | 233.3888889 | 81.73333333 | NC | 208.0333333 |
| | S.D. | 391.2479885 | 71.84144215 | 6.356822744 | NC | 33.0146084 |
| | Best | 293 | 103 | 72 | NC | 117 |
| | Worst | 1719 | 376 | 97 | NC | 250 |
| | Exec. time | 0.1936562 | 0.3856845 | 0.1286414 | NC | 3.080032433 |
| | No. of Succeeds | 30 | 18 | 30 | 0 | 30 |
| | No. of Failures | 0 | 12 | 0 | 30 | 0 |
| F4 | Mean | 419.7666667 | 194.0714286 | 1253.52381 | 219.6896552 | 93.8 |
| | S.D. | 191.3771657 | 65.9005584 | 154.2182282 | 132.8401573 | 24.72515588 |
| | Best | 189 | 64 | 981 | 33 | 47 |
| | Worst | 898 | 367 | 1517 | 636 | 145 |
| | Exec. time | 0.263658433 | 0.674849321 | 2.085545905 | 1.637981655 | 1.34841468 |
| | No. of Succeeds | 30 | 28 | 21 | 29 | 25 |
| | No. of Failures | 0 | 2 | 9 | 1 | 5 |
| F5 | Mean | 99.93333333 | 29 | 54 | NC | 40.3 |
| | S.D. | 22.2539161 | 3.695290572 | 10.22168081 | NC | 26.1786304 |
| | Best | 53 | 22 | 28 | NC | 3 |
| | Worst | 143 | 36 | 68 | NC | 85 |
| | Exec. time | 0.043615233 | 0.0518579 | 0.0923664 | NC | 0.349824067 |
| | No. of Succeeds | 30 | 30 | 30 | 0 | 30 |
| | No. of Failures | 0 | 0 | 0 | 30 | 0 |
| F6 | Mean | 671.8 | 230.1333333 | 75.13333333 | NC | NC |
| | S.D. | 107.4107167 | 48.5107406 | 5.888231005 | NC | NC |
| | Best | 424 | 162 | 61 | NC | NC |
| | Worst | 883 | 341 | 91 | NC | NC |
| | Exec. time | 0.229059867 | 0.646945533 | 0.120696033 | NC | NC |
| | No. of Succeeds | 30 | 30 | 30 | 0 | 0 |
| | No. of Failures | 0 | 0 | 0 | 30 | 30 |
| F7 | Mean | 112.6333333 | 157.6551724 | NC | 30.61111111 | 445.8666667 |
| | S.D. | 30.11527661 | 16.00061575 | NC | 9.798792776 | 33.75033418 |
| | Best | 84 | 125 | NC | 16 | 343 |
| | Worst | 232 | 186 | NC | 58 | 499 |
| | Exec. time | 0.0441853 | 0.534595448 | NC | 0.272917111 | 5.854646267 |
| | No. of Succeeds | 30 | 29 | 0 | 18 | 30 |
| | No. of Failures | 0 | 1 | 30 | 12 | 0 |
| F8 | Mean | 328.6666667 | 32.08 | 644.2333333 | 23.76 | 110.7666667 |
| | S.D. | 31.07425293 | 3.414674216 | 5.992428172 | 3.3326666 | 24.39712156 |
| | Best | 261 | 24 | 633 | 17 | 53 |
| | Worst | 419 | 37 | 661 | 29 | 151 |
| | Exec. time | 0.0566179 | 0.05600776 | 1.142361633 | 0.18997352 | 1.5272697 |
| | No. of Succeeds | 30 | 25 | 30 | 25 | 30 |
| | No. of Failures | 0 | 5 | 0 | 5 | 0 |
| F9 | Mean | 181.5333333 | 175.8 | 794.25 | 201.5769231 | 742.3103448 |
| | S.D. | 44.83743305 | 19.91256751 | 379.0831654 | 144.6715378 | 47.85178266 |
| | Best | 104 | 145 | 154 | 32 | 547 |
| | Worst | 262 | 213 | 1146 | 488 | 807 |
| | Exec. time | 0.1036287 | 0.459873733 | 0.419778435 | 0.98108872 | 9.115128828 |
| | No. of Succeeds | 30 | 30 | 28 | 26 | 29 |
| | No. of Failures | 0 | 0 | 2 | 4 | 1 |
| F10 | Mean | NC | NC | NC | NC | NC |
| | S.D. | NC | NC | NC | NC | NC |
| | Best | NC | NC | NC | NC | NC |
| | Worst | NC | NC | NC | NC | NC |
| | Exec. time | NC | NC | NC | NC | NC |
| | No. of Succeeds | 0 | 0 | 0 | 0 | 0 |
| | No. of Failures | 30 | 30 | 30 | 30 | 30 |
| F11 | Mean | 138.1666667 | 96.83333333 | 122.4827586 | 43 | 149.8148148 |
| | S.D. | 28.06652606 | 4.177801215 | 5.925855512 | 7.541883054 | 38.67421824 |
| | Best | 94 | 85 | 114 | 31 | 50 |
| | Worst | 201 | 103 | 142 | 55 | 210 |
| | Exec. time | 0.059349433 | 0.282379867 | 0.199407138 | 0.424502846 | 2.34202537 |
| | No. of Succeeds | 30 | 30 | 29 | 26 | 27 |
| | No. of Failures | 0 | 0 | 1 | 4 | 3 |
| F12 | Mean | 572.8333333 | 368.2222222 | 936.1 | 409.5882353 | 80.37037037 |
| | S.D. | 122.3739558 | 99.16549234 | 243.0311957 | 225.0005719 | 27.80999841 |
| | Best | 411 | 210 | 605 | 55 | 19 |
| | Worst | 934 | 565 | 1858 | 944 | 120 |
| | Exec. time | 0.310757733 | 1.056444556 | 1.5107343 | 3.453786647 | 1.234736259 |
| | No. of Succeeds | 30 | 27 | 30 | 17 | 27 |
| | No. of Failures | 0 | 3 | 0 | 13 | 3 |
| F13 | Mean | 427.8666667 | 179.8333333 | 764.962963 | 204.9259259 | 118.4333333 |
| | S.D. | 68.73172902 | 39.81689412 | 229.5504774 | 118.616034 | 25.96374572 |
| | Best | 311 | 80 | 647 | 23 | 56 |
| | Worst | 601 | 270 | 1392 | 457 | 163 |
| | Exec. time | 0.123285933 | 0.2651661 | 1.240302815 | 1.956210444 | 1.893014833 |
| | No. of Succeeds | 30 | 30 | 27 | 27 | 30 |
| | No. of Failures | 0 | 0 | 3 | 3 | 0 |
| F14 | Mean | 269.0666667 | 23.85185185 | 642.8333333 | 17.16666667 | 41.35714286 |
| | S.D. | 29.88706713 | 3.134224278 | 6.772578246 | 1.340560125 | 15.61626475 |
| | Best | 206 | 18 | 630 | 15 | 14 |
| | Worst | 339 | 29 | 658 | 20 | 74 |
| | Exec. time | 0.0919827 | 0.066911815 | 1.159890367 | 0.1416615 | 0.626989893 |
| | No. of Succeeds | 30 | 27 | 30 | 24 | 28 |
| | No. of Failures | 0 | 3 | 0 | 6 | 2 |
| F15 | Mean | 135.0333333 | 36.86666667 | 68.33333333 | 31.33333333 | 163.8 |
| | S.D. | 27.06758251 | 3.702127378 | 3.950847428 | 3.283536081 | 25.20180618 |
| | Best | 85 | 27 | 58 | 24 | 87 |
| | Worst | 233 | 44 | 74 | 38 | 207 |
| | Exec. time | 0.033026633 | 0.087536133 | 0.1189461 | 0.236676 | 1.600936967 |
| | No. of Succeeds | 30 | 30 | 30 | 30 | 30 |
| | No. of Failures | 0 | 0 | 0 | 0 | 0 |
| F16 | Mean | NC | NC | NC | NC | NC |
| | S.D. | NC | NC | NC | NC | NC |
| | Best | NC | NC | NC | NC | NC |
| | Worst | NC | NC | NC | NC | NC |
| | Exec. time | NC | NC | NC | NC | NC |
| | No. of Succeeds | 0 | 0 | 0 | 0 | 0 |
| | No. of Failures | 30 | 30 | 30 | 30 | 30 |

**Table (B.6): Basic statistics of the 30-solutions obtained by BSA, DE, PSO, ABC, and FF in Test 2 (two-variable dimensions with R3)**

| Functions | Statistics | BSA | DE | PSO | ABC | FF |
|---|---|---|---|---|---|---|
| F1 | Mean | 556.6333333 | NC | NC | NC | 279.75 |
| | S.D. | 112.7296073 | NC | NC | NC | 33.31962437 |
| | Best | 396 | NC | NC | NC | 171 |
| | Worst | 898 | NC | NC | NC | 315 |
| | Exec. time | 0.189253867 | NC | NC | NC | 5.1484892 |
| | No. of Succeeds | 30 | 0 | 0 | 0 | 20 |
| | No. of Failures | 0 | 30 | 30 | 30 | 10 |
| F2 | Mean | 154.6666667 | 46.16666667 | 847.8571429 | 47.46666667 | 88.46666667 |
| | S.D. | 24.94730078 | 6.411645296 | 367.6544991 | 12.19308264 | 26.17017694 |
| | Best | 113 | 27 | 165 | 32 | 36 |
| | Worst | 201 | 63 | 1152 | 95 | 141 |
| | Exec. time | 0.067592267 | 0.139866267 | 1.5282565 | 0.340741933 | 1.4476881 |
| | No. of Succeeds | 30 | 30 | 28 | 30 | 30 |
| | No. of Failures | 0 | 0 | 2 | 0 | 0 |
| F3 | Mean | 1909.433333 | 182.75 | 84.46666667 | NC | 249.6 |
| | S.D. | 591.0772564 | 82.99949799 | 10.78547341 | NC | 29.66897834 |
| | Best | 1402 | 112 | 67 | NC | 159 |
| | Worst | 3725 | 302 | 118 | NC | 294 |
| | Exec. time | 0.3167525 | 0.139866267 | 0.127532267 | NC | 3.4745561 |
| | No. of Succeeds | 30 | 4 | 30 | 0 | 30 |
| | No. of Failures | 0 | 26 | 0 | 30 | 0 |
| F4 | Mean | 444.0666667 | 220.0357143 | 1482.153846 | 238.2592593 | 118.6153846 |
| | S.D. | 81.262 | 62.51487654 | 824.6896479 | 120.881249 | 31.97258441 |
| | Best | 309 | 100 | 955 | 45 | 51 |
| | Worst | 659 | 365 | 5332 | 415 | 180 |
| | Exec. time | 0.179457933 | 0.7116585 | 1.445961536 | 2.140455593 | 1.692470923 |
| | No. of Succeeds | 30 | 28 | 26 | 27 | 26 |
| | No. of Failures | 0 | 2 | 4 | 3 | 4 |
| F5 | Mean | NC | NC | NC | NC | NC |
| | S.D. | NC | NC | NC | NC | NC |
| | Best | NC | NC | NC | NC | NC |
| | Worst | NC | NC | NC | NC | NC |
| | Exec. time | NC | NC | NC | NC | NC |
| | No. of Succeeds | 0 | 0 | 0 | 0 | 0 |
| | No. of Failures | 30 | 30 | 30 | 30 | 30 |
| F6 | Mean | 1096.466667 | 570.2666667 | 82.76666667 | NC | NC |
| | S.D. | 263.5093054 | 179.2411078 | 8.071590594 | NC | NC |
| | Best | 691 | 295 | 66 | NC | NC |
| | Worst | 1590 | 984 | 100 | NC | NC |
| | Exec. time | 0.3102513 | 1.701697967 | 0.111516034 | NC | NC |
| | No. of Succeeds | 30 | 30 | 30 | 0 | 0 |
| | No. of Failures | 0 | 0 | 0 | 30 | 30 |
| F7 | Mean | 165.1333333 | 158.3103448 | NC | 36.83333333 | 474.3 |
| | S.D. | 44.85281165 | 18.3655412 | NC | 20.98809186 | 31.4896261 |
| | Best | 98 | 117 | NC | 22 | 409 |
| | Worst | 292 | 188 | NC | 116 | 528 |
| | Exec. time | 0.107162467 | 0.423307759 | NC | 0.336140889 | 6.128933267 |
| | No. of Succeeds | 30 | 29 | 0 | 18 | 30 |
| | No. of Failures | 0 | 1 | 30 | 12 | 0 |
| F8 | Mean | 319.0666667 | 35.30769231 | 643.8333333 | 27.2 | 145.3043478 |
| | S.D. | 38.74935669 | 2.412786452 | 6.198349799 | 2.645751311 | 19.00104012 |
| | Best | 271 | 31 | 634 | 22 | 115 |
| | Worst | 401 | 40 | 657 | 31 | 182 |
| | Exec. time | 0.0543112 | 0.060988577 | 1.248304667 | 0.21520208 | 2.085555391 |
| | No. of Succeeds | 30 | 26 | 30 | 25 | 23 |
| | No. of Failures | 0 | 4 | 0 | 5 | 7 |
| F9 | Mean | 145.1666667 | 176.3 | 370.7083333 | 179.9615385 | 789.7666667 |
| | S.D. | 36.97257449 | 17.78831465 | 400.0615691 | 118.135001 | 27.78945885 |
| | Best | 99 | 147 | 118 | 29 | 725 |
| | Worst | 254 | 207 | 1277 | 460 | 842 |
| | Exec. time | 0.08441 | 0.555940233 | 0.588364917 | 0.943245115 | 8.9030898 |
| | No. of Succeeds | 30 | 30 | 24 | 26 | 30 |
| | No. of Failures | 0 | 0 | 6 | 4 | 0 |
| F10 | Mean | NC | NC | NC | NC | NC |
| | S.D. | NC | NC | NC | NC | NC |
| | Best | NC | NC | NC | NC | NC |
| | Worst | NC | NC | NC | NC | NC |
| | Exec. time | NC | NC | NC | NC | NC |
| | No. of Succeeds | 0 | 0 | 0 | 0 | 0 |
| | No. of Failures | 30 | 30 | 30 | 30 | 30 |
| F11 | Mean | 143.9333333 | 98.93333333 | 129.9333333 | 45.42857143 | 195.5517241 |
| | S.D. | 38.00901285 | 4.555658349 | 19.80061534 | 7.593508785 | 30.62513698 |
| | Best | 89 | 88 | 115 | 25 | 91 |
| | Worst | 211 | 108 | 227 | 60 | 242 |
| | Exec. time | 0.1755195 | 0.305618167 | 0.2071013 | 0.423829429 | 3.17409831 |
| | No. of Succeeds | 30 | 30 | 30 | 28 | 29 |
| | No. of Failures | 0 | 0 | 0 | 2 | 1 |
| F12 | Mean | 1379.166667 | 427.9230769 | 898.9 | 569.5714286 | 116.5 |
| | S.D. | 156.3588012 | 117.8514058 | 254.7484696 | 246.4879018 | 38.30820432 |
| | Best | 1101 | 263 | 585 | 342 | 24 |
| | Worst | 1879 | 667 | 1753 | 932 | 181 |
| | Exec. time | 0.4534084 | 0.7652505 | 1.454673867 | 4.816096714 | 1.725299179 |
| | No. of Succeeds | 30 | 26 | 30 | 7 | 28 |
| | No. of Failures | 0 | 4 | 0 | 23 | 2 |
| F13 | Mean | 381.5666667 | 171.5333333 | 700.1785714 | 224.8076923 | 154.1 |
| | S.D. | 102.9180066 | 53.29147783 | 134.0732397 | 135.1523642 | 29.25553293 |
| | Best | 205 | 75 | 645 | 46 | 76 |
| | Worst | 756 | 276 | 1374 | 456 | 199 |
| | Exec. time | 0.1637319 | 0.519127567 | 1.213026786 | 2.209036731 | 2.474901667 |
| | No. of Succeeds | 30 | 30 | 28 | 26 | 30 |
| | No. of Failures | 0 | 0 | 2 | 4 | 0 |
| F14 | Mean | 280.0333333 | 26.35714286 | 644.2333333 | 18.06896552 | 66.78571429 |
| | S.D. | 26.29210362 | 3.291001038 | 6.371722732 | 2.750727624 | 25.65635744 |
| | Best | 235 | 17 | 633 | 10 | 22 |
| | Worst | 344 | 31 | 657 | 21 | 105 |
| | Exec. time | 0.0942649 | 0.082852964 | 1.2564309 | 0.157990172 | 1.06789675 |
| | No. of Succeeds | 30 | 28 | 30 | 29 | 28 |
| | No. of Failures | 0 | 2 | 0 | 1 | 2 |
| F15 | Mean | NC | 39.46666667 | 70.66666667 | 33.3 | 185.5666667 |
| | S.D. | NC | 2.885556581 | 4.309839051 | 3.052980454 | 33.59359887 |
| | Best | NC | 34 | 62 | 26 | 111 |
| | Worst | NC | 44 | 77 | 39 | 241 |
| | Exec. time | NC | 0.1049036 | 0.106110233 | 0.2379349 | 1.4060248 |
| | No. of Succeeds | 0 | 30 | 30 | 30 | 30 |
| | No. of Failures | 30 | 0 | 0 | 0 | 0 |
| F16 | Mean | NC | NC | NC | NC | NC |
| | S.D. | NC | NC | NC | NC | NC |
| | Best | NC | NC | NC | NC | NC |
| | Worst | NC | NC | NC | NC | NC |
| | Exec. time | NC | NC | NC | NC | NC |
| | No. of Succeeds | 0 | 0 | 0 | 0 | 0 |
| | No. of Failures | 30 | 30 | 30 | 30 | 30 |

In Test 3, the ratio of successful minimisation of the functions for Tests 1 and 2 are presented in Tables (B.7) and (B.8).

**Table (B.7): The success and failure ratio for minimising the sixteen benchmark functions in Test 1**

| Variable dimensions | BSA | | DE | | PSO | | ABC | | FF | |
|---|---|---|---|---|---|---|---|---|---|---|
| | **Success** | **Failure** | **Success** | **Failure** | **Success** | **Failure** | **Success** | **Failure** | **Success** | **Failure** |
| Nvar1: 10 | 11 | 2 | 13 | 3 | 9 | 7 | 6 | 10 | 10 | 6 |
| Nvar2: 30 | 13 | 3 | 12 | 4 | 6 | 10 | 5 | 11 | 9 | 7 |
| Nvar3: 60 | 11 | 5 | 10 | 6 | 6 | 10 | 5 | 11 | 9 | 7 |

**Table (B.8): The success and failure ratio for minimising the sixteen benchmark functions in Test 2**

| **Search space** | **BSA** | | **DE** | | **PSO** | | **ABC** | | **FF** | |
|---|---|---|---|---|---|---|---|---|---|---|
| | **Success** | **Failure** | **Success** | **Failure** | **Success** | **Failure** | **Success** | **Failure** | **Success** | **Failure** |
| R1: [-5, 5] | 15 | 1 | 14 | 2 | 12 | 4 | 13 | 3 | 15 | 1 |
| R2: [-250, 205] | 14 | 2 | 13 | 3 | 12 | 4 | 12 | 4 | 13 | 3 |
| R3: [-500, 500] | 12 | 4 | 12 | 4 | 10 | 6 | 10 | 6 | 12 | 4 |

# APPENDIX C

# ECA* Evaluation Results

The tables presented in this section are the explanatory data results related the cluster quality measured by external measures (CI, CSI, and NMI) and internal measures (SSE, nMSE, and ε-ratio) for ECA* compared to its counterpart algorithms for 30 run average. Specifically, the tables presented in this appendix are as follows: Tables (C.1), (C.2), and (C.3) present the cluster quality measured by external measures (CI, CSI, and NMI) for ECA* compared to its counterpart algorithms for 30 run average. Tables (C.4), (C.5), and (C.6) present the cluster quality measured by internal measures (SSE, nMSE, and ε-ratio) for ECA* compared to its counterpart algorithms for 30 run average.

In Table (C.1), it shows a comparison of CI measure for ECA* with its counterpart algorithms. All the algorithms, except GENCLUST++, are successful in determining the right number of clusters for the used 32 datasets. In the meantime, GENCLUST++ is successful in most of the cases. Exceptionally, it was not successful for finding the right number of centroids indexes in A3, Compound, Dim-32, Dim-128, Dim-256, Dim-512, Dim-1024, and G2-1024-80. GENCLUST++ is more successful than in two-dimensional clustering benchmark datasets in comparison to multi-dimensional clustering benchmark datasets.

**Table (C.1): Cluster quality measured by CI for ECA* compared to its counterpart algorithms for 30 run average**

| Datasets | ECA* | KM | KM++ | EM | LVQ | GENCLUST++ | Winner |
|---|---|---|---|---|---|---|---|
| S1 | 0 | 0 | 0 | 0 | 0 | 0 | All |
| S2 | 0 | 0 | 0 | 0 | 0 | 0 | All |
| S3 | 0 | 0 | 0 | 0 | 0 | 0 | All |
| S4 | 0 | 0 | 0 | 0 | 0 | 0 | All |
| A1 | 0 | 0 | 0 | 0 | 0 | 0 | All |
| A2 | 0 | 0 | 0 | 0 | 0 | 0 | All |
| A3 | 0 | 0 | 0 | 0 | 0 | 0 | All |
| Birch1 | 0 | 0 | 0 | 0 | 0 | 22 | ECA*, KM, KM++, EM, and LVQ |
| Un-balance | 0 | 0 | 0 | 0 | 0 | 0 | |
| Aggregation | 0 | 0 | 0 | 0 | 0 | 0 | |
| Compound | 0 | 0 | 0 | 0 | 0 | 1 | ECA*, KM, KM++, EM, and LVQ |
| Path-based | 0 | 0 | 0 | 0 | 0 | 0 | All |
| D31 | 0 | 0 | 0 | 0 | 0 | 0 | All |
| R15 | 0 | 0 | 0 | 0 | 0 | 0 | All |
| Jain | 0 | 0 | 0 | 0 | 0 | 0 | All |
| Flame | 0 | 0 | 0 | 0 | 0 | 0 | All |
| Dim-32 | 0 | 0 | 0 | 0 | 0 | 3 | ECA*, KM, KM++, EM, and LVQ |
| Dim-64 | 0 | 0 | 0 | 0 | 0 | 0 | |
| Dim-128 | 0 | 0 | 0 | 0 | 0 | 3 | ECA*, KM, KM++, EM, and LVQ |
| Dim-256 | 0 | 0 | 0 | 0 | 0 | 4 | ECA*, KM, KM++, EM, and LVQ |
| Dim-512 | 0 | 0 | 0 | 0 | 0 | 9 | ECA*, KM, KM++, EM, and LVQ |
| Dim-1024 | 0 | 0 | 0 | 0 | 0 | 5 | ECA*, KM, KM++, EM, and LVQ |
| G2-16-10 | 0 | 0 | 0 | 0 | 0 | 0 | All |
| G2-16-30 | 0 | 0 | 0 | 0 | 0 | 0 | All |
| G2-16-60 | 0 | 0 | 0 | 0 | 0 | 0 | All |
| G2-16-80 | 0 | 0 | 0 | 0 | 0 | 0 | All |
| G2-16-100 | 0 | 0 | 0 | 0 | 0 | 0 | All |
| G2-1024-10 | 0 | 0 | 0 | 0 | 0 | 0 | All |
| G2-1024-30 | 0 | 0 | 0 | 0 | 0 | 0 | All |
| G2-1024-60 | 0 | 0 | 0 | 0 | 0 | 0 | All |
| G2-1024-80 | 0 | 0 | 0 | 0 | 0 | 1 | ECA*, KM, KM++, EM, and LVQ |
| G2-1024-100 | 0 | 0 | 0 | 0 | 0 | 0 | All |
| Average | 0 | 0 | 0 | 0 | 0 | 1.5 | ECA*, KM, KM++, EM, and LVQ |

Table (C.2) presents a comparison of CSI measure for ECA* with its counterpart algorithms. After ECA*, KM, EM, KM++, LQV and GENGLUST++. In a few cases, the algorithms measured 1 or approximate to 1 for CSI value. Overall, ECA* is successful in having the optimal value of CSI for all the datasets, except S4. Alongside with its success, ECA*'s counterpart algorithms are successful in some datasets. Mainly, KM is successful in almost of shape datasets, including Aggregation, Compound, Path-based, D31, R15, Jain, and Flame, and three of the multi-dimensional clustering benchmark datasets, including G2-16-10, G2-16-30, and G2-1024-100. After that, EM is the third successful technique, in which a winner in six datasets (S2, S4, Compound, R15, Jain, and G2-1024-100), followed by KM++ which is the fourth winner in four datasets (R15, Dim-256, Dim-256, and G2-1024-100). Meanwhile, LVQ and GENCLUST++ are the least successful winner in only two datasets.

**Table (C.2): Cluster quality measured by CSI for ECA* compared to its counterpart algorithms for 30 run average**

| Datasets | ECA* | KM | KM++ | EM | LVQ | GENCLUST++ | Winner |
|---|---|---|---|---|---|---|---|
| S1 | 0.994 | 0.681 | 0.7897 | 0.9865 | 0.6975 | 0.807 | ECA* |
| S2 | 0.977 | 0.4935 | 0.9595 | 0.9795 | 0.704 | 0.9165 | ECA*, and EM |
| S3 | 0.9809 | 0.5175 | 0.87195 | 0.963 | 0.7585 | 0.939 | ECA* |
| S4 | 0.872 | 0.6405 | 0.58025 | 0.9005 | 0.8245 | 0.8315 | EM |
| A1 | 1 | 0.613 | 0.954 | 0.9825 | 0.981 | 0.947 | ECA* |
| A2 | 0.9785 | 0.584 | 0.9395 | 0.9425 | 0.9635 | 0.97 | ECA*, and GENCLUST++ |
| A3 | 0.9995 | 0.8295 | 0.982 | 0.9935 | 0.8855 | 0.954 | ECA* |
| Birch1 | 0.999 | 0.645 | 0.669 | 0.717 | 0.278 | 0.23 | ECA* |
| Un-balance | 1 | 0.503 | 0.818 | 0.921 | 0.9515 | 0.8005 | ECA* |
| Aggregation | 1 | 1 | 0.782 | 0.978 | 0.938 | 0.963 | ECA*, and KM |
| Compound | 1 | 1 | 0.835 | 1 | 0.973 | 0.9395 | ECA*, KM, and EM |
| Path-based | 1 | 1 | 0.3445 | 0.994 | 0.9545 | 0.887 | ECA*, and KM |
| D31 | 1 | 1 | 0.7665 | 0.9945 | 0.9715 | 0.958 | ECA*, and KM |
| R15 | 1 | 1 | 1 | 1 | 1 | 0.955 | ECA*, KM, KM++, EM, and LVQ |
| Jain | 1 | 1 | 0.205 | 1 | 0.9435 | 0.9435 | ECA*, KM, and EM |
| Flame | 1 | 1 | 0.225 | 0.9195 | 0.975 | 0.975 | ECA*, and KM |
| Dim-32 | 0.999 | 0.886 | 0.997 | 0.665 | 0.997 | 0.415 | ECA* |
| Dim-64 | 0.915 | 0.884 | 0.729 | 0.602 | 0.937 | 0.548 | ECA* |
| Dim-128 | 0.907 | 0.78 | 0.999 | 0.78 | 0.997 | 0.543 | ECA* |
| Dim-256 | 0.974 | 0.689 | 0.998 | 0.689 | 0.533 | 0.428 | KM++ |
| Dim-512 | 0.871 | 0.574 | 0.88 | 0.507 | 0.506 | 0.367 | ECA* |
| Dim-1024 | 0.934 | 0.662 | 0.934 | 0.662 | 0.507 | 0.454 | ECA*, and KM++ |
| G2-16-10 | 1 | 1 | 0.998 | 0.999 | 0.999 | 1 | ECA*, KM, and GENCLUST++ |
| G2-16-30 | 1 | 1 | 0.998 | 0.998 | 0.998 | 0.998 | ECA*, and KM |
| G2-16-60 | 0.998 | 0.997 | 0.997 | 0.997 | 0.933 | 0.996 | ECA* |
| G2-16-80 | 0.997 | 0.995 | 0.995 | 0.995 | 0.966 | 0.995 | ECA* |
| G2-16-100 | 0.999 | 0.996 | 0.996 | 0.997 | 0.908 | 0.951 | ECA* |
| G2-1024-10 | 1 | 0.999 | 0.99 | 0.999 | 0.999 | 0.953 | ECA* |
| G2-1024-30 | 0.999 | 0.998 | 0.998 | 0.998 | 0.998 | 0.953 | ECA* |
| G2-1024-60 | 0.998 | 0.997 | 0.997 | 0.997 | 0.997 | 0.952 | ECA* |
| G2-1024-80 | 0.997 | 0.996 | 0.996 | 0.996 | 0.996 | 0.969 | ECA* |
| G2-1024-100 | 0.996 | 0.996 | 0.996 | 0.996 | 0.996 | 0.849 | ECA*, KM, KM++, EM, and LVQ |
| Average | 0.981 | 0.842 | 0.851 | 0.911 | 0.877 | 0.825 | ECA* |

Table (C.3) presents a comparison of cluster quality measured by NMI for ECA* and its competitive algorithms. For all the used datasets, ECA* cluster centroid results have the best correlation between its cluster centroids and ground truth centroids of the same matched clusters. Afterwards, KM has relatively good relations of its cluster centroids with the ground truth centroids of the same matched clusters, Particularly, for the two-dimensional benchmark datasets. In terms of this correlation, KM is followed by KM++, EM, and LVQ. It needs to mention that GENCLUST++ records the least correlation for almost all the current datasets.

**Table (C.3): Cluster quality measured by NMI for ECA* compared to its counterpart algorithms for 30 run average**

| **Datasets** | **ECA*** | **KM** | **KM++** | **EM** | **LVQ** | **GENCLUST++** | **Winner** |
|---|---|---|---|---|---|---|---|
| S1 | 1.000 | 1.000 | 1.000 | 1.000 | 1.000 | 0.972 | ECA*, KM, KM++, EM, and LVQ |
| S2 | 1.000 | 1.000 | 1.000 | 1.000 | 1.000 | 0.994 | ECA*, KM, KM++, EM, and LVQ |
| S3 | 1.000 | 1.000 | 1.000 | 1.000 | 1.000 | 0.993 | ECA*, KM, KM++, EM, and LVQ |
| S4 | 1.000 | 1.000 | 1.000 | 1.000 | 1.000 | 0.972 | ECA*, KM, KM++, EM, and LVQ |
| A1 | 1.000 | 1.000 | 1.000 | 1.000 | 1.000 | 0.865 | ECA*, KM, KM++, EM, and LVQ |
| A2 | 1.000 | 1.000 | 0.997 | 1.000 | 1.000 | 0.970 | ECA*, KM, EM, and LVQ |
| A3 | 1.000 | 1.000 | 1.000 | 1.000 | 1.000 | 0.988 | ECA*, KM, KM++, EM, and LVQ |
| Birch1 | 0.989 | 0.954 | 0.934 | 0.943 | 0.951 | 0.989 | ECA*, and GENCLUST++ |
| Un-balance | 1.000 | 1.000 | 1.000 | 1.000 | 0.935 | 0.870 | ECA*, KM, KM++, and EM |
| Aggregation | 0.967 | 0.967 | 0.824 | 0.853 | 0.737 | 0.796 | ECA*, and KM |
| Compound | 1.000 | 1.000 | 0.767 | 0.876 | 0.832 | 0.654 | ECA*, and KM |
| Path-based | 0.923 | 0.923 | 0.826 | 0.852 | 0.862 | 0.390 | ECA*, and KM |
| D31 | 0.838 | 0.950 | 0.700 | 0.899 | 0.660 | 0.704 | KM |
| R15 | 0.653 | 1.000 | 1.000 | 0.831 | 0.821 | 0.723 | KM, KM++ |
| Jain | 1.000 | 1.000 | 0.866 | 0.866 | 1.000 | 0.397 | ECA*, KM, and LVQ |
| Flame | 0.965 | 0.866 | 0.666 | 0.667 | 0.866 | 0.798 | ECA* |
| Dim-32 | 0.807 | 0.763 | 0.761 | 0.763 | 0.723 | 0.754 | ECA* |
| Dim-64 | 0.708 | 0.661 | 0.663 | 0.661 | 0.624 | 0.675 | ECA* |
| Dim-128 | 0.613 | 0.537 | 0.552 | 0.537 | 0.533 | 0.491 | ECA* |
| Dim-256 | 0.475 | 0.423 | 0.446 | 0.423 | 0.455 | 0.389 | ECA* |
| Dim-512 | 0.330 | 0.318 | 0.312 | 0.318 | 0.319 | 0.296 | ECA* |
| Dim-1024 | 0.283 | 0.204 | 0.249 | 0.204 | 0.310 | 0.198 | LVQ |
| G2-16-10 | 0.715 | 0.715 | 0.715 | 0.715 | 0.715 | 0.709 | ECA*, KM, KM++, EM, and LVQ |
| G2-16-30 | 0.613 | 0.613 | 0.613 | 0.613 | 0.613 | 0.600 | ECA*, KM, KM++, EM, and LVQ |
| G2-16-60 | 0.571 | 0.532 | 0.532 | 0.532 | 0.407 | 0.499 | ECA* |
| G2-16-80 | 0.505 | 0.493 | 0.489 | 0.491 | 0.477 | 0.494 | ECA* |
| G2-16-100 | 0.566 | 0.528 | 0.534 | 0.532 | 0.433 | 0.484 | ECA* |
| G2-1024-10 | 0.702 | 0.688 | 0.688 | 0.688 | 0.688 | 0.549 | ECA* |
| G2-1024-30 | 0.584 | 0.583 | 0.583 | 0.583 | 0.583 | 0.551 | ECA*, KM, KM++, EM, and LVQ |
| G2-1024-60 | 0.527 | 0.513 | 0.513 | 0.513 | 0.513 | 0.503 | ECA* |
| G2-1024-80 | 0.506 | 0.491 | 0.491 | 0.491 | 0.491 | 0.503 | ECA*, and GENCLUST++ |
| G2-1024-100 | 0.498 | 0.473 | 0.473 | 0.473 | 0.473 | 0.337 | ECA* |
| Average | 0.761 | 0.756 | 0.725 | 0.729 | 0.719 | 0.660 | ECA* |

Table (C.4) presents a comparison of external cluster evaluation measured by SSE for ECA* and its counterpart algorithms to measure the squared differences between each data observation with its cluster centroid. In this comparison, ECA* performs well in most of the datasets, and it is the winner in 25 datasets. On the other hand, the counterpart algorithms of ECA* are the winner in a few numbers of datasets. Therefore, ECA* performs well compared to its competitive algorithms in the current datasets.

**Table (C.4): Cluster quality measured by SSE for ECA* compared to its counterpart algorithms for 30 run average**

| Datasets | ECA* | KM | KM++ | EM | LVQ | GENCLUST++ | Winner |
|---|---|---|---|---|---|---|---|
| S1 | 9.093E+12 | 1.217E+13 | 1.423E+13 | 9.085E+12 | 3.107E+13 | 2.137E+13 | ECA*, and EM |
| S2 | 1.420E+13 | 1.854E+13 | 1.328E+13 | 1.389E+13 | 4.751E+13 | 1.321E+13 | GENCLUST++ |
| S3 | 1.254E+13 | 2.284E+13 | 1.892E+13 | 1.193E+12 | 4.640E+13 | 1.987E+13 | ECA* |
| S4 | 9.103E+12 | 1.685E+13 | 1.671E+13 | 2.267E+13 | 3.539E+13 | 2.118E+13 | ECA* |
| A1 | 1.215E+10 | 1.874E+10 | 1.271E+10 | 1.578E+10 | 5.995E+10 | 4.889E+10 | ECA* |
| A2 | 7.103E+09 | 3.797E+10 | 3.375E+10 | 3.266E+10 | 8.132E+10 | 3.358E+10 | ECA* |
| A3 | 2.949E+10 | 5.225E+10 | 4.106E+10 | 4.508E+10 | 1.143E+11 | 5.584E+10 | ECA* |
| Birch1 | 7.007E+09 | 2.102E+11 | 2.102E+10 | 1.752E+10 | 4.204E+11 | 8.408E+09 | ECA* |
| Un-balance | 2.145E+11 | 2.188E+12 | 1.173E+12 | 1.329E+12 | 5.459E+12 | 5.968E+12 | ECA* |
| Aggregation | 1.024E+04 | 1.180E+04 | 1.375E+04 | 1.377E+04 | 6.072E+04 | 9.025E+03 | GENCLUST++ |
| Compound | 4.197E+03 | 5.578E+03 | 5.710E+03 | 5.632E+03 | 1.465E+04 | 8.544E+03 | ECA* |
| Path-based | 4.615E+03 | 8.959E+03 | 8.958E+03 | 9.628E+03 | 1.716E+04 | 4.621E+03 | ECA* |
| D31 | 3.495E+03 | 5.251E+04 | 5.177E+03 | 4.148E+03 | 1.027E+04 | 5.155E+05 | ECA* |
| R15 | 1.092E+02 | 1.636E+02 | 1.086E+02 | 1.704E+02 | 2.097E+02 | 3.244E+02 | ECA* |
| Jain | 1.493E+04 | 2.377E+04 | 2.377E+04 | 2.532E+04 | 4.305E+04 | 2.719E+03 | GENCLUST++ |
| Flame | 3.302E+03 | 3.192E+03 | 3.124E+03 | 3.330E+03 | 3.352E+03 | 1.240E+03 | GENCLUST++ |
| Dim-32 | 1.618E+05 | 3.786E+07 | 2.325E+05 | 7.555E+06 | 2.359E+05 | 2.323E+08 | ECA* |
| Dim-64 | 1.814E+06 | 6.234E+07 | 6.515E+07 | 5.580E+07 | 2.170E+05 | 3.860E+08 | LVQ |
| Dim-128 | 1.958E+04 | 2.451E+09 | 5.684E+07 | 1.092E+08 | 2.691E+05 | 1.787E+07 | ECA* |
| Dim-256 | 1.255E+04 | 1.764E+08 | 4.202E+08 | 4.809E+07 | 2.433E+05 | 9.499E+07 | ECA* |
| Dim-512 | 2.969E+05 | 2.618E+08 | 2.768E+08 | 6.565E+08 | 2.969E+05 | 5.909E+08 | LVQ |
| Dim-1024 | 1.992E+05 | 7.474E+08 | 2.755E+05 | 5.596E+08 | 1.244E+09 | 3.107E+05 | ECA* |
| G2-16-10 | 2.048E+05 | 3.277E+06 | 3.277E+06 | 1.667E+08 | 3.277E+06 | 3.280E+06 | ECA* |
| G2-16-30 | 1.825E+06 | 2.920E+07 | 2.262E+08 | 2.920E+07 | 2.920E+07 | 2.921E+07 | ECA* |
| G2-16-60 | 1.045E+07 | 1.165E+08 | 1.165E+08 | 1.165E+08 | 1.909E+08 | 1.166E+08 | ECA* |
| G2-16-80 | 1.701E+07 | 2.092E+08 | 2.092E+08 | 2.092E+08 | 2.502E+08 | 2.096E+08 | ECA* |
| G2-16-100 | 3.259E+08 | 4.868E+08 | 3.268E+08 | 3.268E+08 | 4.023E+08 | 5.412E+08 | KM++, and EM |
| G2-1024-10 | 2.097E+07 | 2.097E+08 | 2.097E+08 | 2.097E+08 | 2.097E+08 | 1.333E+10 | ECA* |
| G2-1024-30 | 2.086E+08 | 1.883E+09 | 1.883E+09 | 1.883E+09 | 1.883E+09 | 1.501E+10 | ECA* |
| G2-1024-60 | 8.074E+08 | 7.537E+09 | 7.537E+09 | 7.537E+09 | 7.537E+09 | 2.067E+10 | ECA* |
| G2-1024-80 | 3.040E+09 | 1.340E+10 | 1.340E+10 | 1.340E+10 | 1.340E+10 | 1.340E+10 | ECA* |
| G2-1024-100 | 2.221E+09 | 4.247E+10 | 2.093E+10 | 2.093E+10 | 2.093E+10 | 2.616E+10 | ECA* |
| Average | 1.413E+12 | 2.280E+12 | 2.015E+12 | 1.510E+12 | 5.205E+12 | 2.557E+12 | ECA* |

In addition to Table (C.4), Table (C.5) shows a comparison of external cluster evaluation measured by nMSE for ECA* and its counterpart algorithms to measure the average of squared error, and the mean squared difference between the actual value and the estimated values. Similar to the previous evaluation measure, ECA* performs well compared to its counterpart techniques in most of the datasets. After that, GENCLUS++ is the second winner that has good results in four datasets (S2,

Aggregation, Flame, and G2-16-80). On the other hand, KM is not a winner in all cases, whereas KM++ is the winner in two datasets (A3, and G2-1024-100). It is necessary to mention that KM++, EM, and LVQ are the winners for G2-1024-100 dataset. Also, EM is the winner for S1, and LVQ is the winner for two other datasets (Dim-512, and G2-16-100).

**Table (C.5): Cluster quality measured by nMSE for ECA* compared to its counterpart algorithms for 30 run average**

| Datasets | ECA* | KM | KM++ | EM | LVQ | GENCLUST++ | Winner |
|---|---|---|---|---|---|---|---|
| S1 | 9.093E+08 | 1.217E+09 | 1.423E+09 | 9.085E+08 | 3.107E+09 | 2.137E+09 | ECA* |
| S2 | 1.420E+09 | 1.854E+09 | 1.328E+09 | 1.389E+09 | 4.751E+09 | 1.321E+09 | GENCLUST++ |
| S3 | 1.254E+09 | 2.284E+09 | 1.892E+09 | 1.193E+08 | 4.640E+09 | 1.987E+09 | ECA* |
| S4 | 9.103E+08 | 1.685E+09 | 1.671E+09 | 2.267E+09 | 3.539E+09 | 2.118E+09 | ECA* |
| A1 | 2.026E+06 | 3.123E+06 | 2.118E+06 | 2.629E+06 | 9.991E+06 | 8.148E+06 | ECA* |
| A2 | 1.184E+06 | 3.616E+06 | 3.214E+06 | 3.111E+06 | 7.745E+06 | 3.198E+06 | ECA* |
| A3 | 4.915E+06 | 3.483E+06 | 2.737E+06 | 3.006E+06 | 7.619E+06 | 3.723E+06 | |
| Birch1 | 3.504E+04 | 1.051E+06 | 1.051E+05 | 8.759E+04 | 2.102E+06 | 4.204E+04 | ECA* |
| Un-balance | 1.650E+07 | 1.683E+08 | 9.026E+07 | 1.022E+08 | 4.199E+08 | 4.591E+08 | ECA* |
| Aggregation | 6.499E+00 | 7.490E+00 | 8.726E+00 | 8.734E+00 | 3.853E+01 | 5.726E+00 | GENCLUST++ |
| Compound | 5.259E+00 | 6.990E+00 | 7.156E+00 | 7.058E+00 | 1.836E+01 | 1.071E+01 | ECA* |
| Path-based | 7.691E+00 | 1.493E+01 | 1.493E+01 | 1.605E+01 | 2.861E+01 | 7.701E+00 | ECA* |
| D31 | 5.637E-01 | 8.470E+00 | 8.350E-01 | 6.691E-01 | 1.656E+00 | 8.314E+01 | ECA* |
| R15 | 9.099E-02 | 1.364E-01 | 9.052E-02 | 1.420E-01 | 1.747E-01 | 2.704E-01 | ECA* |
| Jain | 2.001E+01 | 3.186E+01 | 3.186E+01 | 3.394E+01 | 5.771E+01 | 3.644E+00 | ECA* |
| Flame | 6.880E+00 | 6.650E+00 | 6.508E+00 | 6.937E+00 | 6.984E+00 | 2.584E+00 | GENCLUST++ |
| Dim-32 | 4.938E+00 | 1.155E+03 | 7.096E+00 | 2.306E+02 | 7.200E+00 | 7.090E+03 | ECA* |
| Dim-64 | 4.324E-01 | 9.512E+02 | 9.942E+02 | 8.515E+02 | 3.311E+00 | 5.890E+03 | ECA* |
| Dim-128 | 1.494E-01 | 1.870E+04 | 4.337E+02 | 8.330E+02 | 2.053E+00 | 1.363E+02 | ECA* |
| Dim-256 | 4.789E-02 | 6.731E+02 | 1.603E+03 | 1.834E+02 | 9.281E-01 | 3.624E+02 | ECA* |
| Dim-512 | 5.663E-01 | 4.993E+02 | 5.279E+02 | 1.252E+03 | 5.663E-01 | 1.127E+03 | ECA*, and LVQ |
| Dim-1024 | 1.900E-01 | 7.128E+02 | 2.627E-01 | 5.337E+02 | 1.186E+03 | 2.963E-01 | ECA* |
| G2-16-10 | 6.250E+00 | 9.999E+01 | 9.999E+01 | 5.086E+03 | 9.999E+01 | 1.001E+02 | ECA* |
| G2-16-30 | 5.569E+01 | 8.911E+02 | 6.902E+03 | 8.911E+02 | 8.911E+02 | 8.915E+02 | ECA* |
| G2-16-60 | 3.190E+02 | 3.556E+03 | 3.556E+03 | 3.555E+03 | 5.825E+03 | 3.558E+03 | ECA* |
| G2-16-80 | 5.192E+02 | 6.385E+03 | 6.385E+03 | 9.975E+01 | 1.193E+02 | 9.995E+01 | GENCLUST++ |
| G2-16-100 | 9.947E+03 | 1.486E+04 | 9.973E+03 | 1.558E+02 | 1.918E+02 | 2.581E+02 | LVQ |
| G2-1024-10 | 1.000E+01 | 9.998E+01 | 9.998E+01 | 9.998E+01 | 9.998E+01 | 6.356E+03 | ECA* |
| G2-1024-30 | 9.947E+01 | 8.979E+02 | 8.979E+02 | 8.979E+02 | 8.979E+02 | 7.157E+03 | ECA* |
| G2-1024-60 | 3.850E+02 | 3.594E+03 | 3.594E+03 | 3.594E+03 | 3.594E+03 | 9.854E+03 | ECA* |
| G2-1024-80 | 1.450E+03 | 6.388E+03 | 6.388E+03 | 6.388E+03 | 6.388E+03 | 6.388E+03 | ECA* |
| G2-1024-100 | 1.059E+04 | 2.025E+04 | 9.982E+03 | 9.982E+03 | 9.982E+03 | 1.248E+04 | KM++, EM, and LVQ |
| Average | 1.412E+08 | 2.256E+08 | 2.004E+08 | 1.498E+08 | 5.151E+08 | 2.511E+08 | ECA* |

Finally, Table (C.6) depicts a comparison of external cluster evaluation measured by ε- ratio for ECA* and its counterpart algorithms to compare the results with the theoretical results achieved for approximation techniques. ECA* performs well compare to its counterpart techniques in most of the datasets. After that, EM is the second winner that has good results in three datasets (S3, R15, and G2-16-100). Meanwhile, KM++, LVQ, and GENCLUS++ are the winner in only one case, whereas KM does not win in any of the 32 datasets.

**Table (C.6): Cluster quality measured by ε- ratio for ECA* compared to its counterpart algorithms for 30 run average**

| Datasets | ECA* | KM | KM++ | EM | LVQ | GENCLUST++ | Winner |
|---|---|---|---|---|---|---|---|
| S1 | 9.093E+15 | 1.217E+16 | 1.423E+16 | 9.085E+15 | 3.107E+16 | 2.137E+16 | ECA* |
| S2 | 1.420E+16 | 1.854E+16 | 1.328E+16 | 1.389E+16 | 4.751E+16 | 1.321E+16 | GENCLUST++ |
| S3 | 1.254E+16 | 2.284E+16 | 1.892E+16 | 1.193E+15 | 4.640E+16 | 1.987E+16 | EM |
| S4 | 9.103E+15 | 1.685E+16 | 1.671E+16 | 2.267E+16 | 3.539E+16 | 2.118E+16 | ECA* |
| A1 | 1.215E+13 | 1.874E+13 | 1.271E+13 | 1.578E+13 | 5.995E+13 | 4.889E+13 | ECA* |
| A2 | 7.103E+12 | 3.797E+13 | 3.375E+13 | 3.266E+13 | 8.132E+13 | 3.358E+13 | ECA* |
| A3 | 2.949E+13 | 5.225E+13 | 4.106E+13 | 4.508E+13 | 1.143E+14 | 5.584E+13 | ECA* |
| Birch1 | 7.007E+12 | 2.102E+14 | 2.102E+13 | 1.752E+13 | 4.204E+14 | 8.408E+12 | ECA* |
| Un-balance | 2.145E+14 | 2.188E+15 | 1.173E+15 | 1.329E+15 | 5.459E+15 | 5.968E+15 | ECA* |
| Aggregation | 1.024E+07 | 1.180E+07 | 1.375E+07 | 1.377E+07 | 6.072E+07 | 9.025E+06 | ECA* |
| Compound | 4.197E+06 | 5.578E+06 | 5.710E+06 | 5.632E+06 | 1.465E+07 | 8.544E+06 | ECA* |
| Path-based | 4.615E+06 | 8.959E+06 | 8.958E+06 | 9.628E+06 | 1.716E+07 | 4.621E+06 | ECA* |
| D31 | 3.495E+06 | 5.251E+07 | 5.177E+06 | 4.148E+06 | 1.027E+07 | 5.155E+08 | ECA* |
| R15 | 1.092E+05 | 1.636E+05 | 1.086E+05 | 1.704E+05 | 2.097E+05 | 3.244E+05 | EM |
| Jain | 1.493E+07 | 2.377E+07 | 2.377E+07 | 2.532E+07 | 4.305E+07 | 2.719E+06 | ECA* |
| Flame | 3.302E+06 | 3.192E+06 | 3.124E+06 | 3.330E+06 | 3.352E+06 | 1.240E+06 | GENCLUST++ |
| Dim-32 | 1.618E+08 | 3.786E+10 | 2.325E+08 | 7.555E+09 | 2.359E+08 | 2.323E+11 | ECA* |
| Dim-64 | 2.834E+07 | 6.234E+10 | 6.515E+10 | 5.580E+10 | 2.170E+08 | 3.860E+11 | ECA* |
| Dim-128 | 1.958E+07 | 2.451E+12 | 5.684E+10 | 1.092E+11 | 2.691E+08 | 1.787E+10 | ECA* |
| Dim-256 | 1.255E+07 | 1.764E+11 | 4.202E+11 | 4.809E+10 | 2.433E+08 | 9.499E+10 | ECA* |
| Dim-512 | 2.969E+08 | 2.618E+11 | 2.768E+11 | 6.565E+11 | 2.969E+08 | 5.909E+11 | ECA*, and LVQ |
| Dim-1024 | 1.992E+08 | 7.474E+11 | 2.755E+08 | 5.596E+11 | 1.244E+12 | 3.107E+08 | ECA* |
| G2-16-10 | 3.259E+11 | 3.277E+09 | 3.277E+09 | 1.667E+11 | 3.277E+09 | 3.280E+09 | ECA* |
| G2-16-30 | 1.825E+09 | 2.920E+10 | 2.262E+11 | 2.920E+10 | 2.920E+10 | 2.921E+10 | ECA* |
| G2-16-60 | 1.045E+10 | 1.165E+11 | 1.165E+11 | 1.165E+11 | 1.909E+11 | 1.166E+11 | ECA* |
| G2-16-80 | 1.701E+10 | 2.092E+11 | 2.092E+11 | 2.092E+11 | 2.502E+11 | 2.096E+11 | ECA* |
| G2-16-100 | 3.879E+11 | 4.868E+11 | 3.268E+11 | 3.268E+11 | 4.023E+11 | 5.412E+11 | KM++, and EM |
| G2-1024-10 | 2.097E+10 | 2.097E+11 | 2.097E+11 | 2.097E+11 | 2.097E+11 | 1.333E+13 | ECA* |
| G2-1024-30 | 2.086E+11 | 1.883E+12 | 1.883E+12 | 1.883E+12 | 1.883E+12 | 1.501E+13 | ECA* |
| G2-1024-60 | 8.074E+11 | 7.537E+12 | 7.537E+12 | 7.537E+12 | 7.537E+12 | 2.067E+13 | ECA* |
| G2-1024-80 | 3.040E+12 | 1.340E+13 | 1.340E+13 | 1.340E+13 | 1.340E+13 | 1.340E+13 | ECA* |
| G2-1024-100 | 2.221E+13 | 4.247E+13 | 2.093E+13 | 2.093E+13 | 2.093E+13 | 2.616E+13 | ECA* |
| Average | 1.413E+15 | 2.280E+15 | 2.015E+15 | 1.510E+15 | 5.205E+15 | 2.557E+15 | ECA* |